\documentclass{article}

\PassOptionsToPackage{numbers, compress}{natbib}

\usepackage[final]{lib/neurips_2021}

\usepackage[utf8]{inputenc} 
\usepackage[T1]{fontenc}    
\usepackage{hyperref}       
\usepackage{url}            
\usepackage{booktabs}       
\usepackage{amsfonts}       
\usepackage{nicefrac}       
\usepackage{microtype}      
\usepackage{xcolor}         
\usepackage{graphicx}

\usepackage{todonotes}
\usepackage{wrapfig}
\usepackage{pbox}
\usepackage{bbm,bm}
\usepackage{mathtools}
\usepackage{stackengine}
\usepackage{amsthm}
\usepackage{multirow}
\usepackage{dsfont}
\usepackage{subcaption}
\usepackage{caption}
\usepackage{cleveref}

\title{\textsc{Doctor}: A Simple Method for Detecting Misclassification {E}rrors}

\author{%
  Federica Granese\thanks{\footnotesize{These authors contributed equally to this work.}}~~\thanks{\footnotesize{This paper is supported by the ERC project Hypatia under the European Unions Horizon 2020 research and innovation program. Grant agreement N. 835294.}}\\
  Lix, Inria, Institute Polytechnique de Paris,\\
  Sapienza University of Rome\\
  \texttt{federica.granese@inria.fr} \\
  \And
  Marco Romanelli\footnotemark[1]\\
L2S, CentraleSupélec,\\CNRS, Université Paris Saclay\\
  \texttt{marco.romanelli@centralesupelec.fr} \\
  \AND
  Daniele Gorla \\
  Sapienza University of Rome\\
  \texttt{gorla@di.uniroma1.it} \\
  \And
  Catuscia Palamidessi$^\dag$ \\
Lix, Inria, Institute Polytechnique de Paris,\\
  \texttt{catuscia@lix.polytechnique.fr} \\
  \And
  Pablo Piantanida\thanks{\footnotesize{This project has received funding
from the European Union’s Horizon 2020 research and innovation programme under the Marie Skłodowska-Curie grant agreement N. 792464.}} \\
L2S, CentraleSupélec,\\CNRS, Université Paris Saclay\\
  \texttt{pablo.piantanida@centralesupelec.fr} \\
}

\theoremstyle{definition}
\newtheorem{theorem}{Theorem}[section]

\newtheorem{proposition}[theorem]{Proposition}

\newtheorem{definition}{Definition}

\crefname{proposition}{Proposition}{Propositions}
\Crefname{proposition}{Proposition}{Propositions}

\crefname{definition}{Definition}{Definitions}
\Crefname{definition}{Definition}{Definitions}

\newcommand{\fun}[1]{\text{#1}}
\def\myeq{\mathrel{\ensurestackMath{\stackon[1pt]{=}{\scriptscriptstyle\Delta}}}}

\newcommand{\pe}{\fun{Pe}}
\newcommand{\pehat}{\widehat{\fun{Pe}}}
\newcommand{\ghat}{1 - \widehat{\fun{g}}}
\newcommand{\gh}{\widehat{\fun{g}}}

\newcommand{\FedRev}[1]{\textcolor{black}{#1}}

\newcommand\Tstrut{\rule{0pt}{2.6ex}}         
\newcommand\Bstrut{\rule[-0.9ex]{0pt}{0pt}}   

\begin{document}

\maketitle

\begin{abstract}
Deep neural networks (DNNs) have shown to perform very well on large scale object recognition problems and lead to widespread use for real-world applications, including situations where DNN are implemented as ``black boxes''.  A promising approach to secure their use is to accept decisions that are likely to be correct while discarding the others. In this work, we propose \textsc{Doctor}, a simple method that aims to identify whether the prediction of a DNN classifier should (or should not) be trusted so that, consequently, it would be possible to accept it or to reject it.  Two scenarios are investigated: Totally Black Box (TBB) where only the soft-predictions are available and Partially Black Box (PBB) where gradient-propagation  to perform input pre-processing is allowed. Empirically, we show that \textsc{Doctor} outperforms all state-of-the-art methods on various well-known images and sentiment analysis datasets. In particular, we observe a reduction of up to $4\%$ of the false rejection rate (FRR) in the PBB scenario. \textsc{Doctor} can be applied to any pre-trained model, it  does not require prior information about the underlying dataset and is as simple as the simplest available methods in the literature.
\end{abstract}

\section{Introduction}
\label{sec:intro}

With the advancement of state-of-the-art Deep Neural Networks (DNNs), there has been rapid adoption of these technologies in a broad range of applications to critical systems, such as autonomous driving vehicles or industrial robots, including--but not limited to--classification and decision making tasks.  Nevertheless, these solutions still exhibit unwanted behaviors as they tend to be overconfident even in presence of wrong decisions~\cite{KristiadiHH2020ICML}.  Developing methods and tools to make these algorithms reliable, in particular for non-specialists who may treat them as “black boxes” with no further checks, constitutes a core challenge. Recently,  the study of safety AI methods has gained ground, and many efforts have been made in several areas ~\cite{GeifmanY2019ICML,GeifmanUY2019ICLR,GuoPSW2017ICML,HeinAB2018CVPR,MeinkeH2020ICLR,XingAZPICLR2020,ZhangDS2019CoRR}. In this paper, we investigate  a simple method capable of detecting  whether a prediction of a classifier is likely to be correct, and therefore  should be trusted, or it is not, and should be rejected. 

Deep learning pursues the idea of learning effective representations from the data itself by training with the implicit assumption that the test data distribution should be similar to the training data distribution. However, when applied to real-world tasks, this assumption does not hold true, leading to a significant increase of misclassification errors. Although classic approaches to Out-Of-Distribution (OOD) detection~\cite{ChenLWLJ2020CoRR,HendrycksG2017ICLR,LeeLLS2019ICLR,LiangLS2018ICLR,VyasJZDKW2018EECV} are not directly concerned with detecting misclassification errors, they are intended to prevent those errors indirectly by identifying potential drifts of the testing distribution. What the above ODD methods have in common with our work is that samples drawn from the in-distribution are more likely to be correctly classified than those from a different distribution. Indeed, the model's soft-predictions for in-distribution samples tend to be generally peaky in correspondence to the correct class label while they tend to be less peaky for input samples drawn from a different distribution~\cite{HendrycksG2017ICLR}. In general, most of these works consider \textit{white-box} scenarios, where the hidden layers of the architecture are accessible or the corresponding weights are tuned during the training phase. A very effective approach to OOD detection is ODIN \cite{LiangLS2018ICLR} which involves the use of temperature scaling and the addition of small perturbations to input samples. A related solution is introduced in~\cite{GeifmanE2017NIPS} where the maximum soft-probability is called \textit{softmax response}. Within this approach, the softmax response decides whether the classifier is confident enough in its prediction or not. A different approach to OOD detection is given by the use of the Mahalanobis distance~\cite{HsuSJK2020IEEEX,LeeLLS2018NeurIPS}, which consists in calculating how much the observed  out-distribution sample deviate from the in-distribution ones but assuming the latter are given.


\subsection{Summary of contributions} 

Our work tackles the problem of  identifying whether the prediction of a classifier should or should not be trusted, no matter if they are made on out or in-distribution samples, and advances the state-of-the-art in multiple ways.
\begin{itemize}
    \item From the theoretical point of view, we derive the trade-off between two types of error probabilities: Type-I, that refers to the rejection of the classification for an input that would be correctly classified,  and Type-II, that refers to the acceptance of the classification for an input  that would not be correctly classified (\cref{prop:1}). The characterization of the optimal discriminator in \cref{eq:opt'} allows us to devise a feasible implementation of it, based on the softmax probability (\cref{prop:2}).

\item From the algorithmic point of view, inspired by our theoretical analysis, we propose \textsc{Doctor} a new discriminator (\cref{def:doctor}), which yields a simple and flexible framework to detect whether a decision made by a model is likely to be correct or not. We distinguish two scenarios under which DOCTOR can be deployed: Totally Black Box (TBB) where only the soft-predictions are available, hence gradient-propagation  to perform input pre-processing is not allowed,  and Partially Black Box (PBB) where we further allow method-specific inputs perturbations.

\item  From the experimental point of view, we show that \textsc{Doctor} outperforms comparable state-of-the-art methods (e.g., ODIN~\cite{LiangLS2018ICLR}, softmax response~\cite{GeifmanE2017NIPS} and Mahalanobis distance~\cite{LeeLLS2018NeurIPS}) on datasets including both in-distribution and out-of-distribution samples, and different architectures  with various  expressibilities, under both TBB and PBB. A key ingredient of \textsc{Doctor} is to fully exploit all available information contained in the soft-probabilities of the predictions (not only their maximum). 
\end{itemize}

\subsection{Related works} 
Recent works have shown that the accuracy of a classifier and its ability to output soft-predictions that represent the true posteriors estimate can be totally disjointed~\cite{GuoPSW2017ICML,KuleshovE2016CoRR,KuleshovL2015NIPS}.  Furthermore, models often tend to be overconfident about their decision even when their predictions  fail~\cite{HeinAB2018CVPR,KristiadiHH2020ICML}. This motivates a novel research area that strives to develop methods to assess when decisions made by classifiers should or should not be trusted. 
Although the detection of OOD samples is a different (domain) problem, it is naturally expected that samples from a distribution that is significantly different from the training one cannot be correctly classified. In~\cite{LiangLS2018ICLR}, the authors propose a method which increases the peakiness of the softmax output by perturbing the input samples and applying temperature scaling~\cite{GuoPSW2017ICML,HintonVD2015NIPSWS,Platt1999ALMC} to the classifier logits in order to better detect in-distribution samples. It is worth noticing that this method requires additional information on the internal structure of the latent code of the model. A very different approach~\cite{HsuSJK2020IEEEX,LeeLLS2018NeurIPS} tackles OOD detection by using the \textit{Mahalanobis distance}. Although this approach appears to be more powerful, it also  requires additional samples to learn the mean by class and the covariance matrix of the in-distribution. In~\cite{DeVriesWT2018CoRR}, classifiers are trained to output calibrated confidence estimates that are used to perform OOD detection. A  related line of research is concerned with the problem of \textit{selective predictions}  (aka~\textit{reject options}) in deep neural networks. The main motivation for selective prediction is reducing the error rate by abstaining from prediction when in doubt, while keeping the number of correctly classified samples as high as possible~\cite{GangradeAISTATS2021,GeifmanE2017NIPS,GeifmanY2019ICML}.  The idea is to combine classifiers with \textit{rejection functions} by observing the classifiers' output, without using any supervision, to decide whether to accept or to reject the classification outcome. In~\cite{GeifmanE2017NIPS}, the authors introduce \textit{softmax response}, a rejection function which compares the maximum soft-probability to a pre-determined threshold to decide whether to accept or reject the class prediction given by the model.

\section{Main Definitions and Preliminaries}
\label{sec:preliminaries}
\subsection{Basic definitions} 

We start by introducing some definitions and background; then, we  describe our statistical model and some useful properties about the underlying detection problem. Let $\mathcal{X}\subseteq \mathbb{R}^d$ be the (possibly continuous) feature space and let  $\mathcal{Y}=\{1,\dots,C\}$ denote the concept of the label space related to some task of interest. Moreover, let $p_{{X}Y}$ be the underlying (unknown) probability density function (pdf) over $\mathcal{X}\times\mathcal{Y}$. Let $\mathcal{D}_{n} = \big\{( \mathbf{x}_1, y_1),\dots,( \mathbf{x}_n, y_n)  \big\}\sim p_{{X}Y}$ be a random realization of $n$ i.i.d. samples according to $p_{{X}Y}$  denoting the \textit{training set}, where $\mathbf{x}_i \in \mathcal{X}$ is the input (feature), $y_i \in \mathcal{Y} $ is the output class among $C$ possible classes and $n$ denotes the size of the training set. A predictor ${f}_{\mathcal{D}_{n}}: \mathcal{X}\rightarrow\mathcal{Y}$  uses the inferred model $P_{\widehat{Y}|{X}}\equiv P_{\widehat{Y}|{X}}(y|\mathbf{x};\mathcal{D}_{n})$ based on the training set, 
$$
f_{\mathcal{D}_{n}}(\mathbf{x}) \equiv  f_{{n}}(\mathbf{x};\mathcal{D}_{n})  \myeq  \arg\max_{y\in\mathcal{Y}}~P_{\widehat{Y}|{X}}(y|\mathbf{x};\mathcal{D}_{n}),
$$
and tries to approximate the optimal  (Bayes) decision rule 
$
f^\star(\mathbf{x})  \myeq   \arg\max\limits_{y\in\mathcal{Y}}~P_{{Y}|{X}}(y|\mathbf{x}).
$
Notice that $P_{\widehat{Y}|{X}}$ can be interpreted as the prediction of the class (label) posterior probability given a sample (e.g., $P_{\widehat{Y}|X}(y~|\mathbf{x})\equiv \fun{softmax}(\mathbf{\mathbf{x}})_y$), while $P_{Y|{X}}$ is the true (unknown)  probability. 
In several practical scenarios $P_{\widehat{Y}|{X}}$ does not perfectly match $P_{{Y}|{X}}$ and still $f_{\mathcal{D}_{n}} \approx f^\star$ (cf. \cite{GuoPSW2017ICML}).

\subsection{Error variable} 

Let $E(\mathbf{x}) \myeq    \mathds{1}\left [Y \neq f_{\mathcal{D}_{n}}(\mathbf{x}) \right]$ denote the error variable for a given $\mathbf{x}\in\mathcal{X}$ corresponding to  $f_{\mathcal{D}_{n}}$, i.e., where we denote with $\mathds{1}[\mathcal{E}]$ the indicator vector which outputs $1$ if the event $\mathcal{E}$ is true and $0$ otherwise. Similarly, we can define the self-error variable $\widehat{E}(\mathbf{x}) \myeq    \mathds{1}\big[\widehat{Y} \neq f_{\mathcal{D}_{n}}(\mathbf{x}) \big]$ also corresponding to the inferred predictor $f_{\mathcal{D}_{n}}$ but based on the prediction model $P_{\widehat{Y}|{X}}$ of the class posterior probability. Notice that $\widehat{E}(\mathbf{x})$ is observable since the underlying distribution is  known. However, ${E}(\mathbf{x})$ cannot be observed and in general these binary variables  do not coincide.  

At this stage, it is convenient to introduce the notions of \textit{probability of classification  error} for a given $\mathbf{x}\in\mathcal{X}$ w.r.t. both the true class posterior and the predicted probabilities: 
\begingroup
\allowdisplaybreaks
\begin{align}
\fun{Pe}(\mathbf{x}) & \myeq   \mathbb{E}\left[ {E}(\mathbf{x}) | \mathbf{x} \right]   
={} 1 - P_{{Y}|X}\left( f_{\mathcal{D}_{n}}(\mathbf{x})  |\mathbf{x}\right), \label{eq:pe}\\
\widehat{\fun{Pe}}(\mathbf{x}) & \myeq  \mathbb{E}\left[ \widehat{E}(\mathbf{x}) | \mathbf{x} \right]  ={} 1 - P_{\widehat{Y}|X}\left( f_{\mathcal{D}_{n}}(\mathbf{x})  |\mathbf{x}\right).  \label{eq:ph}
 \end{align}
\endgroup
Notice that $\widehat{\fun{Pe}}(\mathbf{x})$ represents the probability of misclassification of the sample $\mathbf{x}$ with respect to the softmax probability $P_{\widehat{Y}|X}$, which can be interpreted as the model's approximation of nature  $P_{Y|X}$. Such approximation is close when the model is well-calibrated.
Obviously,  $ \fun{Pe}^\star(\mathbf{x})  \leq \fun{Pe}(\mathbf{x})$ for all $\mathbf{x}\in\mathcal{X}$, where $\fun{Pe}^\star(\mathbf{x})$  corresponds to the minimum error of the  Bayes classifier:  ${\fun{Pe}^\star(\mathbf{x}) =  1- P_{{Y}|X}\left( f^\star(\mathbf{x})  |\mathbf{x}\right)}$. It is worth mentioning that, by averaging  \eqref{eq:pe}  over the data distribution, we obtain the error rate of the classifier $f_{\mathcal{D}_{n}}$. Although $\widehat{\fun{Pe}}(\mathbf{x}) $ provides a valuable candidate to infer the unknown error variable ${E}(\mathbf{x})$, it is easy to check that  \begin{align}
 \max\big\{ \fun{Pe}(\mathbf{x}) , \widehat{\fun{Pe}}(\mathbf{x}) \big\} - \Pr\big(\widehat{Y} =  {Y} | \mathbf{x}\big)\leq\Pr\big\{ \widehat{E}(\mathbf{x})  \neq   {E}(\mathbf{x}) | \mathbf{x} \big\} & \leq  \Pr\big(\widehat{Y} \neq  {Y} | \mathbf{x}\big),\label{eq-misssing-1}
 \end{align}
which in particular implies that the error incurred in using $\widehat{E}(\mathbf{x})$ to predict  ${E}(\mathbf{x})$ is lower bounded by the classification error per sample \eqref{eq:pe}. The proofs are in Supplementary material (\Cref{appendix:proof-eq}).

In this paper, we aim at identifying a discriminator capable of distinguishing between inputs $\mathbf{x}$ for which we can trust the predictions of the classifier $f_{\mathcal{D}_{n}}(\mathbf{x})$ (i.e., $E(\mathbf{x})=0$) and those for which we should  not  trust predictions (i.e., $E(\mathbf{x})=1$). In the next section, we will show  that the function $\fun{Pe}(\mathbf{x}): \mathcal{X} \mapsto [0,1]$ plays a central role in the characterization of the optimal discriminator. However, $\fun{Pe}(\mathbf{x})$ is not available in practical scenarios and the direct estimation (e.g., based on pairs of inputs and labels) of the true class posterior probability $ P_{{Y}|{X}}$ cannot be performed.  Notice that it is not possible to sample the conditional pdf $P_{{Y}|{X}}$ for each input    $\mathbf{x}\in\mathcal{X}$.  As a matter of fact, it is well-known that the application of direct methods for this estimation will 
lead to ill-posed problems, as 
shown in \cite{pmlr-v128-vapnik20a}. 



\subsection{Statistical model for detection} 

Given a data sample $\mathbf{x}\in\mathcal{X}$ and an unobserved random label $y\in\mathcal{Y}$ drawn from  the unknown distribution $p_{XY}$, we wish to predict the realization of the unobserved error variable $E\myeq  \mathds{1} [Y \neq f_{\mathcal{D}_{n}}(\mathbf{X})]$. To this  end, we will model the data distribution as a mixture pdfs, 
\begin{equation*}
p_{XY}(\mathbf{x}, y)  \equiv  P_E(1) p_{XY|E}(\mathbf{x},y | 1) 
+ P_E(0) p_{XY|E}(\mathbf{x},y | 0),\label{eq:join}
\end{equation*}
where $p_{XY|E}(\mathbf{x},y | 1)$ denotes the pdf truncated to the error event $\{E=1\}$ (i.e., the hard decision fails) and $p_{XY|E}(\mathbf{x},y | 0)$ is the pdf truncated to the success  event  $\{E=0\}$ (i.e., the hard decision succeeds).  By taking the marginal of $p_{XY}$ over the labels, we obtain:
$
p_{X}(\mathbf{x}) =  P_E(1) p_{X|E}(\mathbf{x}| 1) 
+ P_E(0) p_{X|E}(\mathbf{x}| 0).  
$
First, observe that the problem at hand is to infer $E$ from $(\mathbf{x}, P_{\widehat{Y}|{X}} )$ since $Y$ is not observed. Second, we further emphasize that in the present framework we  assume that there are no available (extra)  samples for training a discriminator to distinguish between $p_{X|E}(\mathbf{x}| 0)$ and $p_{X|E}(\mathbf{x}| 1)$. It is worth mentioning that a well-trained classifier would imply $ P_E(1) \ll  P_E(0)$, since in that case we should have very few classification errors. However, this also implies that it would be very unlikely to have enough samples available to train a good enough  discriminator.

\section{Performance Metrics and Discriminators}
\label{sec:performanceandmetrics}
\subsection{Performance metrics and optimal discriminator}
\label{subsec:performance}
We aim to distinguish between samples for which the predictions cannot be trusted and  samples for which predictions should be trusted. We first state the optimal rejection region, given by \eqref{eq:opt}, where we suppose the existence of an oracle who knows all the involved probability distributions.
\begin{definition}[Most powerful discriminator]\label{def-decision-region}
For any ${0 < \gamma < \infty}$, define the decision region: 
\begin{align}
\mathcal{A}(\gamma) \myeq \left\{ \mathbf{x} \in\mathcal{X}: {p_{X|E}(\mathbf{x} | 1)}> \gamma\cdot {p_{X|E}(\mathbf{x} | 0)}\right\}.\label{eq:opt}
\end{align}
The most powerful (Oracle) discriminator at threshold $\gamma$ is defined by setting $D(\mathbf{x},\gamma) = 1$  for all $\mathbf{x} \in  \mathcal{A}(\gamma)$  for which the prediction is rejected (i.e., $\widehat{E} = 1$) and otherwise $D(\mathbf{x},\gamma) = 0$ for all  $ \mathbf{x} \notin   \mathcal{A}(\gamma)$ for which the prediction is accepted.
\end{definition}
In~\cref{prop:1}, we establish the characterization of the fundamental performance of the most powerful (Oracle) discriminator by providing a lower bound on the error achieved by any discriminator and show that this bound is achievable by setting $\gamma=1$. Furthermore, we connect this result to the Bayesian error rate of this optimal discriminator.

\begin{proposition}[Performance of the discriminator]
\label{prop:1}
	For any given decision region $\mathcal{A}\subset{\mathcal{X}}$, let 
	\begin{align}
        \epsilon_{0}(\mathcal{A})  \myeq \int_{\mathcal{A}}p_{X|E}(\mathbf{x}  | 0)d\mathbf{x},  \;   \text{~~~~and~~~~}
                {\epsilon}_{1}( \mathcal{A}^c )  \myeq \int_{ \mathcal{A}^c}p_{X|E}(\mathbf{x}  | 1)d\mathbf{x}  \;,\label{eq-Type1}
	\end{align}
be the Type-I (rejection of the class prediction of an input  $\mathbf{x} $ that would be correctly classified) and Type-II (acceptance of the class prediction of an input $\mathbf{x} $ that would not be correctly classified) error probability, respectively.
Then, 
	\begin{align}
		{\epsilon}_{0}(\mathcal{A}) + {\epsilon}_{1}(\mathcal{A}^c) &\geq  1- \left\Vert p_{X|E=1} - p_{X|E=0} \right\Vert_\mathrm{TV}\\
		& =1-  \frac12 \int_{\mathcal{X}} |p_{X|E=1}(\mathbf{x})- p_{X|E=0}(\mathbf{x})| d\mathbf{x}. 
	    \label{eq:prop1}
	\end{align}
Equality is achieved by choosing the optimal decision region  $\mathcal{A}^\star \equiv \mathcal{A}(1)$ in Definition~\ref{def-decision-region}. If the hypotheses are equally distributed, the minimum Bayesian error satisfies:  
	\begin{equation}
 2 \Pr\left\{D(\mathbf{X}) \neq E(\mathbf{X}) \right\} 
 	\geq   1 -  \left\Vert p_{X|E=1} - p_{X|E=0} \right\Vert_\mathrm{TV}.
 	    \label{eq:prop1:2}
	\end{equation}
Equality is achieved by using the optimal decision region.
\end{proposition}
Expressions~\eqref{eq:prop1} and \eqref{eq:prop1:2} provide lower bounds for the total error of an arbitrary discriminator.  The proof of this proposition is relegated to the Supplementary material (\Cref{appendix:proof1}). Using Bayes we can rewrite \eqref{eq:opt} via the posteriors as:
\begin{equation}
	\mathcal{A}(\gamma) 
	= \left\{\mathbf{x} \in\mathcal{X}:  {P_{E|X}( 1|\mathbf{x} )} {P_{E}(0)} > \gamma \cdot \left(1-P_{E|X}(1|\mathbf{x} )  \right)   {P_{E}(1)} \right\}.\label{eq:opt2}
\end{equation}
From \eqref{eq:opt2}, it is easy to check that 
$
	P_{E|X}(1|\mathbf{x}) = 1 - P_{{Y}|X}\left( f_{\mathcal{D}_{n}}(\mathbf{x})  |\mathbf{x}\right) =  \fun{Pe}(\mathbf{x} ), 
$
and hence, the decision region $\mathcal{A}(\gamma)$ can be reformulated as:
\begin{equation}\label{eq:opt'}
\mathcal{A}(\gamma^\prime) =\left\{\mathbf{x} \in\mathcal{X}:\frac{\fun{Pe}(\mathbf{x} )}{1-\fun{Pe}(\mathbf{x} )}> \gamma^\prime \right\}= \left\{\mathbf{x} \in\mathcal{X}:  \fun{Pe}(\mathbf{x} ) > \frac{\gamma^\prime}{(\gamma^\prime + 1)} \right\},
\end{equation}
where $\gamma^\prime  \myeq \gamma \cdot \frac{P_{E}(1)}{P_{E}(0)}$ and $0 < \gamma^\prime < \infty$. According to  \eqref{eq:opt'} and Proposition~\eqref{prop:1}, the optimal discriminator is given by  $D^\star (\mathbf{x},\gamma^\prime) = 1$, whenever $\mathbf{x}\in \mathcal{A}(\gamma^\prime)$, and $D^\star (\mathbf{x},\gamma^\prime) = 0$, otherwise.  The main difficulty arises here since the error probability  function of an input:    $\mathbf{x} \mapsto \fun{Pe}(\mathbf{x} )$ is not known and in general cannot be learned from training samples.  

\subsection{\textsc{Doctor} discriminator}
We start by deriving an approximation to the unknown function $\mathbf{x} \mapsto \fun{Pe}(\mathbf{x} )$  which can be used to devise the decision region in expression  \eqref{eq:opt'}. First, we state the following: 

\begin{proposition}\label{prop:2}
Let $\widehat{\fun{g}}(\mathbf{x})$ be defined by 
\begin{equation}\label{eq-g_hat}
1- \widehat{\fun{g}}(\mathbf{x})  \myeq    \underset{y\in\mathcal{Y}}{\sum} P_{\widehat{Y}|X}(y|\mathbf{x}) \Pr\left (\widehat{Y} \neq y |\mathbf{x}\right)
= 1 - \underset{y\in\mathcal{Y}}{\sum}P_{\widehat{Y}|X}^2(y|\mathbf{x}), 
\end{equation}
for each $\mathbf{x} \in\mathcal{X}$, which indicates the probability of incorrectly classifying a feature $\mathbf{x}$ if it was randomly labeled according to the model distribution $P_{\widehat{Y}|X}$ trained based on the dataset. Then, 
\begingroup
\allowdisplaybreaks
\begin{equation}
 (1 - \sqrt{  \widehat{\fun{g}}(\mathbf{x}) }) - \Delta(\mathbf{x})  \leq \pe(\mathbf{x} )   \leq  (1 - \widehat{\fun{g}}(\mathbf{x})) +  \Delta(\mathbf{x}),\label{bound:h1} \\
 \end{equation}
\endgroup
\end{proposition}
where $\Delta (\mathbf{x}) \myeq {2\sqrt{2~\textrm{KL}\big(P_{Y|X} (\cdot| \mathbf{x} ) \|P_{\widehat{Y}|X}(\cdot| \mathbf{x} ) \big)}}$ and denotes the Kullback–Leibler (KL) divergence (further details are provided in  Supplementary material \Cref{appendix:proof2}). 

\subsection{Discussion} 

It is worth emphasizing that expressions in \eqref{bound:h1} provide bounds to the unknown function $\mathbf{x} \mapsto \fun{Pe}(\mathbf{x} )$  using a known statistics  $\mathbf{x} \mapsto 1- \widehat{\fun{g}}(\mathbf{x})$, which is based on the soft-probability of the predictor. On the other hand, 
$
 0\leq  \widehat{\fun{g}}(\mathbf{x})\leq \sqrt{  \widehat{\fun{g}}(\mathbf{x}) }\leq 1, \, \textrm{for all $\mathbf{x} \in\mathcal{X}$, }
$
which simply follows using the subadditive of the function $t \mapsto \sqrt{t}$ and the definition of $\widehat{\fun{g}}(\mathbf{x})$. By Markov's inequality, 
\begin{equation}
\Pr \big(\Delta (\mathbf{X})\geq  \varepsilon{(\eta)} \big)
 \leq  \eta \ \  \textrm{ with } \ \  \varepsilon(\eta) = 2 \sqrt{{2\mathbb{E}_{\mathbf{X}Y}\big[-\log P_{\widehat{Y}|X}(Y| \mathbf{X})  \big]}/{\eta}}, \label{eq-KL--CE}
\end{equation}
for any $\eta>0$, where $\mathbb{E}_{\mathbf{X}Y}\big[-\log P_{\widehat{Y}|X}(Y| \mathbf{X})  \big]$ in \eqref{eq-KL--CE} is the cross-entropy risk. The  latter is expected to be small provided that the model generalizes well. Thus, $\varepsilon{(\eta)}$ can be expected to be small for a desired confidence $\eta>0$.  Interestingly,  \eqref{eq-g_hat} turns out to be related to the uncertainty of the classifier via the quadratic R\'enyi entropy~\cite{erven2014}: $
   -  \log_2  \big(  \widehat{\fun{g}}(\mathbf{x}) \big)  =  2 H_2(\widehat{Y}| \mathbf{x})  \leq 2 H(\widehat{Y}| \mathbf{x}) , 
$
where the latter is the Shannon entropy, i.e.,  the self-uncertainty of the classifier. 

\subsection{From the theory to a practical discriminator} 
Our previous discussion  suggests that $\pehat(\mathbf{x})$ in \eqref{eq:ph} may be a valuable candidate to approximate $\pe(\mathbf{x})$ in the definition of the optimal discriminator \eqref{eq:opt'}. On the other hand,~\cref{prop:2} suggests that $1-\widehat{\fun{g}}(\mathbf{x})$ can also be a valuable candidate yielding another discriminator. These discriminators, referred to as \textsc{Doctor}, are introduced below. 

\begin{definition}[\textsc{Doctor}]\label{def:doctor}
For any $0<\gamma <\infty$ and $\mathbf{x} \in\mathcal{X}$, define the following discriminators:  
\begin{align}
D_\alpha  (\mathbf{x}, \gamma )  \myeq   \mathds{1} \left[ \ghat(\mathbf{x}) > \gamma \cdot \widehat{\fun{g}}(\mathbf{x}) \right],  & & 
D_\beta  (\mathbf{x}, \gamma )  \myeq 
\mathds{1} \left[ \pehat(\mathbf{x})   > \gamma\cdot (1-\pehat(\mathbf{x})) \right].\label{eq:d_alpha}
\end{align}
\end{definition}
Notice that because of \cref{def:doctor} and  \eqref{eq-g_hat}, $D_\alpha(\mathbf{x}, \gamma) = \mathds{1}[1 - \sum_{y\in\mathcal{Y}}~\fun{softmax}^2(\mathbf{x})_y > \gamma \cdot\sum_{y\in\mathcal{Y}}~\fun{softmax}^2(\mathbf{x})_y]$; similarly because of \cref{def:doctor} and eq. \eqref{eq:ph}, $D_\beta(\mathbf{x}, \gamma) = \mathds{1}[1 - \max_{y\in\mathcal{Y}}~\fun{softmax}(\mathbf{x})_y > \gamma \cdot\max_{y\in\mathcal{Y}}~\fun{softmax}(\mathbf{x})_y].$
The performance of these discriminators will be investigated and compared to state-of-the-art methods in the next section. 
In the supplementary material (\Cref{appendix:toy}), we illustrate how \textsc{Doctor} and the optimal discriminator (\cref{def-decision-region}) work on a synthetic data model that is a mixture of two spherical Gaussians with one component per class.

\setcounter{footnote}{0} 

\section{Experimental Results}
\label{sec:experiments}
In this section we present a collection of experimental results to investigate the effectiveness of \textsc{Doctor}, by applying it to several benchmark datasets. We provide publicly available code\footnote{\url{https://github.com/doctor-public-submission/DOCTOR/}} to reproduce our results, and we give further details on the environment, the parameter setting and the experimental setup in the Supplementary material (\Cref{appendix:experimental_setup}).
We propose a comparison with state-of-the-art methods using similar information. Though we are not concerned with the OOD detection problem, we are confident it is appropriate to compare \textsc{Doctor} to methods which use soft-probabilities or at most the output of the latent code, e.g.,  ODIN~\cite{LiangLS2018ICLR}, softmax response (SR)~\cite{GeifmanE2017NIPS} and Mahalanobis distance (MHLNB)~\cite{LeeLLS2018NeurIPS}. 
Since we are focusing on misclassification detection, it is expected that OOD samples should be also detected as classification errors. 

\textbf{Totally Black Box (TBB) and Partially Black Box (PBB).} We address two different scenarios with respect to the available information about the network.  In the TBB  only the output of the last layer of the network is available, hence gradient-propagation  to perform input pre-processing is not allowed. In the PBB we allow method-specific inputs perturbations. When considering \textsc{Doctor} in PBB, for each testing sample $\mathbf{x}$, we calculate the pre-processed sample $\mathbf{\widetilde{x}}$ by adding a small perturbation:
\begin{align*}\label{eq:perturb_alpha}
    \widetilde{\mathbf{x}}^\alpha &= \mathbf{x} - \epsilon\times \text{sign}\left[-\nabla_{\mathbf{x}}\log\left (\frac{\ghat(\mathbf{x})}{\gh(\mathbf{x})}\right)\right], 
    \text{~and~ }
    \widetilde{\mathbf{x}}^\beta 
    = \mathbf{x} - \epsilon\times\text{sign}\left[-\nabla_{\mathbf{x}}\log\left(\frac{\pehat(\mathbf{x})}{1 - \pehat(\mathbf{x})}\right)\right].
\end{align*}
We will write directly $\widetilde{\mathbf{x}}$  when it is clear from the context which input pre-processing we are referring to.
In Supplementary material (\Cref{appendix:input_pre_processing}) we further analyze the equations above.
When ODIN or MHLNB are used, we pre-process the inputs as in~\cite{LiangLS2018ICLR} and in~\cite{LeeLLS2018NeurIPS}, respectively.

\subsection{Review of related methods}
\label{sub:relatedexperiments}
\textbf{PBB.} We compare \textsc{Doctor} (using input pre-processing and temperature scaling) with ODIN and MHLNB.
ODIN~\cite{LiangLS2018ICLR}  comprises 
temperature scaling and input pre-processing via perturbation. Temperature scaling is applied
to its scoring function, which has $f_i(\mathbf{\widetilde{x}})$ for the logit of the $i$-th class. Formally, given an input sample $\mathbf{x}$:
\begin{equation*}
\textrm{SODIN}(\mathbf{\widetilde{x}})= \max_{i=[1:C]} \frac{\exp(f_i(\mathbf{\widetilde{x}}) /T) }{\sum_{j=1}^C \exp(f_j(\mathbf{\widetilde{x}})/T )}\text{,}~~		\fun{ODIN}(\mathbf{\widetilde{x}}; \delta, T, \epsilon)=
		\begin{cases}
			\textrm{out}, & \text{if }\textrm{SODIN}(\mathbf{\widetilde{x}}) \leq \delta \\
			\textrm{in},  & \text{if }\textrm{SODIN}(\mathbf{\widetilde{x}}) > \delta,
		\end{cases}   
\label{eq:sodin}
\end{equation*}
where $\mathbf{\widetilde{x}}$ represents a magnitude $\epsilon$ perturbation of the original $\mathbf{x}$; $T$ is the temperature scaling parameter; $\delta \in [0,1]$ is the threshold value; $in$ indicates the acceptance decision while $out$ indicates the rejection decision. Notice, however, $\gamma$ in \textsc{Doctor} and $\delta$ in ODIN, respectively,  are defined over two different domains: if $\delta$ denotes a probability, $\gamma$ is a ratio between probabilities. 
Although ODIN originally required tuning the hyper-parameter $T$ with out-of-distribution data, it was also shown that a large value for $T$ is generally desirable, suggesting that this gain is achieved at $1000$. Anyway, in this framework, we notice an improvement of ODIN in performance for low values of $T$. Thus we report the best results obtained by ODIN considering the range of hyper-parameters values tested also for \textsc{Doctor} (cf.~\cref{subsec:exp_res}). 
ENERGY~\cite{LiuNIPS2020} comprises the denominator of the softmax activation: 
\begin{equation*}
\textrm{ES}(\mathbf{x})= -T \cdot\log \sum_{j=1}^C \exp(f_j(\mathbf{x})/T )\text{,}~~		\fun{ENERGY}(\mathbf{x}; \xi, T)=
		\begin{cases}
			\textrm{out}, & \text{if } -\textrm{ES}(\mathbf{x}) \leq \xi \\
			\textrm{in},  & \text{if } -\textrm{ES}(\mathbf{x}) > \xi,
		\end{cases}   
\label{eq:energy}
\end{equation*}
where $\xi\in\mathbb{R}$ is the threshold value.
Unlike all the methods considered in this paper, MHLNB~\cite{LeeLLS2018NeurIPS} requires the knowledge of the training set $\mathcal{D}_n$ which the pre-trained network was trained on to compute its \textit{empirical class mean }${\widehat{\mu}_c}$ for each class $c$ and its 
\textit{empirical covariance }$\widehat{\Sigma}$: 
$$ 
{\widehat{\mu}_c} = \frac{1}{n_c} \sum_{i:\, y_i=c} f(\mathbf{\widetilde{x}}_i);~
\widehat{\Sigma} = \frac{1}{n}\sum_{c\in\mathcal{Y}} \sum_{\forall i:\, y_i=c} (f(\mathbf{\widetilde{x}}_i) - \widehat{\mu}_c)(f(\mathbf{\widetilde{x}}_i) - \widehat{\mu}_c)^\top,
$$ 
where $n_c$ denotes the number of training samples with label $c$ and $f(\mathbf{\widetilde{x}})$ the logits vector. As MHLNB  directly uses the vector of logits, it does not comprise temperature scaling. 
Given an input sample $\mathbf{x}$:
\begin{equation*}
    \fun{M}(\mathbf{\widetilde{x}}) = \max\limits_{ c\in\mathcal{Y} }~-(f(\mathbf{\widetilde{x}}) - \widehat{\mu}_c)^\top \widehat{\Sigma}^{-1} (f(\mathbf{\widetilde{x}}) - \widehat{\mu}_c)\text{,}~~\fun{MHLNB}(\mathbf{\widetilde{x}}; \zeta, \epsilon)=
		\begin{cases}
			\textrm{out}, & \text{if }\textrm{M}(\mathbf{\widetilde{x}}) > \zeta \\
			\textrm{in},  & \text{if }\textrm{M}(\mathbf{\widetilde{x}}) \leq \zeta,
	    \end{cases}
\end{equation*}
as mentioned above, $\mathbf{\widetilde{x}}$ represents a magnitude $\epsilon$ perturbation of the original $\mathbf{x}$; $\zeta \in \mathbb{R}_+$ is the threshold value; $in$ indicates the acceptance decision while $out$ indicates the rejection decision. 

\textbf{TBB.} We compare \textsc{Doctor} (without input pre-processing and temperature scaling) with MHLNB (without input pre-processing and with the softmax output layer in place of the logits) and SR.
Although both \textnormal{\textsc{Doctor}} and \textnormal{SR} have access to the softmax output of the predictor, a fundamental difference is that, while the former utilizes the softmax output in its entirety, the latter only uses the maximum value, therefore discarding potentially useful information. As it will be shown, this leads to better results for \textsc{Doctor} on several datasets (see~\cref{tab:best_aurocs}). 
We emphasize that, by setting $T=1$ and $\epsilon=0$, ODIN reduces to softmax response~\cite{GeifmanE2017NIPS} since $\textrm{SR}(\mathbf{x}) \equiv  \textrm{SODIN}(\mathbf{x})$.

\subsection{Detection of misclassification errors, experimental setup and evaluation metrics}
\label{sec:experiments_setup}

\begin{figure}[!htb]
	\centering
	\begin{subfigure}[b]{0.23\textwidth}
		\centering
		\includegraphics[width=\textwidth]{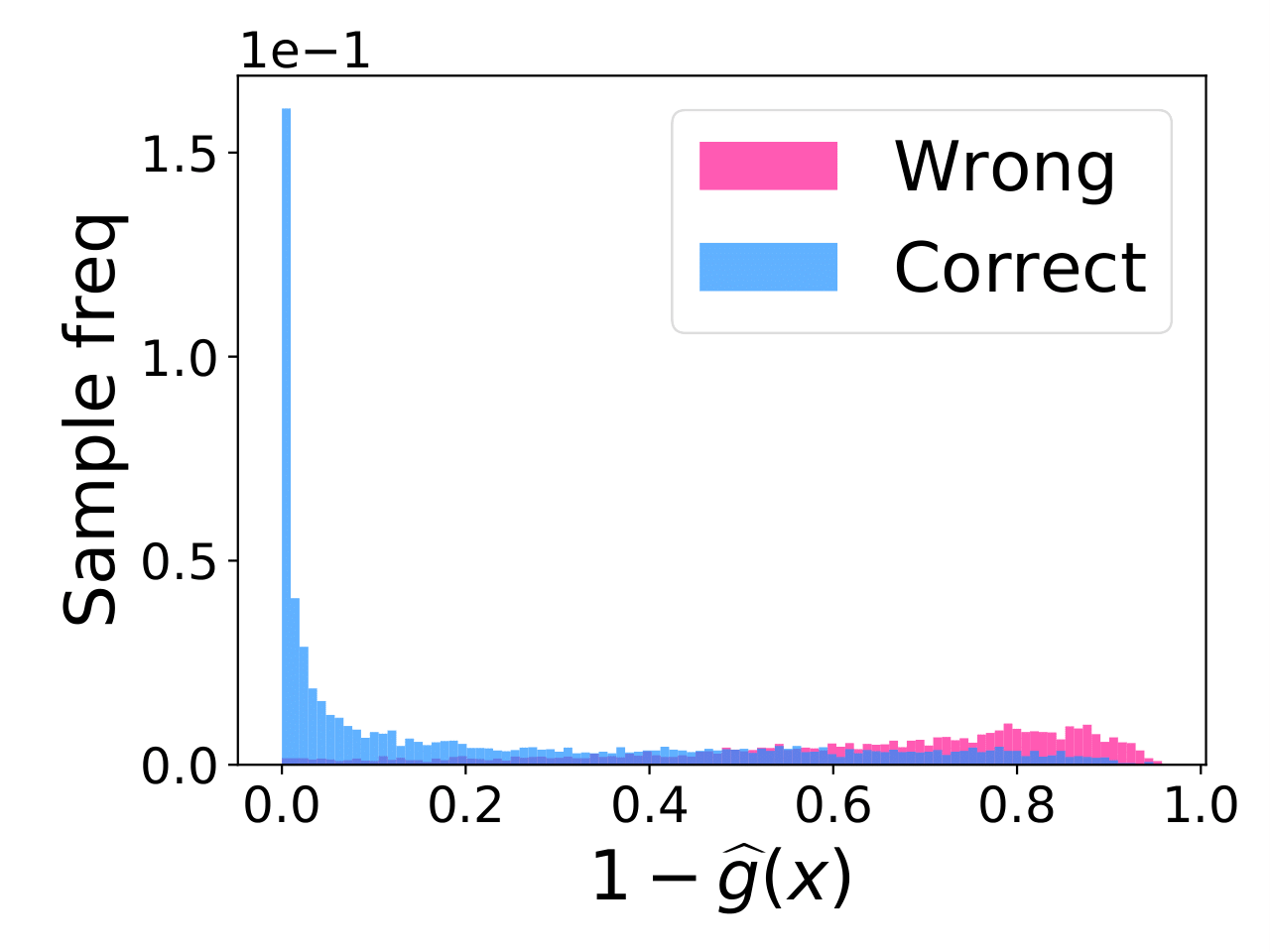}
		\vspace{-1.5\baselineskip}
		\caption{$D_\alpha$ - TBB}
		\label{fig:hist_alpha}
	\end{subfigure}
	\hfill
		\begin{subfigure}[b]{0.23\textwidth}
		\centering
		\includegraphics[width=\textwidth]{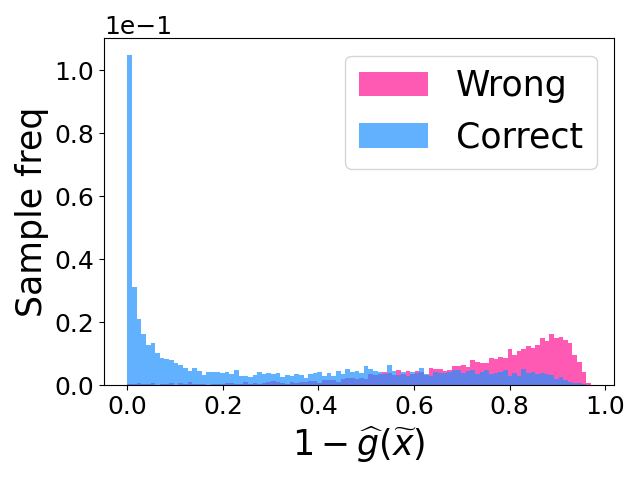}
		\vspace{-1.5\baselineskip}
		\caption{$D_\alpha$ - PBB}
		\label{fig:hist_alpha_perturbed}
	\end{subfigure}
	\hfill
	\begin{subfigure}[b]{0.23\textwidth}
		\centering
		\includegraphics[width=\textwidth]{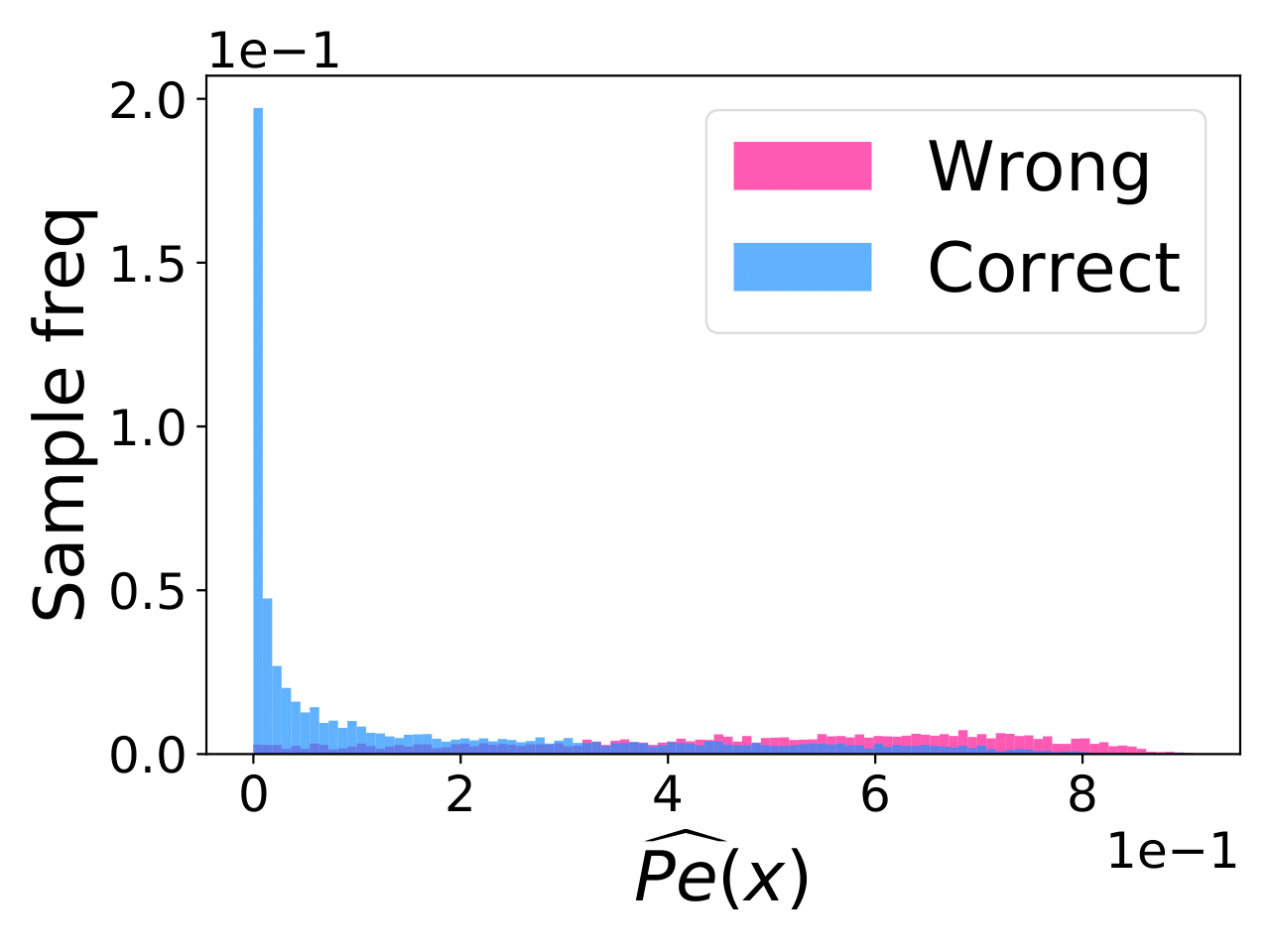}
		\vspace{-1.5\baselineskip}
		\caption{$D_\beta$ - TBB}
		\label{fig:hist_beta}
	\end{subfigure}
	\hfill
	\begin{subfigure}[b]{0.23\textwidth}
		\centering
		\includegraphics[width=\textwidth]{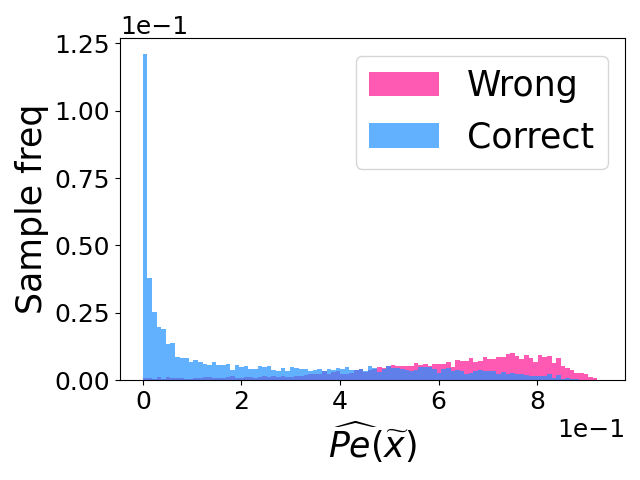}
		\vspace{-1.5\baselineskip}
		\caption{$D_\beta$ - PBB}
		\label{fig:hist_beta_perurbed}
	\end{subfigure}
	\hfill
	\begin{subfigure}[b]{0.23\textwidth}
	    \centering
	    \includegraphics[width=\textwidth]{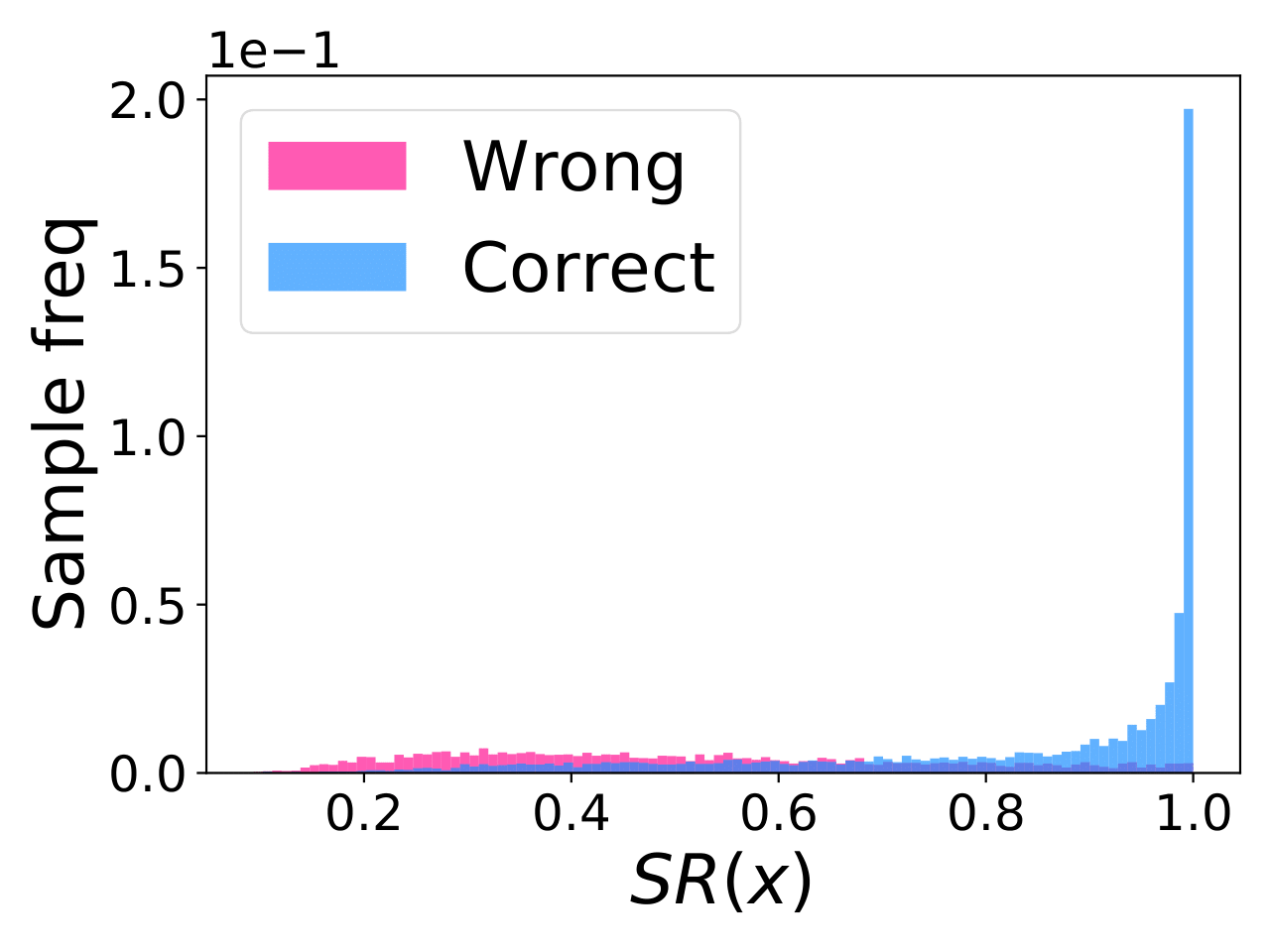}
	    \vspace{-1.5\baselineskip}
	    \caption{SR - TBB}
	    \label{fig:hist_sr}
	\end{subfigure}
    \hfill
	\begin{subfigure}[b]{0.23\textwidth}
		\centering
		\includegraphics[width=\textwidth]{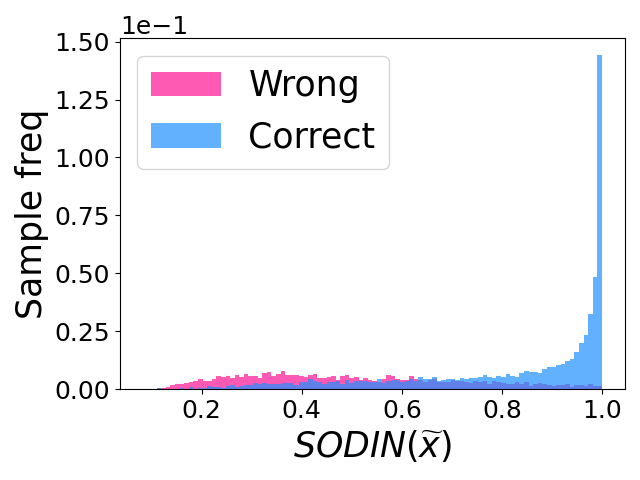}
		\vspace{-1.5\baselineskip}
		\caption{ODIN - PBB}
		\label{fig:hist_odin}
	\end{subfigure}
	\hfill
	\begin{subfigure}[b]{0.23\textwidth}
	    \centering
	    \includegraphics[width=\textwidth]{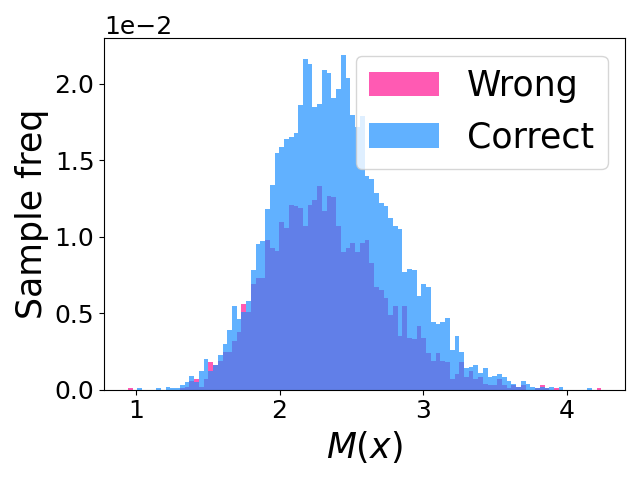}
	    \vspace{-1.5\baselineskip}
	    \caption{MHLNB - TBB}
	    \label{fig:hist_mhlnb}
	\end{subfigure}
	\hfill
	\begin{subfigure}[b]{0.23\textwidth}
	    \centering
	    \includegraphics[width=\textwidth]{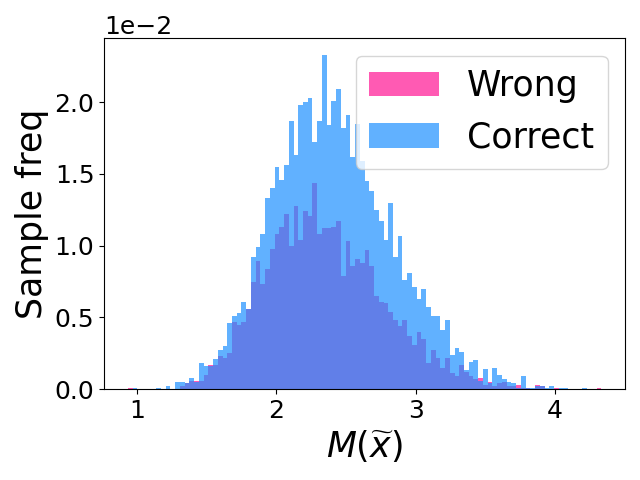}
	    \vspace{-1.5\baselineskip}
	    \caption{MHLNB - PBB}
	    \label{fig:hist_mhlnb_perturbed}
	\end{subfigure}
	\caption{DOCTOR, ODIN, SR and MHLNB to split data samples in TinyImageNet both under TBB and PBB: (a) - (b) show the results for expressions~\eqref{eq:ph}; (c) - (d) show the results for~\eqref{eq-g_hat}; (e) shows the results for SR; (f)  shows the results for ODIN; (g) - (h) show the results for MHLNB. Histograms for wrongly classified samples (red) and  correctly classified samples (blue).}
	\label{fig:hist}
\end{figure}

Before digging into the detailed discussion of our numerical results, we present an empirical analysis of the behavior of \textsc{Doctor}, ODIN, SR and MHLNB when faced with the task of choosing whether to accept or reject the prediction of a given classifier for a certain sample. 
In~\Cref{fig:hist}, we propose a graphical interpretation of the discrimination performance, considering the labeled samples in the dataset TinyImageNet and the ResNet network as the classifier.  We separate correctly and incorrectly classified samples according to their true labels in blue and in red, respectively.  We remind that the label information is \emph{not} necessary for the discriminators to define acceptance and rejection regions.
Then, for each sample we compute the corresponding discriminators’ output. These values are binned and reported on the horizontal axis of~\Cref{fig:hist_alpha} and~\Cref{fig:hist_alpha_perturbed} for $D_\alpha$,~\Cref{fig:hist_beta} and~\Cref{fig:hist_beta_perurbed} for $D_\beta$,~\Cref{fig:hist_sr} for SR, ~\Cref{fig:hist_odin} for ODIN,~\Cref{fig:hist_mhlnb} and~\Cref{fig:hist_mhlnb_perturbed} for MHLNB. In each each plot, and according to the corresponding discriminator, the bins' heights represent the frequency of the samples whose value falls within that bin. The intuition is that, if moving along the horizontal axis it is possible to pick a threshold value such that, w.r.t. this value, blue bars are on one side of the plot and red bars on the other, this threshold would correspond to the optimal discriminator, i.e. the discriminator that chooses the optimal acceptance and rejection regions. 

In~\Cref{fig:hist_mhlnb} through~\Cref{fig:hist_mhlnb_perturbed}, we observe that, for MHLNB, no matter how well we choose the threshold value, it is hard to fully separate red and blue bars both in TBB and PBB, i.e. the discriminator fails at defining acceptance and rejection regions so that all the hits can be assigned to the first one and all the mis-classification to the second one.
The samples distribution for SR and ODIN in~\Cref{fig:hist_sr} and~\Cref{fig:hist_odin}, respectively, does not look significantly different from the one related to $D_\alpha$ and $D_\beta$ in TBB (\Cref{eq:d_alpha}).
However, the discrimination between samples becomes evident in PBB. This is shown in ~\Cref{fig:hist_beta_perurbed} for $D_\beta$ (eq. \eqref{eq:d_alpha}) and even more in~\Cref{fig:hist_alpha_perturbed} for $D_\alpha$ (eq. \eqref{eq:d_alpha}) where, quite clearly, rightly classified samples are clustered on the left-end side of the plot and incorrectly classified samples tend to cluster on the right-end side. This intuition is supported by the results in~\Cref{tab:best_aurocs}. 

\textbf{Datasets and pre-trained networks.}
We run experiments on both image and textual datasets. We use CIFAR10 and CIFAR100~\cite{Cifar}, TinyImageNet~\cite{WuZX2017Report} and SVHN~\cite{SVHN} as image datasets; IMDb~\cite{MaasEtAl2011ACL-HLT}, AmazonFashion and AmazonSoftware~\cite{NiLM2019EMNLP} as textual datasets. Note that, for all the aforementioned datasets, we consider only the test set since we rely on pre-trained models.
Along the same lines of~\cite{LiangLS2018ICLR}, we use the pre-trained {DenseNet} models \cite{HuangLW2016CoRR} for CIFAR10, CIFAR100 and SVHN. In addition, we use a pre-trained {ResNet} model \cite{HeZRS2016CVPR} for TinyImageNet, and BERT \cite{DevlinJCMLKT2019NAACLHLT,Wolfetall2020NLP} for the Amazon datasets and IMDb.
The accuracy achieved by the aforementioned networks on the test sets is showed in~\Cref{tab:best_aurocs}.
According to the invariant properties of the discriminator 
 (see Def. 2) with respect to the soft-probability of the underlying model, 
 permutations of the posterior probabilities vector, due different initialization of the models before the training, do not change the output 
 of Eq.~\eqref{eq:opt'}, 
 as it is a sum of squared values of the softmax probabilities. This variety of models/datasets characterizes the performance of the proposed method in scenarios with different accuracy levels.

\textbf{Evaluation metrics.} 
We will evaluate the performance according to  Proposition~(\ref{prop:1}) via the empirical estimates of Type-I and Type-II errors in expressions \eqref{eq-Type1}. Throughout this section, when the model's decision for a sample is correct (hit) but is rejected by the discriminator, we refer to such event as \textit{false rejection}; when the model's decision for a sample is not correct (miss) and is rejected by the discriminator, we refer to such event as \textit{true 
rejection}. Similarly, we refer to a \textit{false 
	acceptance} when a miss is not rejected and to a \textit{true acceptance} when a hit is not rejected. More specifically,  
	let $\mathcal{T}_m = \{(\mathbf{x}_1, y_1), \dots, (\mathbf{x}_m, y_m)\}\sim p_{XY}$ be the \textit{testing set}, where $\mathbf{x}_i\in\mathcal{X}$ is the input sample, $y_i\in\{1,\dots,C\}$ is the true class of $\mathbf{x}_i$,  
	and $m$ denotes the size of the testing set. 
	With $j\in\{\alpha, \beta\}$: 
	\begin{align}
	    \mathcal{FR}_j (\gamma)  & = \{(\mathbf{x}, y)\in\mathcal{T}_m : y =  f_{\mathcal{D}_n}(\mathbf{x}) ,~{ D_j }(\mathbf{x}, \gamma)=1\},\\ \mathcal{TR}_j (\gamma) & = \{(\mathbf{x}, y)\in\mathcal{T}_m : y \not=  f_{\mathcal{D}_n}(\mathbf{x}),~{D_j}(\mathbf{x},\gamma)=1\},\\  \mathcal{FA}_j(\gamma)  & = \{(\mathbf{x}, y)\in\mathcal{T}_m: y \not=  f_{\mathcal{D}_n}(\mathbf{x}),~{D_j}(\mathbf{x}, \gamma) = 0\},\\
	 \mathcal{TA}_j(\gamma) & = \left\{(\mathbf{x}, y)\in\mathcal{T}_m: y =  f_{\mathcal{D}_n}(\mathbf{x}),~{D_j}(\mathbf{x}, \gamma) = 0\right\}.
		\end{align}

We measure the performance of the test in terms of: 
\begin{itemize}
    \item \textbf{FRR} versus \textbf{TRR}. The false rejection rate (FRR) represents the probability that a hit is rejected, while the true rejection rate (TRR) is the probability that a miss is rejected.
    \item \textbf{AUROC}. The area under the \textit{Receiver Operating Characteristic curve} (ROC)~\cite{DavisGICML2006} depicts the relationship between TRR and FRR. The perfect detector corresponds to a score of $100\%$. 
    \item \textbf{FRR at 95 $\%$ TRR.} This is the probability that a hit is rejected when the TRR is at 95 $\%$.
\end{itemize}
\subsection{Experimental results: comparison between different discriminators}
\label{subsec:exp_res}
\textbf{\textsc{Doctor}: comparison between $\normalfont\textit{D}_\alpha$ and $\normalfont\textit{D}_\beta$.}
We compare the discriminators $\textit{D}_\alpha$ and $\textit{D}_\beta$ introduced in~\eqref{eq:d_alpha} to show how the AUROCs for CIFAR10, CIFAR100, TinyImageNet and SVHN change when varying the parameters $T$ and $\epsilon$. It is observed that $D_\alpha$ is less sensitive to the selection of $T$: for all the datasets, $D_\alpha$ outperforms $D_\beta$ achieving the best AUROCs by setting $T=1$. Contrary to $D_\alpha$, $D_\beta$ is more sensitive to the value selected for $T$ in the sense that small changes may result in very different values for the measured AUROCs (cf.~\Cref{appendix:alpha_beta}).
In contrast,
the best results are obtained for the same epsilon values of $D_\alpha$ and $D_\beta$ across all the datasets.

\textbf{Comparison in TBB}. We compare \textsc{Doctor} with MHLNB (without input pre-processing and with the softmax output in place of the logits) and SR. It is worth to emphasize that $D_\alpha$ does not coincide (in general) with SR since the former consists in the sum of squared values of all  probabilities involved in the softmax. 
To complete the comparison, we include the results for both methods in~\Cref{tab:best_aurocs}.

\begin{figure*}[!htb]
	\centering
	\begin{subfigure}[b]{0.24\textwidth}
	\centering
	   	\includegraphics[width=\textwidth]{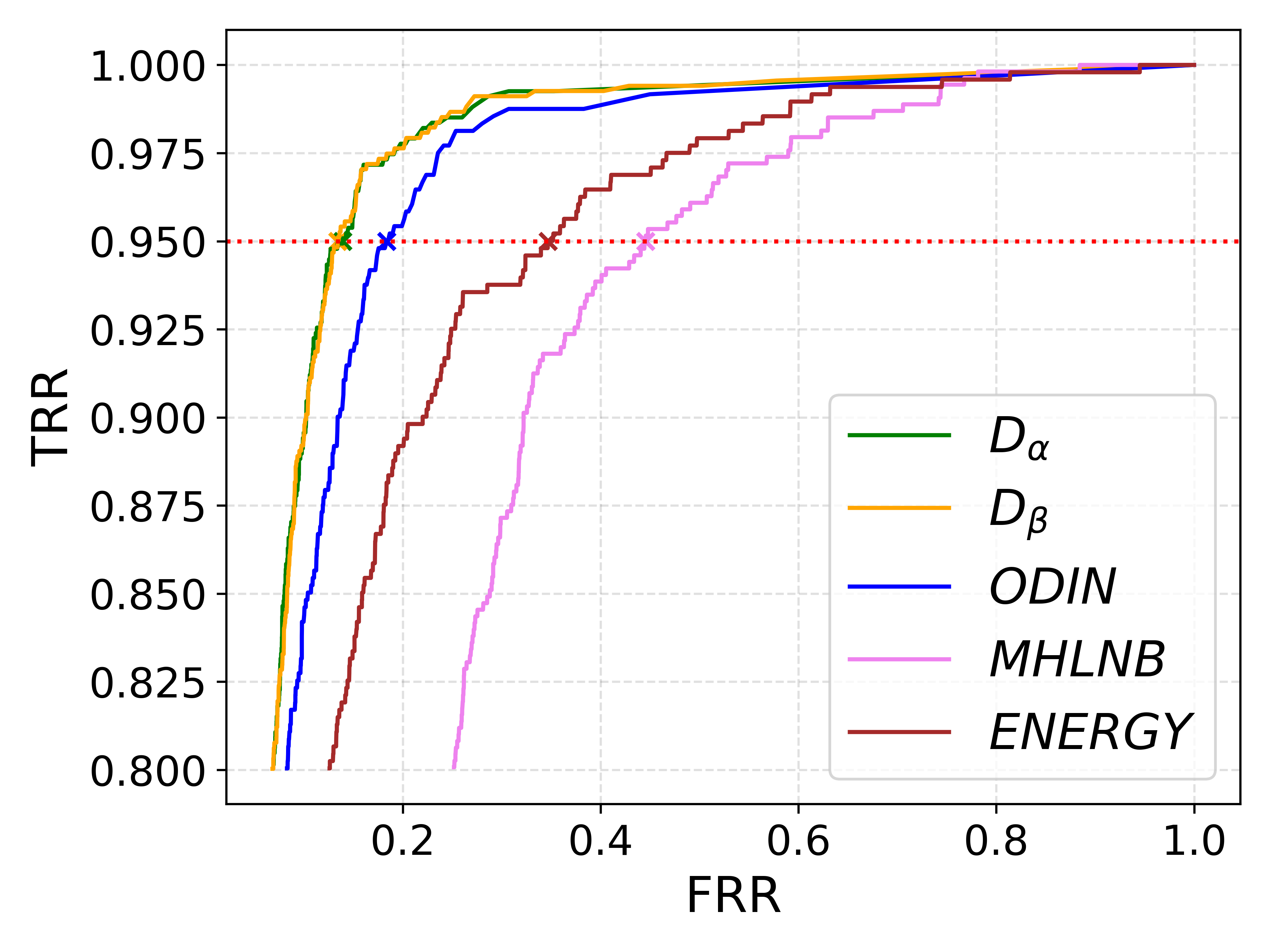}
	    \vspace{-1.5\baselineskip}
	    \caption{CIFAR10 - PBB}
	    \label{fig:cifar10_pbb}
	\end{subfigure}
	\begin{subfigure}[b]{0.24\textwidth}
	\centering
	    \includegraphics[width=\textwidth]{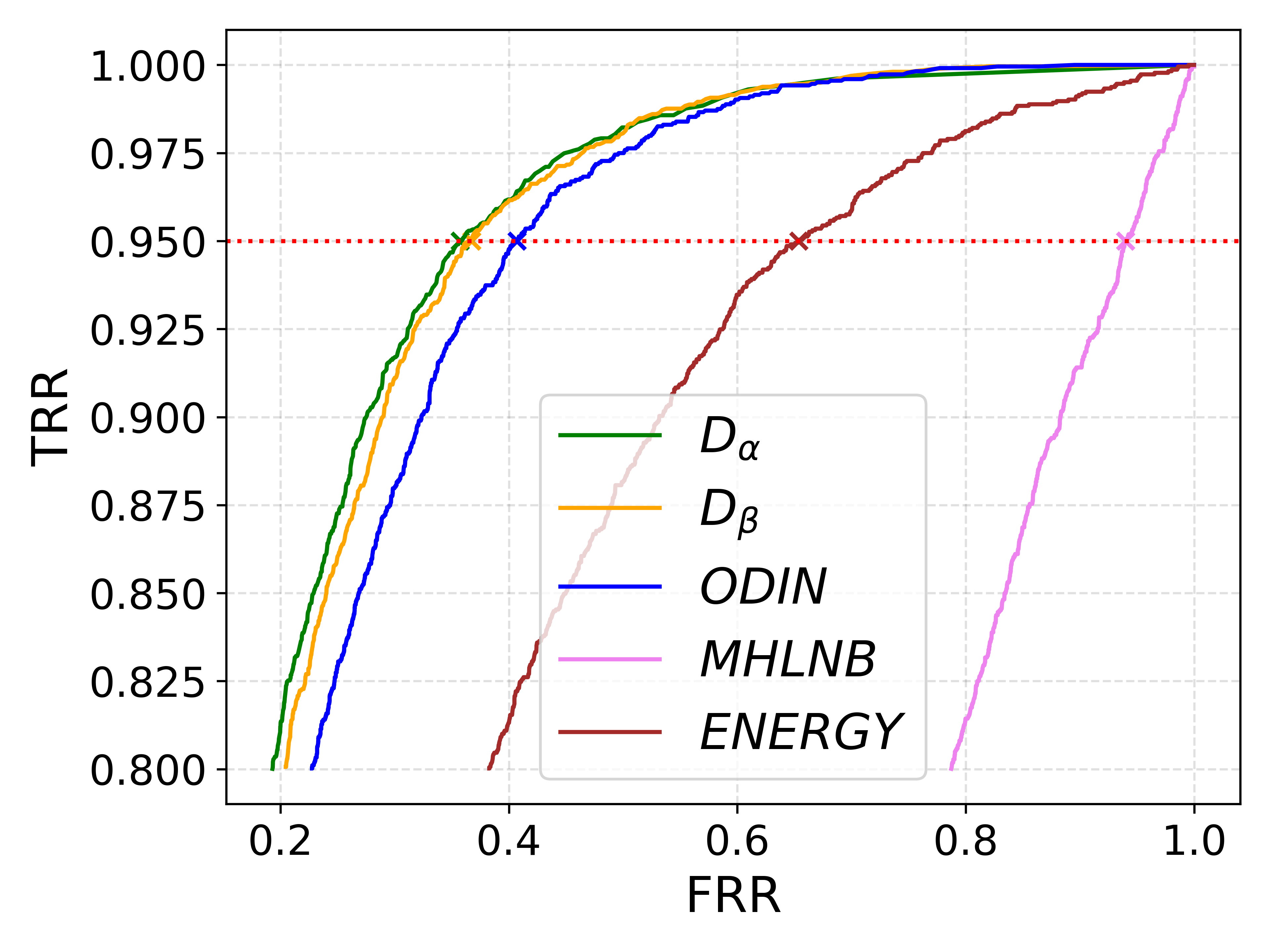}
	    \vspace{-1.5\baselineskip}
	    \caption{CIFAR100 - PBB}
	    \label{fig:cifar100_pbb}
	\end{subfigure}
	\begin{subfigure}[b]{0.24\textwidth}
	\centering
	    \includegraphics[width=\textwidth]{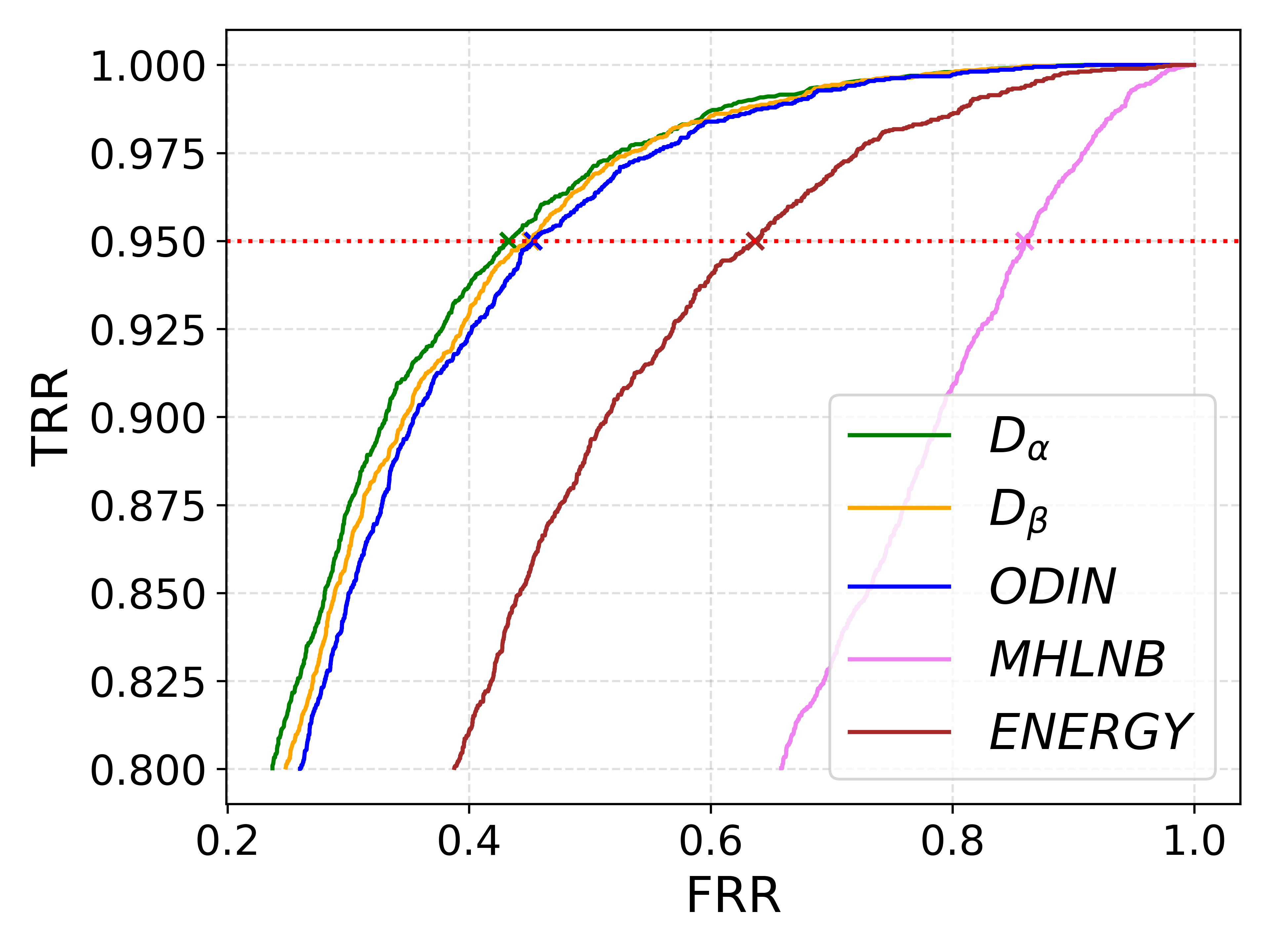}
	    \vspace{-1.5\baselineskip}
	    \caption{TinyImageNet - PBB}
	    \label{fig:tiny_pbb}
	\end{subfigure}
	\begin{subfigure}[b]{0.24\textwidth}
	\centering
	    \includegraphics[width=\textwidth]{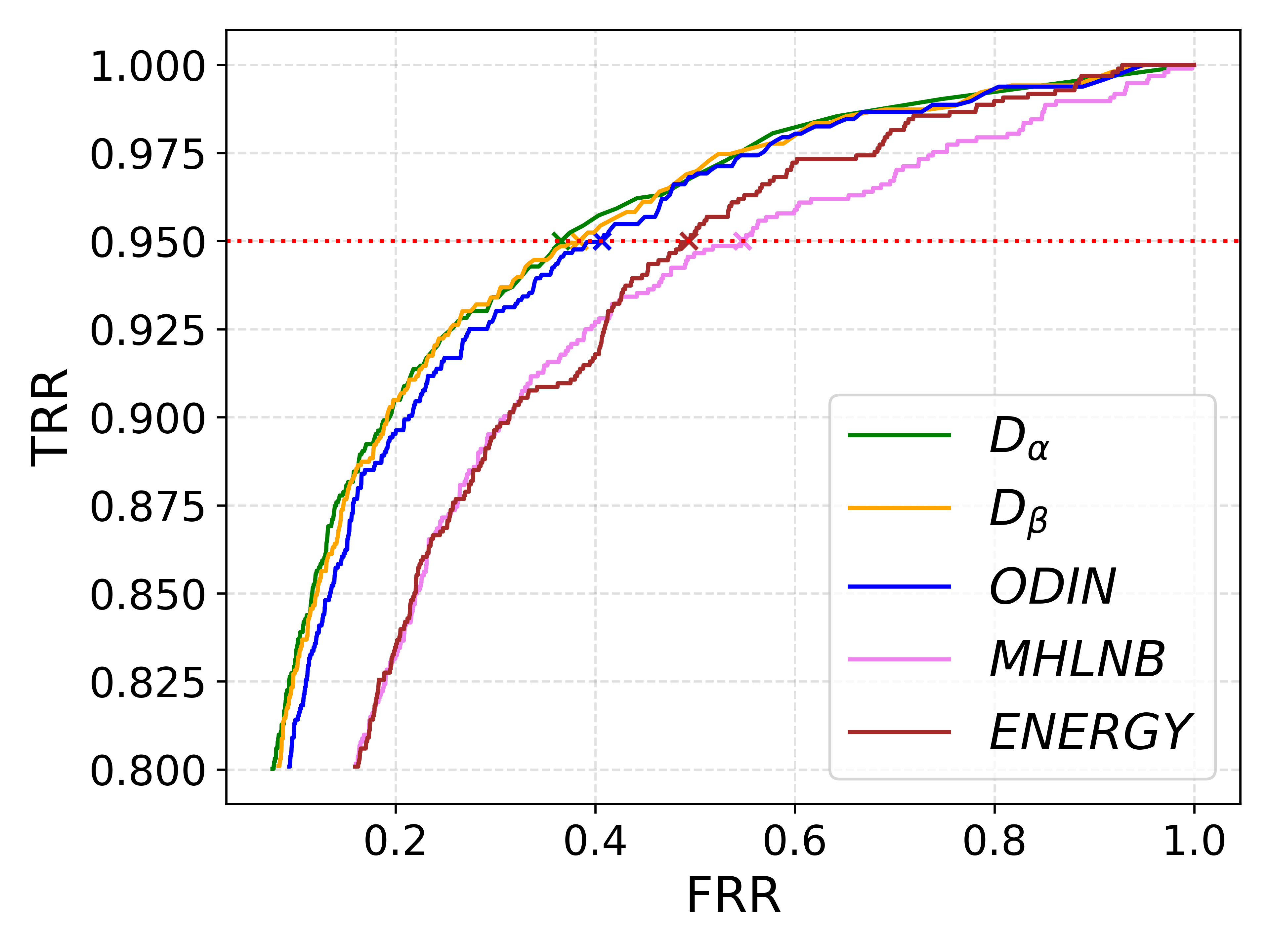}
	    \vspace{-1.5\baselineskip}
	    \caption{SVHN - PBB}
	    \label{fig:svhn_pbb}
	\end{subfigure}
	\caption{ROC curves. Comparison between $D_\alpha$ ($T_\alpha = 1$ and $\epsilon_\alpha = 0.00035$), $D_\beta$ ($T_\beta = 1.5$ and $\epsilon_\alpha = 0.00035$), ODIN ($T_\text{ODIN} = 1.3$ and $\epsilon_\text{ODIN} = 0$), MHLNB ($T_\text{MHLNB} = 1$ and $\epsilon_\text{MHLNB} = 0.0002$) and ENERGY ($T_\text{ENERGY} = 1$ and $\epsilon_{\text{ENERGY}}=0$). Red dashed lines mark the $95\%$ threshold of TRR.}
	\label{fig:best}
\end{figure*}
\textbf{Comparison in PBB.} We compare \textsc{Doctor} with ODIN, MHLNB and {ENERGY}. We keep the same parameter setting for all the methods. In the case of \textsc{Doctor} and ODIN where temperature scaling is allowed, we test, for each dataset, $24$ different values of $\epsilon$ for each of the $11$ different values of $T$, see (\Cref{appendix:overall}) for the set of ranges. In the case of MHLNB, which directly uses the logits, we keep $T=1$ and we vary $\epsilon$ for each dataset. {In the case of ENERGY, where no perturbation is allowed, we keep $\epsilon=0$ and we maintain $T=1$ (as in~\cite{LiuNIPS2020})}. According to our framework, no validation samples are available; consequently, in order to be consistent across the datasets, we only report the experimental settings and values for which, on average, we obtain favorable results for all the considered domains {(cf.~\Cref{fig:best}).}
{In order to be fair, we update ODIN's parameters from those in~\cite{LiangLS2018ICLR} to new values which are more suitable to the task at hand (cf. plots in~\Cref{appendix:overall}).}

\textsc{Doctor}'s performance compared to ODIN's, MHLNB's and {ENERGY's}, are collected in~\Cref{tab:best_aurocs} and in~\Cref{fig:best}. The results in the table show that noise further improves the performance of \textsc{Doctor} (cf. PBB) up to $1\%$ over our previous experiments without noise (cf. TBB) in terms of AUROC. The improvement is even more significant in terms of FRR at $95\%$ TRR: {\textit{a $4\%$ decrease is obtained in terms of predictions incorrectly rejected for \textsc{Doctor} when passing from TBB to PBB}}. 
Note that only the softmax output is available when we consider the pre-trained models for AmazonFashion, AmazonSoftware and IMDb datasets; therefore, we cannot access any internal layer and test \textsc{Doctor} for values of $T$ which differ from the default value $T=1$. Consequently, temperature scaling and input pre-processing cannot be applied in these cases and thus these datasets cannot be tested in PBB. Moreover, even in TBB, these datasets cannot be tested through MHLNB and {ENERGY} since the dataset on which the network was trained is not available.
We provide simulations on how the range of interval for the different thresholds
can affect the results in~\Cref{appendix:intervals}.

\textbf{Misclassification detection in presence of OOD samples}. 
We evaluate \textsc{Doctor}'s performance in 
misclassifcation detection considering a mixture of both in (\textsc{Dataset-in}) and out-of-distribution (OOD) samples (\textsc{Dataset-out}), i.e. input samples for which the decision should not be trusted. The results are compared with ODIN.
We test the two methods when one sample to reject out of  
five ($\clubsuit$), three ($\diamondsuit$) or two ($\spadesuit$) 
is OOD. The details of the simulations, the considered dataset, and the complete experimental results are relegated~\Cref{appendix:ood}.
In~\Cref{tab:ood_main}  we report an extract of the results for the PBB scenario {in terms of \textit{mean / standard deviation}}: \textsc{Doctor} achieves, and most of the time outperforms ODIN's performance. 
We emphasize that, even though \textsc{Doctor} is not tuned for the OOD detection problem, it represents the best choice for deciding whether to accept or reject the prediction of the classifier also on mixed data scenarios where the percentage of OOD samples, as long as it is not dominant, can sensitively vary.

\begin{table}[!htb]
\centering
	\caption{For all methods, in TBB, we set $T=1$ and $\epsilon=0$; in PBB we set : $\epsilon_\alpha = \epsilon_\beta = 0.00035$,
	$T_\alpha = 1$, 
	$T_\beta = 1.5$, $\epsilon_\text{ODIN} = 0$ and $T_\text{ODIN} = 1.3$, $\epsilon_\text{MHLNB} = 0.0002$ and $T_\text{MHLNB} = 1$, $\epsilon_{\text{ENERGY}}=0$ and $T_{\text{ENERGY}} = 1$. 
	In TBB, ODIN and SR coincide ($T=1$ and $\epsilon=0$).}
	\begin{center}
			%
			\begin{small}
			\resizebox{0.9\textwidth}{!}{
						\begin{sc}
						\begin{tabular}{c|c|c|c|c|c}
								\toprule
							\multirow{2}{5em}{\textbf{DATASET}} & \multirow{2}{5em}{\textbf{METHOD}} & \multicolumn{2}{c}{\textbf{AUROC} $\%$} & \multicolumn{2}{|c}{\textbf{FRR $\%$ (95 $\%$ TRR) }}\Bstrut\\\cline{3-6}
								& & TBB & PBB & TBB & PBB\Tstrut\Bstrut\\
								\hline
								\midrule
								\multirow{5}{5em}{\parbox[t][][c]{3cm}{
 CIFAR10\\\textnormal{Acc.} $95\%$}}  
& $D_\alpha$ & \textbf{94} & \textbf{95.2} & \textbf{17.9} & 13.9\Bstrut\\\cline{2-6}
& $D_\beta$  &   68.5 & 94.8 & 18.6 & \textbf{13.4}\Tstrut\Bstrut\\\cline{2-6}
& ODIN       & 93.8 & 94.2 & 18.2 & 18.4\Tstrut\Bstrut\\\cline{2-6}
& SR         & 93.8 & - & 18.2 & -\Tstrut\Bstrut\\\cline{2-6}
& MHLNB      &   92.2  & 84.4 & 30.8 & 44.6\Tstrut\Bstrut\\\cline{2-6}
& ENERGY      &   -   & 91.1 & - & 34.7\Tstrut\\
								\midrule
								\multirow{5}{5em}{\parbox[t][][c]{3cm}{
 CIFAR100\\\textnormal{Acc.} $78\%$}} 
& $D_\alpha$ & \textbf{87} &\textbf{88.2} & 40.6 & \textbf{35.7}\Bstrut\\\cline{2-6}
& $D_\beta$  &   84.2  & 87.4 & 40.6 & 36.7\Tstrut\Bstrut\\\cline{2-6}
& ODIN       &  86.9 & 87.1 & 40.5 & 40.7\Tstrut\Bstrut\\\cline{2-6}
& SR         &  86.9 & - & \textbf{40.5} &-\Tstrut\Bstrut\\\cline{2-6}
& MHLNB      &   82.6   & 50 & 66.7 & 94\Tstrut\Bstrut\\\cline{2-6}
& ENERGY      &   -   & 78.7 & - & 65.4\Tstrut\\
								\midrule
								\multirow{5}{5em}{\parbox[t][][c]{3cm}{
 Tiny\\ImageNet\\\textnormal{Acc.} $63\%$}}   
& $D_\alpha$ & \textbf{84.9} &\textbf{86.1} & \textbf{45.8} &  \textbf{43.3} \Bstrut\\\cline{2-6}
& $D_\beta$  &  \textbf{84.9}  & 85.3 & \textbf{45.8} & 45.1 \Tstrut\Bstrut\\\cline{2-6}
& ODIN       & 84.9 & 84.9 & 45.8 & 45.3\Tstrut\Bstrut\\\cline{2-6}
& SR         & \textbf{84.9} & - & \textbf{45.8} & -\Tstrut\Bstrut\\\cline{2-6}
& MHLNB      &    78.4 & 59 & 82.3&  86\Tstrut\Bstrut\\\cline{2-6}
& ENERGY      &   -   & 78.2 & - & 63.7\Tstrut\\
								\bottomrule
							\end{tabular}
												\begin{tabular}{c|c|c|c|c|c}
							\toprule
							\multirow{2}{5em}{\textbf{DATASET}} & \multirow{2}{5em}{\textbf{METHOD}} & \multicolumn{2}{c}{\textbf{AUROC} $\%$} & \multicolumn{2}{|c}{\textbf{FRR $\%$ (95 $\%$ TRR) }}\Bstrut\\\cline{3-6}
								& & TBB & PBB & TBB & PBB\Tstrut\Bstrut\\
								\hline
													        \midrule
													        \multirow{5}{5em}{\parbox[t][][c]{3cm}{ SVHN\\ \textnormal{Acc.} $96\%$}} 
							& $D_\alpha$ & \textbf{92.3}  & \textbf{93} & \textbf{38.6} & \textbf{36.6}\Bstrut\\\cline{2-6}
							& $D_\beta$  &  92.2  &  92.8 & 39.7 & 38.4\Tstrut\Bstrut\\\cline{2-6}
							& ODIN       & 92.3  & 92.3 &  38.6 & 40.7  \Tstrut\Bstrut\\\cline{2-6}
							& SR         & \textbf{92.3} &  - & \textbf{38.6}  & - \Tstrut\Bstrut\\\cline{2-6}
							& MHLNB      &  87.3    & 88 & 85.8 & 54.7\Tstrut\Tstrut\Bstrut\\\cline{2-6}
& ENERGY      &   -   & 88.9 & - & 49.4\Tstrut\\
								\midrule
															\multirow{3}{5em}{\parbox[t][][c]{3cm}{
							 Amazon\\Fashion\\\textnormal{Acc.} $85\%$}}  
							& $D_\alpha$ & \textbf{89.7}  & - & 27.1 & -\Bstrut\\\cline{2-6}
							& $D_\beta$  & \textbf{89.7} &  - & \textbf{26.3} & -\Tstrut\Bstrut\\\cline{2-6}
							& SR         & 87.4 & - & 50.1 &  -\Tstrut\\
															\midrule
															\multirow{3}{5em}{\parbox[t][][c]{3cm}{
							 Amazon\\Software\\\textnormal{Acc.} $73\%$}} 
							& $D_\alpha$ & \textbf{68.8} &  - & \textbf{73.2} & -\Bstrut\\\cline{2-6}
							& $D_\beta$  &  \textbf{68.8}  & - & \textbf{73.2} & -\Tstrut\Bstrut\\\cline{2-6}
							& SR         &  67.3 & - & 86.6 & -\Tstrut\Bstrut\Tstrut\\
															\midrule
															\multirow{3}{5em}{\parbox[t][][c]{3cm}{
							 IMDb\\\textnormal{Acc.} $90\%$}}           
							& $D_\alpha$ & \textbf{84.4}  & - & \textbf{54.2} & -\Bstrut\\\cline{2-6}
							& $D_\beta$  &   \textbf{84.4}    & - & 54.4 & -\Tstrut\Bstrut\\\cline{2-6}
							& SR         & 83.7  & - & 61.7 & -\Tstrut\Bstrut\Tstrut\\
							\bottomrule
						\end{tabular}
						\end{sc}
				}
			\end{small}
		\end{center}
	\label{tab:best_aurocs}
\end{table}

\begin{table}[!htb]
\centering
	\caption{Same parameter setting as in~\cref{tab:best_aurocs} (PBB) for $D_\alpha$, $D_\beta$, ODIN, ENERGY; as in~\cite{LiangLS2018ICLR} for  ODIN$_\text{OOD}$ and as in~\cite{LeeLLS2018NeurIPS} for MHLNB$_\text{WB}$. 
	Results presented in terms of \textit{mean / standard deviation}.
	}
	\begin{center}
		\begin{small}
		\resizebox{1\textwidth}{!}{
		\begin{sc}
			\begin{tabular}{c|c|c|c|c|c|c|c|c|c|c|c|c|c}
				\toprule
				\multirow{2}{5em}{\centering\textbf{DATASET-In}} & \multirow{2}{5em}{\centering\textbf{DATASET-Out}} & \multicolumn{6}{c}{\textbf{AUROC $\%$}} & \multicolumn{6}{|c}{\textbf{FRR $\%$ (95 $\%$ TRR)}}\Bstrut\\\cline{3-14}
								& & $D_\alpha$ & $D_\beta$ & ODIN & ODIN$_{\text{ood}}$ & ENERGY & MHLNB$_{\text{WB}}$& $D_\alpha$ & $D_\beta$ & ODIN & ODIN$_{\text{ood}}$ & ENERGY & MHLNB$_{\text{WB}}$ \Tstrut\\
								\hline
								\midrule
								\multirow{2}{5em}{\parbox[t][][t]{1cm}{\centering
 CIFAR10\\$\clubsuit$}}  
& iSUN & \textbf{95.4} / 0.1  & 95.1 / 0.1  & 94.6 / 0.1  & 89.6 / 0 & 92.4 / 0.1 & 54.5 / 0.1 &  14 / 0.5 &  \textbf{13.5} / 0.4 &  17.2 / 0.3 &  38.9 / 0 & 32.2 / 0.1 & 92 / 0.1\Bstrut\\\cline{2-14}
& Tiny (res) & \textbf{95.2} / 0.1  & 94.9 / 0  & 94.6 / 0.1  & 89.6 / 0  & 92.3 / 0.1 & 56.2 / 0 & \textbf{14} / 0.4 &  \textbf{14} / 0.5 &  17.8 / 0.4 &  38.9 / 0 & 32.2 / 0.1 & 90.3 / 0.2 \Tstrut\Bstrut\\
\midrule
\multirow{2}{5em}{\parbox[t][][t]{1cm}{\centering
 CIFAR10\\$\diamondsuit$}}  
& iSUN  & \textbf{95.5} / 0.1  & 95.3 / 0.1  & 94.9 / 0.1 & 91.5 / 0 & 92.9 / 0 & 54.5 / 0.1 & 14.4 / 0.6 &  \textbf{13.4} / 0.2 &  16.8 / 0.5 & 34/ 0.1 & 27 / 1 & 92 / 0.2 \Bstrut\\\cline{2-14}
& Tiny (res) & \textbf{95.4} / 0.1  & 95 / 0.1  & 94.8 / 0.1  & 91.4 / 0& 92.8 / 0 & 56.2 / 0.1 &  15 / 0.1 &  1\textbf{4.8} / 0.7 &  17 / 0.5 & 34.5 / 0.9 & 28.8 / 1.9 & 90 / 0.3\Tstrut\Bstrut\\
\midrule
\multirow{2}{5em}{\parbox[t][][t]{1cm}{\centering
 CIFAR10\\$\spadesuit$}}  
& iSUN  &  \textbf{95.6} / 0.1  & \textbf{95.6} / 0  & 95.4 / 0 & 93.5 / 0 & 93.6 / 0.1 & 54.6 / 0 & 15.1 / 0.1 &  \textbf{13.6} / 0.5 &  16.1 / 0.2 & 30.6 / 0.4 & 25.1 / 0.2 & 92 / 0.2  \Bstrut\\\cline{2-14}
& Tiny (res) &  \textbf{95.5} / 0.1  & 95.2 / 0.1  & 95.1 / 0.1  & 93.2 & 93.5 / 0 & 56.2 / 0.2 & \textbf{14.7} / 0.3 &  14.8 / 0.5 &  17.1 / 0.4 & 31/ 0 & 25.6 / 0.3 & 90.2 / 0.1\Tstrut\Bstrut\\
\bottomrule
\end{tabular}
		\end{sc}
		}
	\end{small}
\end{center}
\label{tab:ood_main}
\end{table}

\section{Summary and Concluding Remarks}
\label{sec:conclusion}
We introduced a simple and effective method  to detect misclassification errors, i.e., whether a prediction of a classifier should or should not be trusted.
We provided theoretical results on the optimal statistical model for misclassification detection and we presented our empirical discriminator \textsc{Doctor}. Experiments on real (textual and visual) datasets--including OOD samples and comparison to state-of-the-art methods-- demonstrate the effectiveness of our proposed methods. {Whilst methods for ODD frameworks do not necessarily perform well in predicting misclassification errors,} our result advances the state-of-the-art, and the main takeaway is that \textsc{Doctor} can be applied to both partially black-box (PBB) setups and totally black-box (TBB) ones. In the latter, information about the model's architecture may be undisclosed for security reason when dealing with sensitive data). 
\textsc{Doctor} 
uses all the information 
in the softmax output, which results in equal or better performance with respect to the other methods: the results in PBB, where we observe a reduction up to $4\%$ in terms of predictions incorrectly rejected with respect to the ones in TBB are particularly promising.
Moreover, \textsc{Doctor} does not require training data and, thanks to its flexibility, it can be easily deployed in real-world scenarios.
{
Currently, DOCTOR does not exploit information across the layers yet. Only the soft-predictions are used. Besides, the most important obstacle is the calibration of the threshold ($\gamma$) between the desired fault rejection and acceptance rates, which would require additional validation samples. However, quite often, the cost of collecting data for this operation can be prohibitive, making it difficult or too expensive to perform such calibration.}
As future work, we shall combine \textsc{Doctor} with other related lines of research such as: the extension to white-box incorporating additional information across the different latent codes of the model.
{Moreover, we shall investigate the possibility of combining the two proposed discriminators.}

\bibliography{biblio}

\begin{thebibliography}{10}

\bibitem{ChenLWLJ2020CoRR}
J.~Chen, Y.~Li, X.~Wu, Y.~Liang, and S.~Jha.
\newblock Robust out-of-distribution detection in neural networks.
\newblock {\em arXiv preprint arXiv:2003.09711}, 2020.

\bibitem{DavisGICML2006}
J.~Davis and M.~Goadrich.
\newblock The relationship between precision-recall and {ROC} curves.
\newblock In {\em Machine Learning, Proceedings of the Twenty-Third
  International Conference {(ICML} 2006), Pittsburgh, Pennsylvania, USA, June
  25-29, 2006}, volume 148, pages 233--240, 2006.

\bibitem{DevlinJCMLKT2019NAACLHLT}
J.~Devlin, M.-W. Chang, K.~Lee, and K.~Toutanova.
\newblock Bert: Pre-training of deep bidirectional transformers for language
  understanding.
\newblock In {\em Proceedings of the 2019 Conference of the North American
  Chapter of the Association for Computational Linguistics: Human Language
  Technologies, NAACL-HLT 2019, Minneapolis, MN, USA, June 2-7, 2019, Volume 1
  (Long and Short Papers)}, pages 4171--4186, 2019.

\bibitem{DeVriesWT2018CoRR}
T.~DeVries and G.~W. Taylor.
\newblock Learning confidence for out-of-distribution detection in neural
  networks.
\newblock {\em CoRR}, abs/1802.04865, 2018.

\bibitem{GangradeAISTATS2021}
A.~Gangrade, A.~Kag, and V.~Saligrama.
\newblock Selective classification via one-sided prediction.
\newblock In A.~Banerjee and K.~Fukumizu, editors, {\em The 24th International
  Conference on Artificial Intelligence and Statistics, {AISTATS} 2021, April
  13-15, 2021, Virtual Event}, volume 130 of {\em Proceedings of Machine
  Learning Research}, pages 2179--2187. {PMLR}, 2021.

\bibitem{GeifmanE2017NIPS}
Y.~Geifman and R.~El{-}Yaniv.
\newblock Selective classification for deep neural networks.
\newblock In I.~Guyon, U.~von Luxburg, S.~Bengio, H.~M. Wallach, R.~Fergus,
  S.~V.~N. Vishwanathan, and R.~Garnett, editors, {\em Advances in Neural
  Information Processing Systems 30: Annual Conference on Neural Information
  Processing Systems 2017, December 4-9, 2017, Long Beach, CA, {USA}}, pages
  4878--4887, 2017.

\bibitem{GeifmanY2019ICML}
Y.~Geifman and R.~El{-}Yaniv.
\newblock Selectivenet: {A} deep neural network with an integrated reject
  option.
\newblock In {\em Proceedings of the 36th International Conference on Machine
  Learning, {ICML} 2019, 9-15 June 2019, Long Beach, California, {USA}},
  volume~97, pages 2151--2159, 2019.

\bibitem{GeifmanUY2019ICLR}
Y.~Geifman, G.~Uziel, and R.~El{-}Yaniv.
\newblock Reduced uncertainty estimation for deep neural classifiers.
\newblock In {\em 7th International Conference on Learning Representations,
  {ICLR} 2019, New Orleans, LA, USA, May 6-9, 2019}, 2019.

\bibitem{GuoPSW2017ICML}
C.~Guo, G.~Pleiss, Y.~Sun, and K.~Q. Weinberger.
\newblock On calibration of modern neural networks.
\newblock In {\em Proceedings of the 34th International Conference on Machine
  Learning, {ICML} 2017, Sydney, NSW, Australia, 6-11 August 2017}, volume~70,
  pages 1321--1330, 2017.

\bibitem{HeZRS2016CVPR}
K.~He, X.~Zhang, S.~Ren, and J.~Sun.
\newblock Deep residual learning for image recognition.
\newblock In {\em CVPR}, pages 770--778, 2016.

\bibitem{HeinAB2018CVPR}
M.~Hein, M.~Andriushchenko, and J.~Bitterwolf.
\newblock Why networks yield high-confidence predictions far away from the
  training data and how to mitigate the problem.
\newblock In {\em {IEEE} Conference on Computer Vision and Pattern Recognition,
  {CVPR} 2019, Long Beach, CA, USA, June 16-20, 2019}, pages 41--50. Computer
  Vision Foundation / {IEEE}, 2019.

\bibitem{HendrycksG2017ICLR}
D.~Hendrycks and K.~Gimpel.
\newblock A baseline for detecting misclassified and out-of-distribution
  examples in neural networks.
\newblock In {\em 5th International Conference on Learning Representations,
  {ICLR} 2017, Toulon, France, April 24-26, 2017, Conference Track
  Proceedings}, 2017.

\bibitem{HintonVD2015NIPSWS}
G.~Hinton, O.~Vinyals, and J.~Dean.
\newblock Distilling the knowledge in a neural network.
\newblock In {\em NIPS Deep Learning and Representation Learning Workshop},
  2015.

\bibitem{HsuSJK2020IEEEX}
Y.~Hsu, Y.~Shen, H.~Jin, and Z.~Kira.
\newblock Generalized {ODIN:} detecting out-of-distribution image without
  learning from out-of-distribution data.
\newblock In {\em 2020 {IEEE/CVF} Conference on Computer Vision and Pattern
  Recognition, {CVPR} 2020, Seattle, WA, USA, June 13-19, 2020}, pages
  10948--10957, 2020.

\bibitem{HuangLW2016CoRR}
G.~Huang, Z.~Liu, and K.~Q. Weinberger.
\newblock Densely connected convolutional networks.
\newblock {\em CoRR}, abs/1608.06993, 2016.

\bibitem{WuZX2017Report}
Q.~Z. Jiayu~Wu and G.~Xu.
\newblock Tiny imagenet challenge.
\newblock Technical report, 2017.

\bibitem{KristiadiHH2020ICML}
A.~Kristiadi, M.~Hein, and P.~Hennig.
\newblock Being bayesian, even just a bit, fixes overconfidence in relu
  networks.
\newblock In {\em Proceedings of the 37th International Conference on Machine
  Learning, {ICML} 2020, 13-18 July 2020, Virtual Event}, volume 119 of {\em
  Proceedings of Machine Learning Research}, pages 5436--5446. {PMLR}, 2020.

\bibitem{Cifar}
A.~Krizhevsky.
\newblock Learning multiple layers of features from tiny images.
\newblock Technical report, 2009.

\bibitem{KuleshovE2016CoRR}
V.~Kuleshov and S.~Ermon.
\newblock Reliable confidence estimation via online learning.
\newblock {\em CoRR}, abs/1607.03594, 2016.

\bibitem{KuleshovL2015NIPS}
V.~Kuleshov and P.~Liang.
\newblock Calibrated structured prediction.
\newblock In C.~Cortes, N.~D. Lawrence, D.~D. Lee, M.~Sugiyama, and R.~Garnett,
  editors, {\em Advances in Neural Information Processing Systems 28: Annual
  Conference on Neural Information Processing Systems 2015, December 7-12,
  2015, Montreal, Quebec, Canada}, pages 3474--3482, 2015.

\bibitem{LeeLLS2019ICLR}
K.~Lee, H.~Lee, K.~Lee, and J.~Shin.
\newblock Training confidence-calibrated classifiers for detecting
  out-of-distribution samples.
\newblock In {\em 6th International Conference on Learning Representations,
  {ICLR} 2018, Vancouver, BC, Canada, April 30 - May 3, 2018, Conference Track
  Proceedings}, 2018.

\bibitem{LeeLLS2018NeurIPS}
K.~Lee, K.~Lee, H.~Lee, and J.~Shin.
\newblock A simple unified framework for detecting out-of-distribution samples
  and adversarial attacks.
\newblock In {\em Advances in Neural Information Processing Systems 31: Annual
  Conference on Neural Information Processing Systems 2018, NeurIPS 2018,
  December 3-8, 2018, Montr{\'{e}}al, Canada}, pages 7167--7177, 2018.

\bibitem{LiangLS2018ICLR}
S.~Liang, Y.~Li, and R.~Srikant.
\newblock Enhancing the reliability of out-of-distribution image detection in
  neural networks.
\newblock In {\em 6th International Conference on Learning Representations,
  {ICLR} 2018, Vancouver, BC, Canada, April 30 - May 3, 2018, Conference Track
  Proceedings}, 2018.

\bibitem{LiuNIPS2020}
W.~Liu, X.~Wang, J.~D. Owens, and Y.~Li.
\newblock Energy-based out-of-distribution detection.
\newblock In H.~Larochelle, M.~Ranzato, R.~Hadsell, M.~Balcan, and H.~Lin,
  editors, {\em Advances in Neural Information Processing Systems 33: Annual
  Conference on Neural Information Processing Systems 2020, NeurIPS 2020,
  December 6-12, 2020, virtual}, 2020.

\bibitem{MaasEtAl2011ACL-HLT}
A.~L. Maas, R.~E. Daly, P.~T. Pham, D.~Huang, A.~Y. Ng, and C.~Potts.
\newblock Learning word vectors for sentiment analysis.
\newblock In {\em Proceedings of the 49th Annual Meeting of the Association for
  Computational Linguistics: Human Language Technologies}, pages 142--150, June
  2011.

\bibitem{MeinkeH2020ICLR}
A.~Meinke and M.~Hein.
\newblock Neural networks that provably know when they don't know.
\newblock In {\em 8th International Conference on Learning Representations,
  {ICLR} 2020, Addis Ababa, Ethiopia, April 26-30, 2020}, 2020.

\bibitem{SVHN}
Y.~Netzer, T.~Wang, A.~Coates, A.~Bissacco, B.~Wu, and A.~Ng.
\newblock Reading digits in natural images with unsupervised feature learning.
\newblock {\em NIPS}, 01 2011.

\bibitem{NiLM2019EMNLP}
J.~Ni, J.~Li, and J.~J. McAuley.
\newblock Justifying recommendations using distantly-labeled reviews and
  fine-grained aspects.
\newblock In {\em Proceedings of the 2019 Conference on Empirical Methods in
  Natural Language Processing and the 9th International Joint Conference on
  Natural Language Processing, {EMNLP-IJCNLP} 2019, Hong Kong, China, November
  3-7, 2019}, pages 188--197, 2019.

\bibitem{Platt1999ALMC}
J.~Platt.
\newblock Probabilistic outputs for support vector machines and comparisons to
  regularized likelihood methods.
\newblock {\em Adv. Large Margin Classif.}, 10, 06 2000.

\bibitem{10.5555/1522486}
A.~B. Tsybakov.
\newblock {\em Introduction to Nonparametric Estimation}.
\newblock Springer Publishing Company, Incorporated, 1st edition, 2008.

\bibitem{erven2014}
T.~van Erven and P.~Harremos.
\newblock {R\'enyi Divergence and Kullback-Leibler Divergence}.
\newblock {\em IEEE Transactions on Information Theory}, 60(7):3797--3820,
  2014.

\bibitem{pmlr-v128-vapnik20a}
V.~Vapnik and R.~Izmailov.
\newblock Complete statistical theory of learning: learning using statistical
  invariants.
\newblock In {\em Proceedings of the Ninth Symposium on Conformal and
  Probabilistic Prediction and Applications}, volume 128, pages 4--40, 2020.

\bibitem{VyasJZDKW2018EECV}
A.~Vyas, N.~Jammalamadaka, X.~Zhu, D.~Das, B.~Kaul, and T.~L. Willke.
\newblock Out-of-distribution detection using an ensemble of self supervised
  leave-out classifiers.
\newblock In {\em Computer Vision - {ECCV} 2018 - 15th European Conference,
  Munich, Germany, September 8-14, 2018, Proceedings, Part {VIII}}, volume
  11212, pages 560--574, 2018.

\bibitem{Wolfetall2020NLP}
T.~Wolf, L.~Debut, V.~Sanh, J.~Chaumond, C.~Delangue, A.~Moi, P.~Cistac,
  T.~Rault, R.~Louf, M.~Funtowicz, J.~Davison, S.~Shleifer, P.~von Platen,
  C.~Ma, Y.~Jernite, J.~Plu, C.~Xu, T.~L. Scao, S.~Gugger, M.~Drame, Q.~Lhoest,
  and A.~M. Rush.
\newblock Transformers: State-of-the-art natural language processing.
\newblock In {\em Proceedings of the 2020 Conference on Empirical Methods in
  Natural Language Processing: System Demonstrations}, pages 38--45, oct 2020.

\bibitem{XingAZPICLR2020}
C.~Xing, S.~Arik, Z.~Zhang, and T.~Pfister.
\newblock Distance-based learning from errors for confidence calibration.
\newblock In {\em International Conference on Learning Representations}, 2020.

\bibitem{ZhangDS2019CoRR}
Z.~Zhang, A.~V. Dalca, and M.~R. Sabuncu.
\newblock Convolutional neural networks using structured dropout.
\newblock {\em CoRR}, abs/1906.09551, 2019.

\end{thebibliography}
\bibliographystyle{abbrv}
\newpage
\begin{figure*}[ht]
	\centering
	{\LARGE\textbf{Supplementary Material}}
\end{figure*}
\appendix
\section{Proofs}
\label{appendix:proofs}
The following section shows the proofs for Proposition (\ref{prop:1}), Proposition (\ref{prop:2}) and Inequalities (\ref{eq-misssing-1}).
\subsection{Proof of Proposition \ref{prop:1}}
\label{appendix:proof1}

We  recall the definition of the total variation distance when applied to distributions $P$, $Q$ on a set $\mathcal{X}\subseteq \mathbb{R}^d $ and the Scheff\'e's identity, Lemma~2.1 in \cite{10.5555/1522486}:
\begin{equation}
  \label{eq:scheffe}
\| P - Q\|_\text{TV} \myeq \sup_{\mathcal{A}\in  \mathcal{B}^d} |P(\mathcal{A}) - Q(\mathcal{A})| =\frac12 \int |p_{X}(\mathbf{x})- q_{X}(\mathbf{x})|d\mu(\mathbf{x})
\end{equation}
with respect to a base measure $\mu$, where $\mathcal{B}^d$ denotes the class of all Borel sets on $\mathbb{R}^d$.

\begin{proof}
	 First of all, we prove the equality for $\gamma=1$. Let us denote with $ \mathcal{A}^\star\equiv \mathcal{A}(1)$ and ${ \mathcal{A}^\star}^c \equiv \mathcal{A}^c(1)$ the optimal decision regions from \eqref{eq:opt2}. Let $\epsilon_{\text{0}}({ \mathcal{A}^\star})$ and $\epsilon_{\text{1}}( {\mathcal{A}^\star}^c )$ be the Type-I and Type-II errors, respectively. Then, 
	\begin{align}
\epsilon_{\text{0}}( \mathcal{A}^\star  ) + \epsilon_{\text{1}}({ \mathcal{A}^\star}^c ) &=  \int_{ \mathcal{A}^\star} p_{X|E}(\mathbf{x}|0) d \mathbf{x} + \int_{{ \mathcal{A}^\star}^c}p_{X|E}(\mathbf{x}|1) d \mathbf{x} \nonumber\\
	& = \int_{ \mathcal{A}^\star} \min \Big\{p_{X|E}(\mathbf{x}|0) \,,\, p_{X|E}(\mathbf{x}|1) \Big\} d \mathbf{x} \nonumber \\ 
& + \int_{{ \mathcal{A}^\star}^c} \min \Big\{p_{X|E}(\mathbf{x}|0)  \,,\,  p_{X|E}(\mathbf{x}|1)  \Big\}d \mathbf{x} \nonumber \\
	&= \int_{\mathcal{X}} \min\Big \{p_{X|E}(\mathbf{x}|0) \,,\, p_{X|E}(\mathbf{x}|1)  \Big\}d \mathbf{x}\nonumber\\
&=	1- \left\| p_{X|E=1} -p_{X|E=0} \right\|_\text{TV},\label{eq-last-id}
		\end{align}
		where the last identity follows by applying Scheff\'e's identity~\eqref{eq:scheffe}. From the last identity in \eqref{eq-last-id} and any decision region  $ \mathcal{A}\subseteq \mathcal{X}$, we have
	\begin{align}
	1- \big\| p_{X|E=1} -p_{X|E=0} \big\|_\text{TV} &= \int_{ \mathcal{X}} \min\Big \{p_{X|E}(\mathbf{x}|0)   \,,\, p_{X|E}(\mathbf{x}|1)  \Big\} d \mathbf{x}  \nonumber\\
&= \int_{ \mathcal{A}} \min\Big \{p_{X|E}(\mathbf{x}|0) \,,\,  p_{X|E}(\mathbf{x}|1)  \Big\}d \mathbf{x} \nonumber\\
&+ \int_{ \mathcal{A}^c} \min\Big \{p_{X|E}(\mathbf{x}|0) \,,\, p_{X|E}(\mathbf{x}|1)  \Big\}d \mathbf{x} \nonumber\\
&\leq  \int_{ \mathcal{A}} p_{X|E}(\mathbf{x}|0)d \mathbf{x} + \int_{ \mathcal{A}^c}  p_{X|E}(\mathbf{x}|1) d \mathbf{x} \nonumber \\
& =  \epsilon_{\text{0}}( \mathcal{A}   ) + \epsilon_{\text{1}}( { \mathcal{A}}^c ).
	\end{align}
	
It remains to show the last statement related to the Bayesian error of the test. Assume that $P_E(1)=P_E(0)=1/2$. By using the last identity in \eqref{eq-last-id}, we have 
	\begin{align}
\frac12\Big[	1- \big\| p_{X|E=1} -p_{X|E=0} \big\|_\text{TV}\Big]   & = \frac12\int_{\mathcal{X}} \min\Big \{p_{X|E}(\mathbf{x}|0), p_{X|E}(\mathbf{x}|1)  \Big\}d \mathbf{x} \nonumber\\
& = \int_{\mathcal{X}} \min\Big \{p_{X E}(\mathbf{x}, E = 0), p_{XE}(\mathbf{x}, E=1)\Big\} d\mathbf{x} \nonumber\\
& = \mathbb{E}_{X} \left[ \min\Big \{P_{E|X}(0|\mathbf{X}), P_{E|X}(1|\mathbf{X}) \Big\} \right]\nonumber\\
& =\frac12\big[ \epsilon_{\text{0}}( {\mathcal{A}^\star}) + \epsilon_{\text{1}}( { \mathcal{A}^\star}^c )\big]\nonumber  \\
& \equiv  \inf_D \, \Pr\left\{D(\mathbf{X})\neq E \right\},
	\end{align}
where the last identity follow by the definition of the decision regions in \eqref{eq:opt2}. 
\end{proof}

\subsection{Proof of Proposition \ref{prop:2}}
\label{appendix:proof2}

\begin{proof}
We begin by showing that 
	\begin{align}
	|\widehat{\fun{Pe}}(\mathbf{x}) - \fun{Pe}(\mathbf{x})|   &=   \Big |\mathbb{E} \big[\mathds{1}[\widehat{Y} \neq f_{\mathcal{D}_{n}}(\mathbf{x})] \big |  \mathbf{x} \big] - \mathbb{E}\big[\mathds{1}[Y \neq  f_{\mathcal{D}_{n}}(\mathbf{x})] \big | \mathbf{x} \big]\Big|\nonumber\\
	& =  
	\Bigg| \sum_{\{y\in\mathcal{Y} \,| \, y \neq f_{\mathcal{D}_{n}}(\mathbf{x}) \}} \big[P_{\widehat{Y}|X}(y|\mathbf{x}) - P_{Y|X}(y|\mathbf{x})\big] \Bigg|\nonumber\\
&\leq \sum_{\{y\in\mathcal{Y} \,| \, y \neq f_{\mathcal{D}_{n}}(\mathbf{x}) \}} \left| P_{\widehat{Y}|X}(y|\mathbf{x}) - P_{Y|X}(y|\mathbf{x})\right|\nonumber\\	
&\leq \sum_{ y\in\mathcal{Y} } \left| P_{\widehat{Y}|X}(y|\mathbf{x}) - P_{Y|X}(y|\mathbf{x})\right|\nonumber\\
& \leq 2 \left \| P_{\widehat{Y}|X}(\cdot|\mathbf{x}) - P_{Y|X}(\cdot|\mathbf{x}) \right\|_{\fun{TV}}\nonumber\\
& \leq 2 \sqrt{ 2 \fun{KL}\left(P_{Y|\mathbf{x}} \| P_{\widehat{Y}|\mathbf{x}}\right) }, \label{eq-mis0}
	\end{align}
	where $\| \cdot  \|_{\fun{TV}}$ denotes the \textit{Total Variation distance}, $\fun{KL}(\cdot \| \cdot)$ is the \textit{Kullback–Leibler divergence} and the last step is due to \textit{Pinsker's inequality}. On the other hand, 
		\begin{align}
1-	\widehat{\fun{g}}(\mathbf{x}) & = 1 - \underset{y\in\mathcal{Y}}{\sum}P_{\widehat{Y}|X}^2(y|\mathbf{x})\nonumber\\
& = 1 - \mathbb{E}_{\widehat{Y}|X}\left[ P_{\widehat{Y}|X}(\widehat{Y}|\mathbf{x})  | \mathbf{x} \right]\nonumber\\
&\geq 1 - \mathbb{E}_{\widehat{Y}|X}\left[ \max_{y\in\mathcal{Y}} P_{\widehat{Y}|X}(y|\mathbf{x}) \big | \mathbf{x} \right]\nonumber\\
& = 1 - \max_{y\in\mathcal{Y}} P_{\widehat{Y}|X}(y|\mathbf{x})\nonumber\\
& \equiv \widehat{\fun{Pe}}(\mathbf{x}). \label{eq-mis1}
\end{align}
Similarly, 
	\begin{align}
\widehat{\fun{g}}(\mathbf{x}) & = 	\underset{y\in\mathcal{Y}}{\sum}P_{\widehat{Y}|X}^2(y|\mathbf{x})\nonumber\\
& =  P_{\widehat{Y}|X}^2(y^\star|\mathbf{x})  + \underset{y\neq y^\star }{\sum} P_{\widehat{Y}|X}^2(y|\mathbf{x})\nonumber\\
&\geq  \max_{y\in\mathcal{Y}} P_{\widehat{Y}|X}^2 (y|\mathbf{x})\nonumber\\
& \equiv \left( 1 -   \widehat{\fun{Pe}}(\mathbf{x}) \right)^2,  \label{eq-mis2}
\end{align}
where $y^\star =\arg\max_{y\in\mathcal{Y}} P_{\widehat{Y}|X}(y|\mathbf{x}) $. 
By replacing expressions \eqref{eq-mis1} and \eqref{eq-mis2} in \eqref{eq-mis0} we obtained the desired inequalities, which concludes the proof.  
\end{proof}

\subsection{Proof of Inequalities in  \texorpdfstring{\eqref{eq-misssing-1}}{(3)}}
\label{appendix:proof-eq}
\begin{proof}
The event can be decomposed as follows: 
\begin{equation}
\big\{ \widehat{E}(\mathbf{x})  \neq   {E}(\mathbf{x}) | \mathbf{x} \big\} \equiv  \big\{ Y \neq   \widehat{Y} \big\}  \cap  \left\{ \big\{ \widehat{Y}  = f_{\mathcal{D}_{n} }(\mathbf{x}) \big\}\textrm{ or } 
 \big\{ {Y}  = f_{\mathcal{D}_{n}}(\mathbf{x}) \big\}  | \mathbf{x} \right\}
\end{equation}
for all $\mathbf{x}\in\mathcal{X}$. Thus, 
 \begin{align}
 \big\{ \widehat{E}(\mathbf{x})\neq   {E}(\mathbf{x})  | \mathbf{x} \big\}   & \subseteq \big\{ Y \neq   \widehat{Y}  | \mathbf{x}\big\}, \\
\big\{ Y \neq   \widehat{Y} \big\}  \cap  \big\{ {Y}  \neq  f_{\mathcal{D}_{n}}(\mathbf{x}) \big  | \mathbf{x}\}  &  \subseteq \big\{ \widehat{E}(\mathbf{x}) \neq   {E}(\mathbf{x}) | \mathbf{x}\big\},   \\
\big\{ Y \neq   \widehat{Y} \big\}  \cap  \big\{ \widehat{Y}  \neq  f_{\mathcal{D}_{n}}(\mathbf{x}) | \mathbf{x} \big\} & \subseteq   \big\{ \widehat{E}(\mathbf{x}) \neq   {E}(\mathbf{x})  | \mathbf{x} \big\} ,
 \end{align}
which imply 
 \begin{align}
\Pr\big( \{\widehat{E}(\mathbf{x})  \neq   {E}(\mathbf{x})  | \mathbf{x}\}  \big) & \leq  \Pr\big(\{\widehat{Y} \neq  {Y}\} | \mathbf{x}\big),\\
 \fun{Pe}(\mathbf{x}) - \Pr\big(\{\widehat{Y} =  {Y}\} | \mathbf{x}\big)  &\leq \Pr\big( \{\widehat{E}(\mathbf{x}) \neq   {E}(\mathbf{x}) \} \big| \mathbf{x}  \big) ,\\
\widehat{\fun{Pe}}(\mathbf{x}) - \Pr\big(\{\widehat{Y} =  {Y}\} | \mathbf{x}\big) &\leq \Pr\big( \{\widehat{E}(\mathbf{x})  \neq   {E}(\mathbf{x})\} \big| \mathbf{x} \big),  
 \end{align}
 for all $\mathbf{x}\in\mathcal{X}$, where the last inequalities follows by noticing that $\Pr(\mathcal{A} \cap \mathcal{B} ) \geq \Pr(\mathcal{A}) - \Pr(\mathcal{B}^c)$ for arbitrary measurable sets $\mathcal{A}, \mathcal{B}\subset \mathcal{X}$. This concludes the proof of these inequalities. 
\end{proof}


\section{Logistic Regression and Gaussian Model}
\label{appendix:toy}
Throughout this section we test \textsc{Doctor} in a controlled setting were all the involved distributions are known. We refer to that setting as \textit{logistic regression and Gaussian model} since we collect data points from Gaussians distributions and we test on the logistic regression setup.

\subsection{Theoretical analysis}
\label{appendix:gaussianv}
Let $\mathcal{X}=\mathbb{R}^d$ be the feature space and $\mathcal{Y}=\{-1,1\}$  be the label space.
We focus on a binary classification task in which $\mathbf{X}\sim\mathcal{N}(y\bm{\mu}, \sigma^2I)$ and $Y\sim\mathcal{U}(\mathcal{Y})$, where $\bm{\mu}\in\mathbb{R}^n$ is the mean vector, $\sigma^2>0$ is the variance and $I$ is the identity matrix and $\mathcal{U}(\mathcal{Y})$ denotes the uniform distribution over $\mathcal{Y}$.
For a fixed $\bm{\theta}\in\mathbb{R}^d$, consider $f_{\bm{\theta}}:\mathcal{X}\rightarrow\mathcal{Y}$ s.t. $f_{\bm{\theta}}(\mathbf{x}) = \fun{sign}(\fun{sigmoid}(\mathbf{x}^T\bm{\theta})-\nicefrac{1}{2})$. For a given $\mathbf{x}\in\mathcal{X}$, we adapt to the current setting the definition of $E(\mathbf{x})$ in~\cref{sec:preliminaries} as follows:
\begin{align}
\mathds{1}\left[Y \not = f_{\bm{\theta}}(\mathbf{x})\right]
	  & =  \mathds{1}\left[Y \cdot \fun{sign}\left(\fun{sigmoid}\left(\mathbf{x}^T\bm{\theta}\right)-\frac{1}{2}\right) < 0\right].
\end{align}
Let us denote by $\mathds{1}\left[y \not = f_{\bm{\theta}}(\mathbf{x})\right]$ the realization of the random variable $E(\mathbf{x})$. We can compute the probability of classification error $\fun{Pe}(\mathbf{x})$ in~\eqref{eq:pe} w.r.t. the true class posterior probabilities:
\begin{align}
\fun{Pe}(\mathbf{x}) =  \mathbb{E}\Big[\mathds{1}\left[Y \not = f_{\bm{\theta}}(\mathbf{x})\right] | \mathbf{x}\Big] 
=&{}\sum_{y\in \mathcal{Y}}\mathds{1}\left[y \not = f_{\bm{\theta}}(\mathbf{x})\right]\cdot \frac{p_{\mathbf{X}|Y}(\mathbf{x}|y)P_{Y}(y)}{p_{\mathbf{X}}(\mathbf{x})}\nonumber \\
=& \sum_{y\in \mathcal{Y}}\mathds{1}\left[y \not = f_{\bm{\theta}}(\mathbf{x})\right]\cdot \frac{\frac{1}{2}\mathcal{N}(\mathbf{x}; y\mu, \sigma^2I)}{\frac{1}{2}\sum_{y^\prime\in\mathcal{Y}}\mathcal{N}(\mathbf{x}; y^\prime\mu, \sigma^2I)}\nonumber\\
=&{}\frac{\sum_{y\in \mathcal{Y}}\mathds{1}\left[y \not = f_{\bm{\theta}}(\mathbf{x})\right]\cdot\mathcal{N}(\mathbf{x}; y\mu, \sigma^2I)}{\sum_{y\in\mathcal{Y}}\mathcal{N}(\mathbf{x}; y\mu, \sigma^2I)}.
\end{align}
Following~\eqref{eq:opt'}, the decision region corresponding to the most powerful discriminator for the logistic regression and the Gaussian model are given by 
\begin{align}
\mathcal{A}(\gamma) = \left\{\mathbf{x}\in\mathcal{X}:\frac{\sum_{y\in\mathcal{Y}}\mathds{1}\left[y \not = f_{\bm{\theta}}(\mathbf{x})\right] \cdot \mathcal{N}(\mathbf{x}, y\bm{\mu}, \sigma^2I)}{\sum_{y\in\mathcal{Y}}\mathds{1}\left[y  = f_{\bm{\theta}}(\mathbf{x})\right] \cdot \mathcal{N}(\mathbf{x}, y\bm{\mu}, \sigma^2I)}> \gamma\right\}.
\label{toy:opt}
\end{align}
We are now able to state the optimal discriminator for this setting.
\begin{definition}[Optimal discriminator for the logistic regression and the Gaussian model]\label{def:optgaussian}
For any $0<\gamma<\infty$ and $\mathbf{x}\in\mathcal{X}$, the optimal  discriminator follows as:
\begin{align}
    D^\star(\mathbf{x}, \gamma)\myeq\mathds{1}\left[\sum_{y\in \mathcal{Y}} \mathds{1}\left[y \not = f_{\bm{\theta}}(\mathbf{x})\right] \cdot \mathcal{N}(\mathbf{x}, y\bm{\mu}, \sigma^2I) > \gamma \cdot \sum_{y\in \mathcal{Y}} \mathds{1}\left[y = f_{\bm{\theta}}(\mathbf{x})\right] \cdot \mathcal{N}(\mathbf{x}, y\bm{\mu}, \sigma^2I)\right].
    \label{eq:d-star-gaussian}
\end{align}
\end{definition}
Since we cannot analytically evaluate~\cref{prop:1}, we proceed numerically in the next experiment.
\newpage 
\subsection{Experiments}
\begin{wraptable}[21]{r}{6cm}
\caption{
Accuracy on the test set:  $f_{\bm{\theta}_i}\text{ for } i=1,\dots,8$ represents the $i$-th model in $\mathcal{F}$, $f_{avg}$ is the arithmetic mean of the accuracy over each $f_{\bm{\theta}_i} \in \mathcal{F}$. The value $f_{avg}^\star$ represents the accuracy Bayesian classifier averaged on the test set corresponding to the $8$ splits. We show results for both standard deviations, namely $\sigma=2$ and $\sigma=4$.}
\begin{center}
\begin{small}
\begin{sc}
\scalebox{0.9}{
\begin{tabular}{c|cc}
\toprule
\textbf{Classifier} & \multicolumn{2}{c}{\textbf{Accuracy}$\%$}\\
\hline
\midrule
& $\sigma=2$ & $\sigma=4$\\
\midrule
$f_{\bm{\theta}_1}$ & $82$ & $65$\\
$f_{\bm{\theta}_2}$ & $83$ & $77$\\
$f_{\bm{\theta}_3}$ & $82$ & $77$\\
$f_{\bm{\theta}_4}$ & $82$ & $76$\\
$f_{\bm{\theta}_5}$ & $83$ & $76$\\
$f_{\bm{\theta}_6}$ & $81$ & $66$\\
$f_{\bm{\theta}_7}$ & $82$ & $76$\\
$f_{\bm{\theta}_8}$ & $83$ & $83$\\
$f_{avg}$ & $82$ & $74$\\
$f_{avg}^\star$ &  $83$ & $78$\\
\bottomrule
\end{tabular}
}
\end{sc}
\end{small}
\end{center}
\label{tab:toy_acc}
\end{wraptable}

In this section, we will numerically evaluate~\cref{prop:1} via empirical estimates of Type-I and Type-II errors in expressions~\eqref{eq-Type1}. Note that unlike~\cref{sec:experiments}, in this case all the involved distributions are known and hence it is also possible to compute the \emph{true posterior distribution $P_{Y|X}$}. 

We adopt the same notation as in~\cref{sec:performanceandmetrics} for \textsc{Doctor}, i.e., $D_\alpha$, and $D_\beta$ according to according to expressions~\eqref{eq:d_alpha}.
$D^\star$, as in~\cref{def:optgaussian}, denotes the optimal discriminator.

\subsubsection{Experimental setup and evaluation metrics}
\label{appendix:toy_setup}
\textbf{Dataset.} We create a synthetic dataset that consists of $5000$ data points drawn from $\mathcal{N}_0\myeq\mathcal{N}(\bm{\mu}_0, \sigma^2I)$ and $5000$ data points drawn from $\mathcal{N}_1\myeq\mathcal{N}(\bm{\mu}_1, \sigma^2I)$, where $\bm{\mu}_0 = [-1 ~ -1]$, $\bm{\mu}_1 = [1~ 1]$. We consider two values for sigma, namely $\sigma=2$ and $\sigma=4$. These values produce two different distributions which will let us showcase the advantages of \textsc{Doctor}. 
To each data point $\mathbf{x}$ is assigned as class $0$ or $1$ depending on whether $\mathbf{x}\sim\mathcal{N}_0$ or $\mathbf{x}\sim\mathcal{N}_1$, respectively. The aforementioned dataset is divided into a training set, i.e. $\mathcal{D}_{n}=\{(\mathbf{x}_1,y_1),\dots,(\mathbf{x}_{n}, y_{n})\}$ where $n=6700$,  and a testing set, i.e. $\mathcal{T}_{m}=\{(\mathbf{x}_{n+1}, y_{n+1}),\dots,(\mathbf{x}_{n+m},y_{n+m})\}$ where $m=3300$.

\textbf{Training configuration.}
We use a linear classifier, with one hidden layer, sigmoid activation function and binary cross entropy loss. The neural network is trained with gradient descent considering learning rate $r=0.1$. Specifically, we train our network for $5$ epochs. We randomly split our dataset $8$ times, each time keeping $n$ samples to train, and $m$ to test. We consider the same model architecture (described above) for each split and we come up with 8 different binary discriminators $\mathcal{F} = \{ f_{\bm{\theta}_1}, \dots, f_{\bm{\theta}_8}\}$.
Since in this example all the involved distributions are known, we compute the optimal predictor, i.e. the Bayes classifier, and we denote it with $f^\star$. The value $f_{avg}^\star$ reported in~\cref{tab:toy_acc}, represents its accuracy averaged on the test set corresponding to the $8$ splits.

\textbf{Accuracy of trained networks.}
In~\cref{tab:toy_acc} the accuracy of $f^\star$ and the models in $\mathcal{F}$ on the test set.

\textbf{Evaluation metric.} We consider the same metric as in~\cref{sec:experiments_setup}.

\subsubsection{Numerical evaluation of \texorpdfstring{~\cref{prop:1}}{3.1}}
To evaluate~\cref{prop:1} we proceed in a Monte Carlo fashion by computing Type-I and Type-II errors for each of the network in $\mathcal{F}$ and then averaging over the results. Schematically, consider any $f_{\bm{\theta}_i}\in\mathcal{F}$ and $\gamma=1$, we compute:
\begin{enumerate}
\item $\mathcal{A}_i\myeq\mathcal{A}_i(1)$ as defined in~\cref{toy:opt} and its complement $\mathcal{A}^c_i$.

\item For each classifier $f_{\bm{\theta}_i} \in \mathcal{F}$, $\mathcal{T}_{E=1; \bm{\theta}_i} \myeq \{(\mathbf{x}, y) \in\mathcal{T}_{m} ~|~ y \not = f_{\bm{\theta}_i}(\mathbf{x})\}$ represents the set of mis-classified test samples, and $\mathcal{T}_{E=0; \bm{\theta}_i} \myeq \{(\mathbf{x}, y) \in\mathcal{T}_{m} ~|~ y  = f_{\bm{\theta}_i}(\mathbf{x})\}$ is the set of correctly classified test samples.

\item $\mathcal{FR}_i \myeq \left\{(\mathbf{x}, y)\in\mathcal{T}_{E=0; \bm{\theta}_i} : \mathbf{x}\in\mathcal{A}_i\right\}$, $\mathcal{TR}_i \myeq \left\{(\mathbf{x}, y)\in\mathcal{T}_{E=1; \bm{\theta}_i} : \mathbf{x}\in\mathcal{A}_i\right\}$,
$\mathcal{FA}_i \myeq \left\{(\mathbf{x}, y)\in\mathcal{T}_{E=1; \bm{\theta}_i} : \mathbf{x}\in\mathcal{A}^c_i\right\}$
and $\mathcal{TA}_i \myeq \left\{(\mathbf{x}, y)\in\mathcal{T}_{E=0; \bm{\theta}_i} : \mathbf{x}\in\mathcal{A}^c_i\right\}$, i.e. the set of false rejections, true rejections, false acceptances and true acceptance, respectively.

\item $\epsilon_0(\mathcal{A}_i) \myeq \frac{|\mathcal{FR}_i|}{|\mathcal{T}_{E=0; \bm{\theta}_i}|}$ and $
\epsilon_1(\mathcal{A}_i^c) \myeq\frac{|\mathcal{FA}_i|}{|\mathcal{T}_{E=1; \bm{\theta}_i}|}$, i.e. Type-I and Type-II errors. 
\end{enumerate}

At the end of $|\mathcal{F}|$ iterations, we empirically estimate Type-I and Type-II errors of~\cref{prop:1} as follows
\begin{align*}
	\epsilon_0(\mathcal{A}) \approx \frac{1}{|\mathcal{F}|}\sum_{i = 1}^{|\mathcal{F}|}\epsilon_0(\mathcal{A}_i)= 0.0607 \hspace{0.3cm}\text{and}\hspace{0.3cm}
	\epsilon_1(\mathcal{A}^c) \approx \frac{1}{|\mathcal{F}|}\sum_{i = 1}^{|\mathcal{F}|}\epsilon_1(\mathcal{A}_i^c) = 0.7389.
\end{align*}

\subsubsection{FRR versus TRR}
We present the experimental results obtained by running experiments similar to those described in~\cref{sec:experiments} considering the experimental setup in~\ref{appendix:toy_setup} in TBB. In addition to the usual discriminators, we are going to consider the optimal discriminator $D^\star$, as in~\cref{def:optgaussian}.

\begin{figure}[!htb]
\centering
    \begin{subfigure}[b]{0.32\textwidth}
    \centering
        \includegraphics[width=\textwidth]{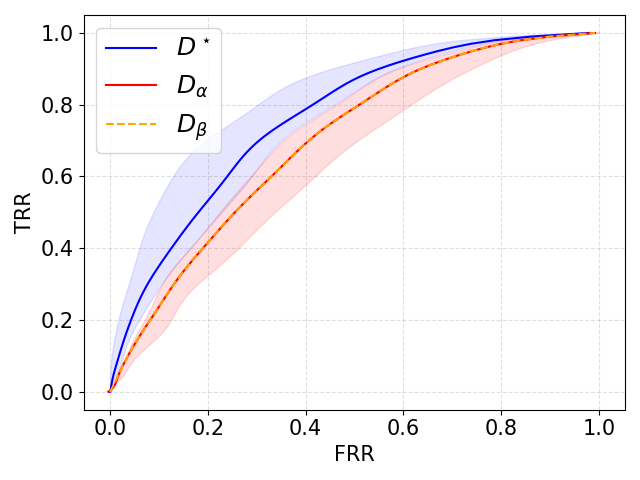}
           \caption{$\sigma=2$}
	\label{fig:pe_pe_hat_g_hat}
    \end{subfigure}
	\begin{subfigure}[b]{0.32\textwidth}
	    \centering
	    \includegraphics[width=\textwidth]{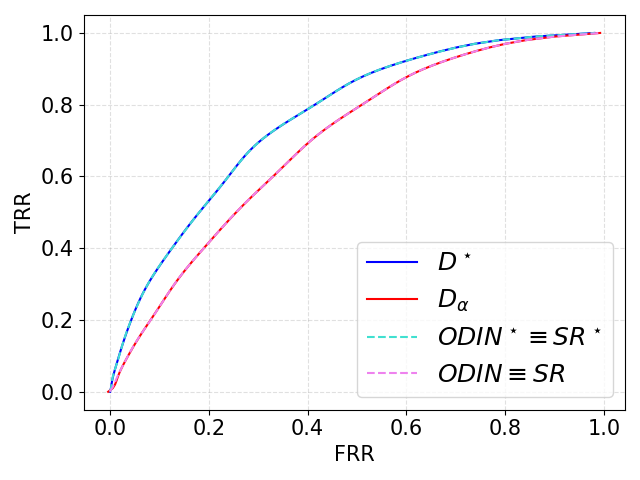}
	    \caption{$\sigma = 2$}
	    \label{fig:odin_sr}
	\end{subfigure}
		\begin{subfigure}[b]{0.32\textwidth}
	    \centering
	    \includegraphics[width=\textwidth]{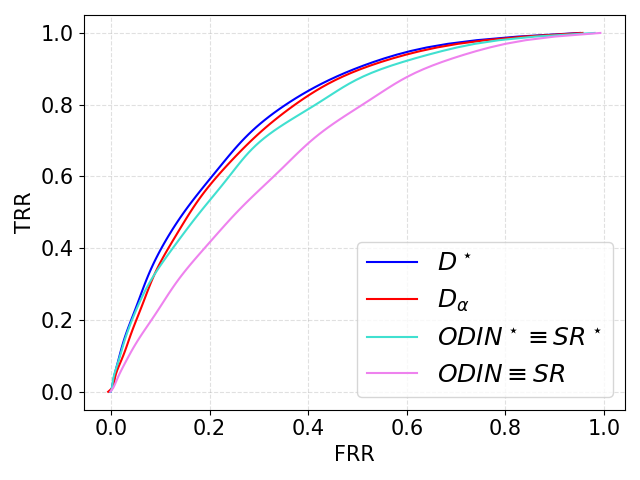}
	    \caption{$\sigma = 4$}
	    \label{fig:odin_sr_2}
	\end{subfigure}
	\caption{ROC curves for $D^\star$, $D_\alpha$ and $D_\beta$, respectively. We denote by SR$^\star$ the softmax response method based on $P_{Y|X}$. Since in this case $T=1$ and $\epsilon=0$, SR$~\equiv~$ODIN as well as  SR$^\star\equiv~$ODIN$^\star$.
    (a)  We apply each discriminator to all the classifiers in $\mathcal{F}$. The obtained ROCs are represented by the colored areas. Inside each area the mean ROC is represented by the thick line.
    Orange and red areas completely overlap as well as the mean ROC. $D^\star$ always outperforms both $D_\alpha$ and $D_\beta$ as expected.
	In (b) $D^\star$ and SR$^\star$ overlap (as also $D_\alpha$ and SR), instead in (c) where $\sigma=4$ and hence the distribution is smoother, SR discards useful information and indeed both $D^\star$ and $D_\alpha$ outperform SR.}
	\label{fig:odin_sr_main}
\end{figure}

\textbf{\textsc{Doctor}: comparison between $D^\star$, $D_\alpha$ and $D_\beta$.}
Let us present the result obtained with \textsc{Doctor} showing how $D^\star$~\eqref{eq:d-star-gaussian} works compared to $D_\alpha$ and $D_\beta$ in~\eqref{eq:d_alpha} when they have to decide whether to trust or not the decision made by a classifier. 
We test the discriminators on the dataset constructed as in~\ref{appendix:toy_setup} by considering $\sigma=2$. Let us analyze~\cref{fig:pe_pe_hat_g_hat}: we apply each discriminator to all the classifiers in $\mathcal{F}$. The obtained ROCs are represented by the colored areas. Inside each area the mean ROC is represented by the thick line.
$D_\alpha$ and $D_\beta$ reach same results as the colored areas and the thick lines are overlapped. For a given $\mathbf{x}\in\mathcal{X}$, we recall that $D^\star$ uses $\fun{Pe}(\mathbf{x})$~\eqref{eq:pe} whilst $D_\alpha$ and $D_\beta$ uses $1 - \gh(\mathbf{x})$~\eqref{eq-g_hat} and $\pehat(\mathbf{x})$~\eqref{eq:ph}, respectively. $D^\star$ always outperforms both $D_\alpha$ and $D_\beta$ since it relies on the probability of classification error based on $P_{Y|X}$ while $D_\alpha$ and $D_\beta$ use $P_{\widehat{Y}|X}$. 

\textbf{Comparison between $D^\star$, $D_\alpha$, ODIN and SR.}
We conclude this section by investigating how our competitors, namely ODIN and SR, work in this setting. 

From now on, we will put ODIN $\equiv$ SR to mean that the two methods coincide (remember we set $T=1$ and $\epsilon=0$ for all the simulations).
We show the results of the comparison in~\cref{fig:odin_sr_main}:~\cref{fig:odin_sr} considers data points from $\mathcal{N}(y\bm{\mu}, 2^2 I)$ whilst~\cref{fig:odin_sr_2} consider data points from $\mathcal{N}(y\bm{\mu}, 4^2 I)$. If in~\cref{fig:odin_sr} we cannot see an advantage in using $D_\alpha$ in place of SR, the situation is totally different
\begin{wraptable}[12]{r}{9cm}
	\caption{ AUROCs: the values for $D_\alpha$, $D_\beta$, SR, and ODIN correspond to the results for the thick lines in~\cref{fig:odin_sr_main}. $D^\star$ and ODIN$^\star\equiv\text{ }$SR$^\star$ are  obtained using $P_{Y|X}$.}
	\begin{center}
		\begin{small}
			\begin{sc}
					\begin{tabular}{l|ccccc}
						\toprule
					      &\multicolumn{5}{|c}{\textbf{AUROC $\%$}} \\
					     \hline
						\midrule
						$\sigma$ & $D^\star$  & $D_\alpha$ & $D_\beta$ & SR$~\equiv~$ ODIN  & SR$^\star\equiv~$ ODIN$^\star$\\
						\midrule
						2 & $\mathbf{76}$ & $70$ & $70$ & $70$ & $76$\\
						\midrule
						4 & $\mathbf{79}$ & $78$ & $78$ & $70$ & $76$\\
						\bottomrule
					\end{tabular}
			\end{sc}
		\end{small}
	\end{center}
	\label{table:toy_aurocs}
\end{wraptable}
in~\cref{fig:odin_sr_2}, where $D^\star$ and $D_\alpha$ clearly outperform the competitors. 
We would like to recall that \textsc{Doctor} uses all the softmax output while SR only uses the maximum value of the softmax output. Therefore, when the underlying distribution $p_{XY}$ is more smooth like in~\cref{fig:odin_sr_2}, SR discards useful information. As result, not only $D^\star$ outperforms SR$^\star$ but even $D_\alpha$ does the same. This is more evident if we look to~\cref{tab:toy_acc}, where for $\sigma=4$ we notice an improvement in terms of AUROC from $70\%$ to $78\%$ when passing from SR to $D_\alpha$.

\section{Supplementary Results of \texorpdfstring{~\Cref{sec:experiments}}{4}}
\label{appendix:experimental_setup}
\subsection{Experimental environment}
\label{appendix:enviroment}
We run each experiment on a machine equipped with an Intel(R) Xeon(R) CPU E5-2623 v4, 2.60GHz clock frequency, and a GeForce GTX 1080 Ti GPU. The execution time for the execution the tests are the following (interval size $10000$):
\begin{itemize}
    \item[TBB.] $D_\alpha$: 12.5 s. $D_\beta$: 13.6 s. SR: 15.9 s. MHLNB: 15.9 s.
    \item[PBB:] $D_\alpha$: 13 s. $D_\beta$: 25.7 s. ODIN: 14.7 s. MHLNB: 32.22 s.
\end{itemize}
\subsection{On the input pre-processing in \textsc{Doctor}}
\label{appendix:input_pre_processing}
In the following we further study \textsc{Doctor}-specific input pre-processing techniques allowed under PBB. We focus on $D_\beta$ since for $D_\alpha$ the reasoning is the same.
Formally, let $\mathbf{x}_0\in\mathcal{X}$ be a testing sample. We are looking for the minimum way to perturb the input such that the discriminator value at $\mathbf{x}_0$ is increased:
\begin{align*}
    r^* = \min_{r \text{ s.t. } \|r\|_\infty\leq\epsilon} -\log\left(\frac{\pehat(\mathbf{x}_0 + r)}{1 - \pehat(\mathbf{x}_0 + r)}\right),
\end{align*}
or equivalently, we are looking to the sample $\widetilde{\mathbf{x}}_0^\beta$ in the $\epsilon$-ball around $\mathbf{x}_0$ which maximize the discriminator value at $\widetilde{\mathbf{x}}_0^\beta$:
\begin{align*}
    \widetilde{\mathbf{x}}_0^\beta &= \mathbf{x}_0 - \epsilon\times\text{sign}\left[-\nabla_{\mathbf{x}_0}\log\left(\frac{\pehat(\mathbf{x}_0)}{1 - \pehat(\mathbf{x}_0)}\right)\right].
\end{align*}
Note that, because of~\cref{eq:pe}
\begin{align*}
    -\log\left(\frac{\pehat(\mathbf{x}_0)}{1 - \pehat(\mathbf{x}_0)}\right)  =& -\log\left(\frac{1 - P_{\widehat{Y}|X}(f_{\mathcal{D}_n}(\mathbf{x}_0)|\mathbf{x}_0)}{P_{\widehat{Y}|X}(f_{\mathcal{D}_n}(\mathbf{x}_0)|\mathbf{x}_0)}\right)\\
    =&-\log(1 - P_{\widehat{Y}|X}(f_{\mathcal{D}_n}(\mathbf{x}_0)|\mathbf{x}_0)) + \log(P_{\widehat{Y}|X}(f_{\mathcal{D}_n}(\mathbf{x}_0)|\mathbf{x}_0))\\
    =&- \log(1 - P_{\widehat{Y}|X}(f_{\mathcal{D}_n}(\mathbf{x}_0)|\mathbf{x}_0)) - \log \text{SODIN}(\mathbf{x}_0).
\end{align*}
\subsection{On the effect the intervals considered for \texorpdfstring{$\gamma$}{gamma}, \texorpdfstring{$\delta$}{delta} and \texorpdfstring{$\zeta$}{zeta} have on the AUROC computation}
\label{appendix:intervals}
\begin{table}
	\caption{AUROCs and FRR at $95\%$ TRR obtained via $D_\alpha$, $D_\beta$, ODIN, SR and MHLNB for CIFAR10 considering different size for $\Gamma_{D_\alpha \text{ or } D_\beta}$, $\Delta_{\text{ODIN or SR}}$ and $Z_{\text{MHLNB}}$ in both TBB and PBB. The column \textsc{Interval size} represents the number of equidistant values considered in the sets defined in~\eqref{alpha_interval},~\eqref{beta_interval},~\eqref{odin_interval},~\eqref{sr_interval} and in~\eqref{mhlnb_interval}, respectively.}
	\begin{center}
		\begin{small}
			\begin{sc}
			    \scalebox{0.6}{
			    	\begin{tabular}{c|c|c|c|c|c}
								\toprule
								\multirow{3}{5em}{\textbf{INTERVAL SIZE}} & \multirow{3}{5em}{\textbf{METHOD}} & \multicolumn{2}{c}{\textbf{TBB}} & \multicolumn{2}{|c}{\textbf{PBB}}\Bstrut\\\cline{3-6}
								& & \multirow{2}{4em}{AUROC} & FRR & \multirow{2}{4em}{AUROC} & FRR\Tstrut\Bstrut\\
								& & & (95 $\%$ TRR) & & (95 $\%$ TRR)\Bstrut\\
								\hline
								\midrule
								\multirow{5}{5em}{\parbox[t][][c]{3cm}{
 10}}  
& $D_\alpha$ & 69.8 & 91.6 & 77.4 & 88.4\Bstrut\\\cline{2-6}
& $D_\beta$  &  50  & 69.7 & 79.8 & 86.2\Tstrut\Bstrut\\\cline{2-6}
& ODIN       & 75.7 & 89.3 & 81.4 & 85.4\Tstrut\Bstrut\\\cline{2-6}
& SR         & 75.7 & 89.3 & - & -\Tstrut\Bstrut\\\cline{2-6}
& MHLNB      & 76.6 & 88.8 & 83.2 & 47.1 \Tstrut\\
\midrule
\multirow{5}{5em}{\parbox[t][][c]{3cm}{
 100}}  
& $D_\alpha$ & 85.1 & 80.6 & 92.5 & 42.6\Bstrut\\\cline{2-6}
& $D_\beta$  & 61.8 & 63.4 & 94.1 & 13.8 \Tstrut\Bstrut\\\cline{2-6}
& ODIN       & 88 & 73.5 & 91.5 & 49.9 \Tstrut\Bstrut\\\cline{2-6}
& SR         & 88 & 73.5 & - & -\Tstrut\Bstrut\\\cline{2-6}
& MHLNB      & 88.3 & 72.6 & 84.4 & 44.6 \Tstrut\\
\bottomrule
\end{tabular}
\begin{tabular}{c|c|c|c|c|c}
								\toprule
								\multirow{3}{5em}{\textbf{INTERVAL SIZE}} & \multirow{3}{5em}{\textbf{METHOD}} & \multicolumn{2}{c}{\textbf{TBB}} & \multicolumn{2}{|c}{\textbf{PBB}}\Bstrut\\\cline{3-6}
								& & \multirow{2}{4em}{AUROC} & FRR & \multirow{2}{4em}{AUROC} & FRR\Tstrut\Bstrut\\
								& & & (95 $\%$ TRR) & & (95 $\%$ TRR)\Bstrut\\
								\hline
								\midrule
\multirow{5}{5em}{\parbox[t][][c]{3cm}{
 1000}}  
& $D_\alpha$ & 91.3 & 53.1 & 94.7 & 13.8 \Bstrut\\\cline{2-6}
& $D_\beta$  & 66.5 & 48.3 & 94.8 & 13.4 \Tstrut\Bstrut\\\cline{2-6}
& ODIN       & 92.5 & 28.9 & 94 & 18.3 \Tstrut\Bstrut\\\cline{2-6}
& SR         & 92.5 & 28.9 & - & -\Tstrut\Bstrut\\\cline{2-6}
& MHLNB      & 92.2 & 35.3 & 84.4 & 44.5 \Tstrut\\
\midrule
\multirow{5}{5em}{\parbox[t][][c]{3cm}{
 10000}}  
& $D_\alpha$ & 93.7 & 18.4 & 95.2 & 13.9 \Bstrut\\\cline{2-6}
& $D_\beta$  & 68.5 & 18.6 & 94.8 & 13.4\Tstrut\Bstrut\\\cline{2-6}
& ODIN       & 93.9 & 18 & 94.2 & 18.4\Tstrut\Bstrut\\\cline{2-6}
& SR         & 93.9 & 18 & - & -\Tstrut\Bstrut\\\cline{2-6}
& MHLNB      & 92.1 & 31 & 84.4 & 44.6 \Tstrut\\
						\bottomrule
					\end{tabular}
					}%
			\end{sc}
		\end{small}
	\end{center}
	\label{table:gamma}
\end{table}
Let us consider the AUROC as a performance measure for the discriminators. The computation of the AUROC of $D_\alpha$, as well as those of ODIN and SR,  heavily depend on the choice of the range values for the decision region thresholds. In the following paragraph, we will discuss how we chose these ranges, namely $\gamma\in\Gamma_{D_\alpha \text{ or } D_\beta}\subseteq\mathbb{R}$, $\delta\in\Delta_{\text{ODIN or SR}}\subseteq[0,1]$ and $\zeta\in Z_{\text{MHLNB}}\subseteq\mathbb{R}$.
In the experiments of~\cref{sec:experiments}, 
we therefore proceed by fixing the aforementioned ranges as follows:
\begin{align}
\Gamma_{D_\alpha}&\myeq\left[\underset{(\mathbf{x},y)\in\mathcal{T}_m}{\text{min}}\frac{1 - \gh(\mathbf{x})}{\gh(\mathbf{x})}, \underset{(\mathbf{x},y)\in\mathcal{T}_m}{\text{max}}\frac{1 - \gh(\mathbf{x})}{\gh(\mathbf{x})}\right],\label{alpha_interval}\\
\Gamma_{D_\beta}&\myeq\left[\underset{(\mathbf{x},y)\in\mathcal{T}_m}{\text{min}}\frac{\pehat(\mathbf{x})}{1-\pehat(\mathbf{x})}, \underset{(\mathbf{x},y)\in\mathcal{T}_m}{\text{max}}\frac{\pehat(\mathbf{x})}{1 -\pehat(\mathbf{x})}\right],\label{beta_interval}\\
\Delta_{\text{ODIN}}&\myeq\left[\underset{(\mathbf{x},y)\in\mathcal{T}_m}{\text{min}}\text{SODIN}(\mathbf{x}), \underset{(\mathbf{x},y)\in\mathcal{T}_m}{\text{max}}\text{SODIN}(\mathbf{x})\right],\label{odin_interval}\\
\Delta_{\text{SR}}&\myeq\left[\underset{(\mathbf{x},y)\in\mathcal{T}_m}{\text{min}}\text{SR}(\mathbf{x}), \underset{(\mathbf{x},y)\in\mathcal{T}_m}{\text{max}}\text{SR}(\mathbf{x})\right],\label{sr_interval}\\
Z_{\text{MHLNB}}&\myeq\left[\underset{(\mathbf{x},y)\in\mathcal{T}_m}{\text{min}}\text{M}(\mathbf{x}), \underset{(\mathbf{x},y)\in\mathcal{T}_m}{\text{max}}\text{M}(\mathbf{x})\right].\label{mhlnb_interval}
\end{align}
Secondly, we fix the number of values to consider in $\Gamma_{D_\alpha \text{ or } D_\beta}$, $\Delta_{\text{ODIN or SR}}$ and $Z_{\text{MHLNB}}$: we test the AUROCs for CIFAR10 for different values of the size of $\Gamma_{D_\alpha \text{ or } D_\beta}$, $\Delta_{\text{ODIN or SR}}$ and $Z_{\text{MHLNB}}$ in both TBB and PBB scenarios. The results are collected in~\cref{table:gamma}. Let us denote by $I$ a generic interval between the ones of~\cref{alpha_interval},~\cref{beta_interval},~\cref{odin_interval},~\cref{sr_interval} and~\cref{mhlnb_interval}, throughout the experiments we set the size of $I$ to $(\max I - \min I)*10000$.
\begin{figure*}[!htb]
	\centering
	\begin{subfigure}[b]{ 0.23\textwidth}
	    \centering
	    \includegraphics[width=\textwidth]{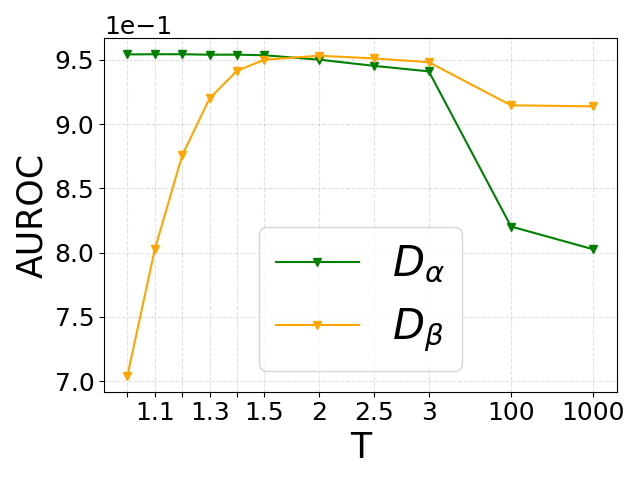}
	    \caption{CIFAR10\\\centering{$\epsilon = 0.0003$}}
	    \label{fig:cifar10_best_T}
	\end{subfigure}
		\begin{subfigure}[b]{ 0.23\textwidth}
	    \centering
	    \includegraphics[width=\textwidth]{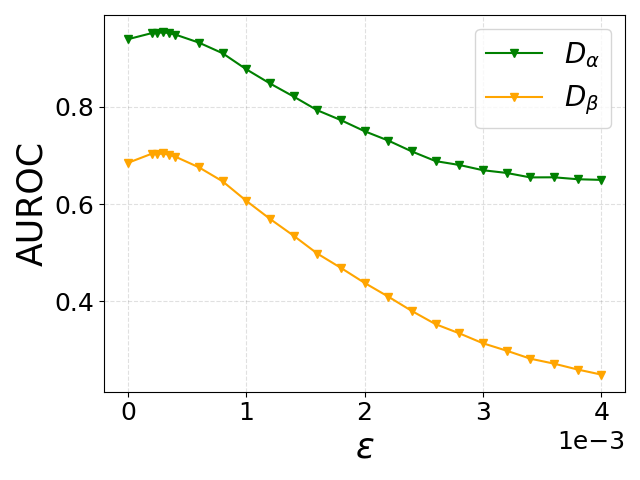}
	    \caption{CIFAR10\\\centering{$T = 1$}}
	    \label{fig:cifar10_best_eps_alpha}
	\end{subfigure}
		\begin{subfigure}[b]{ 0.23\textwidth}
	    \centering
	    \includegraphics[width=\textwidth]{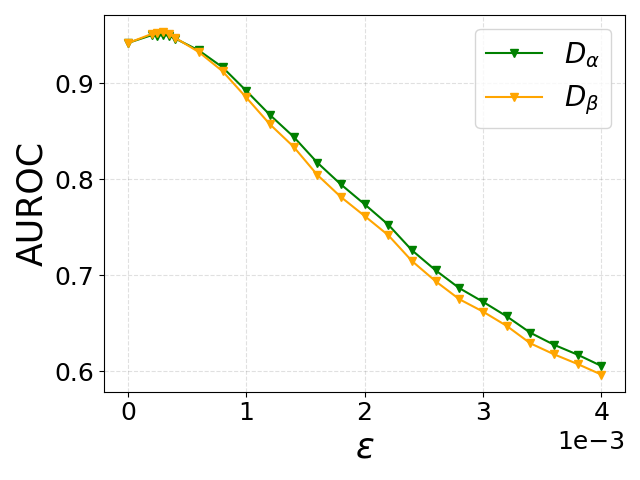}
	    \caption{CIFAR10\\\centering{$T = 2$}}
	    \label{fig:cifar10_best_eps_beta}
	\end{subfigure}
	\begin{subfigure}[b]{ 0.23\textwidth}
	    \centering
	    \includegraphics[width=\textwidth]{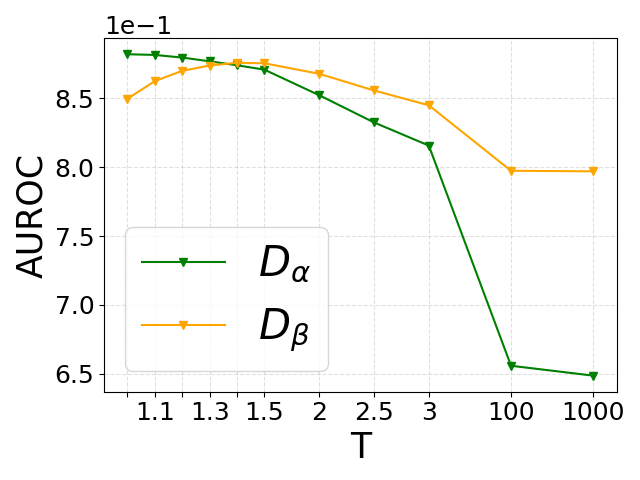}
	    \caption{CIFAR100\\\centering{$\epsilon = 0.0003$}}
	    \label{fig:cifar100_best_T}
	\end{subfigure}
	\begin{subfigure}[b]{ 0.23\textwidth}
	    \centering
	    \includegraphics[width=\textwidth]{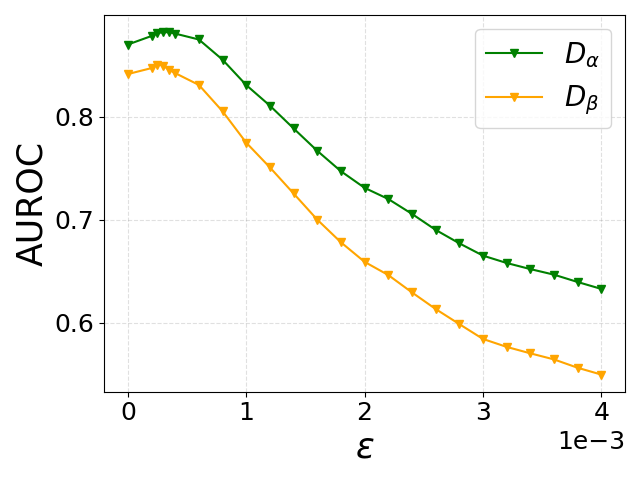}
	    \caption{CIFAR100\\\centering{$T = 1$}}
	    \label{fig:cifar100_best_eps_alpha}
	\end{subfigure}
	\begin{subfigure}[b]{ 0.23\textwidth}
	    \centering
	    \includegraphics[width=\textwidth]{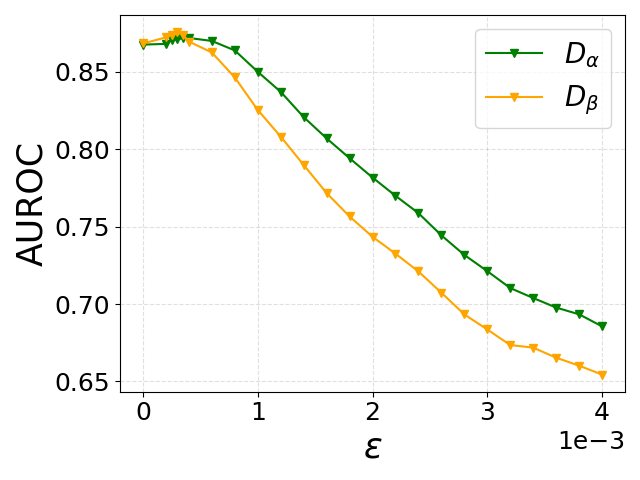}
	    \caption{CIFAR100\\\centering{$T = 1.5$}}
	    \label{fig:cifar100_best_eps_beta}
	\end{subfigure}
	\begin{subfigure}[b]{ 0.23\textwidth}
	    \centering
	    \includegraphics[width=\textwidth]{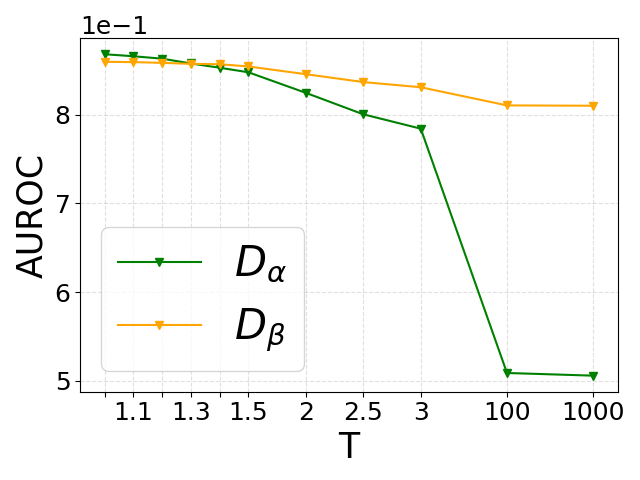}
	    \caption{TinyImageNet\\\centering{$\epsilon = 0.0006$}}
	    \label{fig:tiny_best_T}
	\end{subfigure}
	\begin{subfigure}[b]{ 0.23\textwidth}
	    \centering
	    \includegraphics[width=\textwidth]{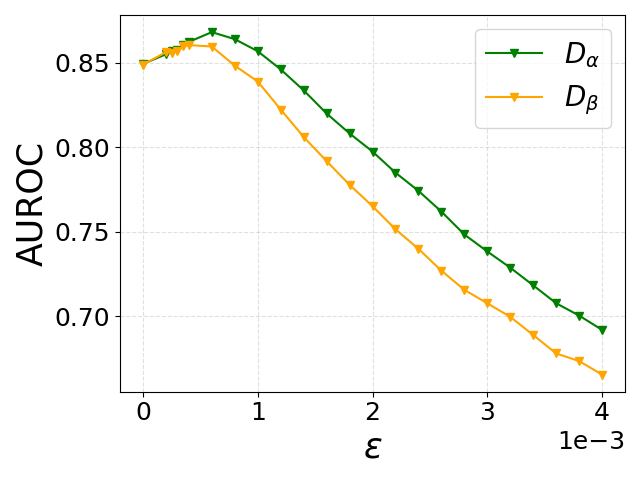}
	    \caption{TinyImageNet\\\centering{$T = 1$}}
	    \label{fig:tiny_best_eps_alpha}
	\end{subfigure}
	\begin{subfigure}[b]{ 0.23\textwidth}
	    \centering
	    \includegraphics[width=\textwidth]{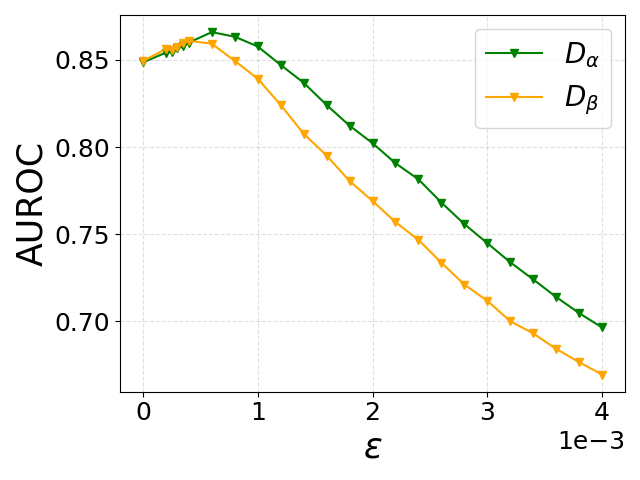}
	    \caption{TinyImageNet\\\centering{$T = 1.1$}}
	    \label{fig:tiny_best_eps_beta}
	\end{subfigure}
	\begin{subfigure}[b]{ 0.23\textwidth}
	    \centering
	    \includegraphics[width=\textwidth]{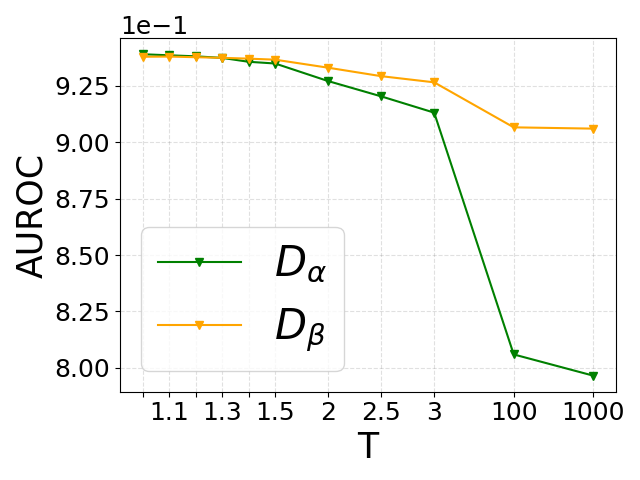}
	    \caption{SVHN\\\centering{$\epsilon = 0.001$}}
	    \label{fig:svhn_best_T}
	\end{subfigure}
	\begin{subfigure}[b]{ 0.23\textwidth}
	    \centering
	    \includegraphics[width=\textwidth]{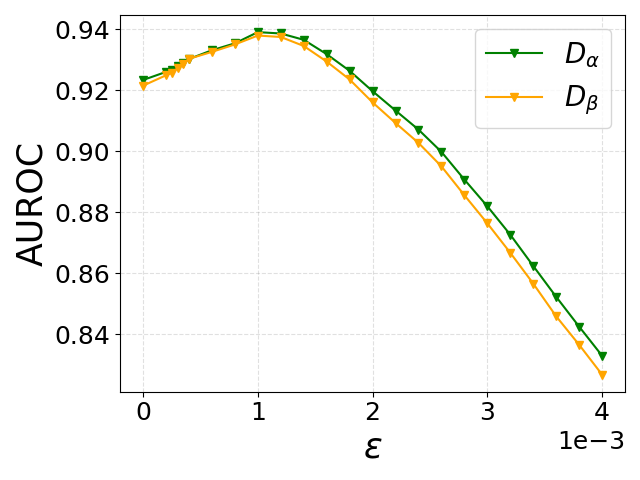}
	    \caption{SVHN\\\centering{$T = 1$}}
	    \label{fig:svhn_best_eps_alpha}
	\end{subfigure}
		\begin{subfigure}[b]{ 0.23\textwidth}
	    \centering
	    \includegraphics[width=\textwidth]{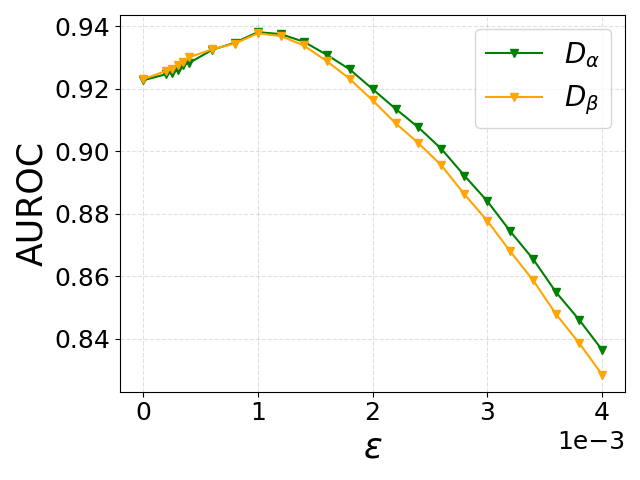}
	    \caption{SVHN\\\centering{$T = 1.2$}}
	    \label{fig:svhn_best_eps_beta}
	\end{subfigure}
	\caption{\FedRev{Comparison of AUROCs obtained via $D_\alpha$ (in green) and via $D_\beta$ (in orange) for different values of $T$ and $\epsilon$.}}
	\label{fig:auc}
\end{figure*}
\subsection{Additional plots and results}
In the next sections, we show graphically the set of results obtained from the experiments in~\cref{subsec:exp_res}. We first specify the range of values for the parameters $T$ and $\epsilon$ considered throughout the experiments. For temperature scaling, $T$ is selected among  $\{1,~ 1.1,~1.2,$ $1.3,~1.4,~1.5,$ $2,~2.5,~3,$ $100,~1000\}$, whilst for input pre-processing, $\epsilon$ is selected among $\{0,~.0002,~.00025,$ $.0003,~.00035,~.0004,$ $.0006,~.0008,~.001,$ $.0012,~.0014,~.0016,$ $.0018, .002,~.0022,$ $.0024,~.0026,~.0028,$ $.003,~.0032,~.0034,$ $.0036,~.0038,~.004\}$.
\subsubsection{Comparison \texorpdfstring{$D_\alpha$}{D-alpha} and \texorpdfstring{$D_\beta$}{D-beta}}
\label{appendix:alpha_beta}
We include the plots for \textit{\textsc{Doctor}: comparison between $D_\alpha$ and $D_\beta$} (\cref{subsec:exp_res}).
In~\cref{fig:cifar10_best_T}, \cref{fig:cifar100_best_T},~\cref{fig:tiny_best_T} and~\cref{fig:svhn_best_T}, we set $\epsilon$ at its best value which is found to coincide in the case of $D_\alpha$ and $D_\beta$. In~\cref{fig:cifar10_best_eps_alpha},~\cref{fig:cifar100_best_eps_alpha},~\cref{fig:tiny_best_eps_alpha} and~\cref{fig:svhn_best_eps_alpha} we do the opposite and we set $T$ to its best value w.r.t. $D_\alpha$ whilst in \cref{fig:cifar10_best_eps_beta},~\cref{fig:cifar100_best_eps_beta},~\cref{fig:tiny_best_eps_beta} and~\cref{fig:svhn_best_eps_beta}, the value of $T$ is chosen w.r.t. the best value for $D_\beta$.
\subsubsection{Comparison \texorpdfstring{$D_\alpha$}{D-alpha}, \texorpdfstring{$D_\beta$}{D-beta}, ODIN and MHLNB}
\label{appendix:overall}
\begin{figure*}[!htb]
	\centering
	\begin{subfigure}[b]{ 0.23\textwidth}
	    \centering
	    \includegraphics[width=\textwidth]{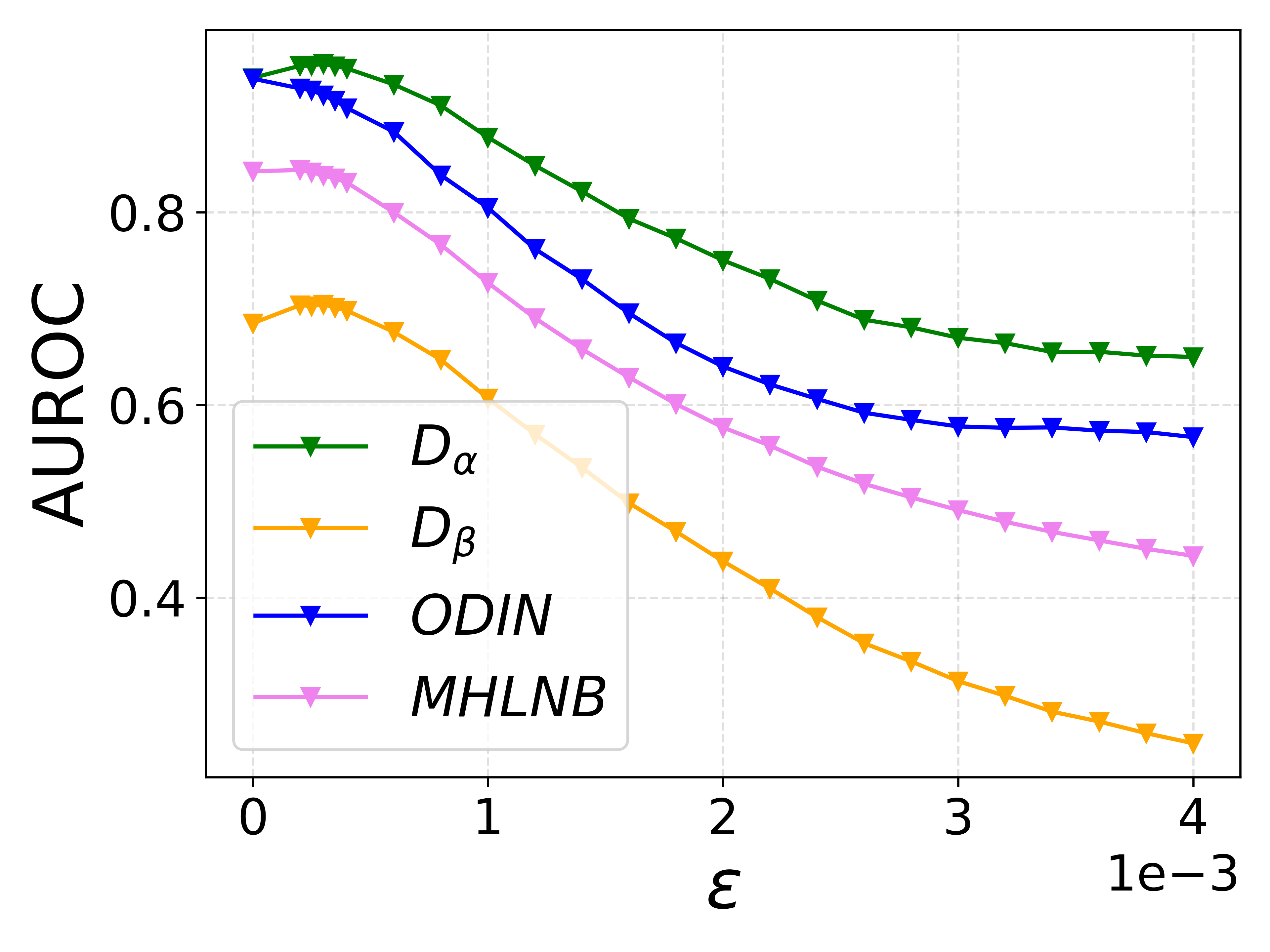}
	    \vspace{-1.5\baselineskip}
	    \caption{CIFAR10\\\centering{$T=1$}}
	    \label{fig:cifar10_T_1_all}
	\end{subfigure}
	\begin{subfigure}[b]{ 0.23\textwidth}
	    \centering
	    \includegraphics[width=\textwidth]{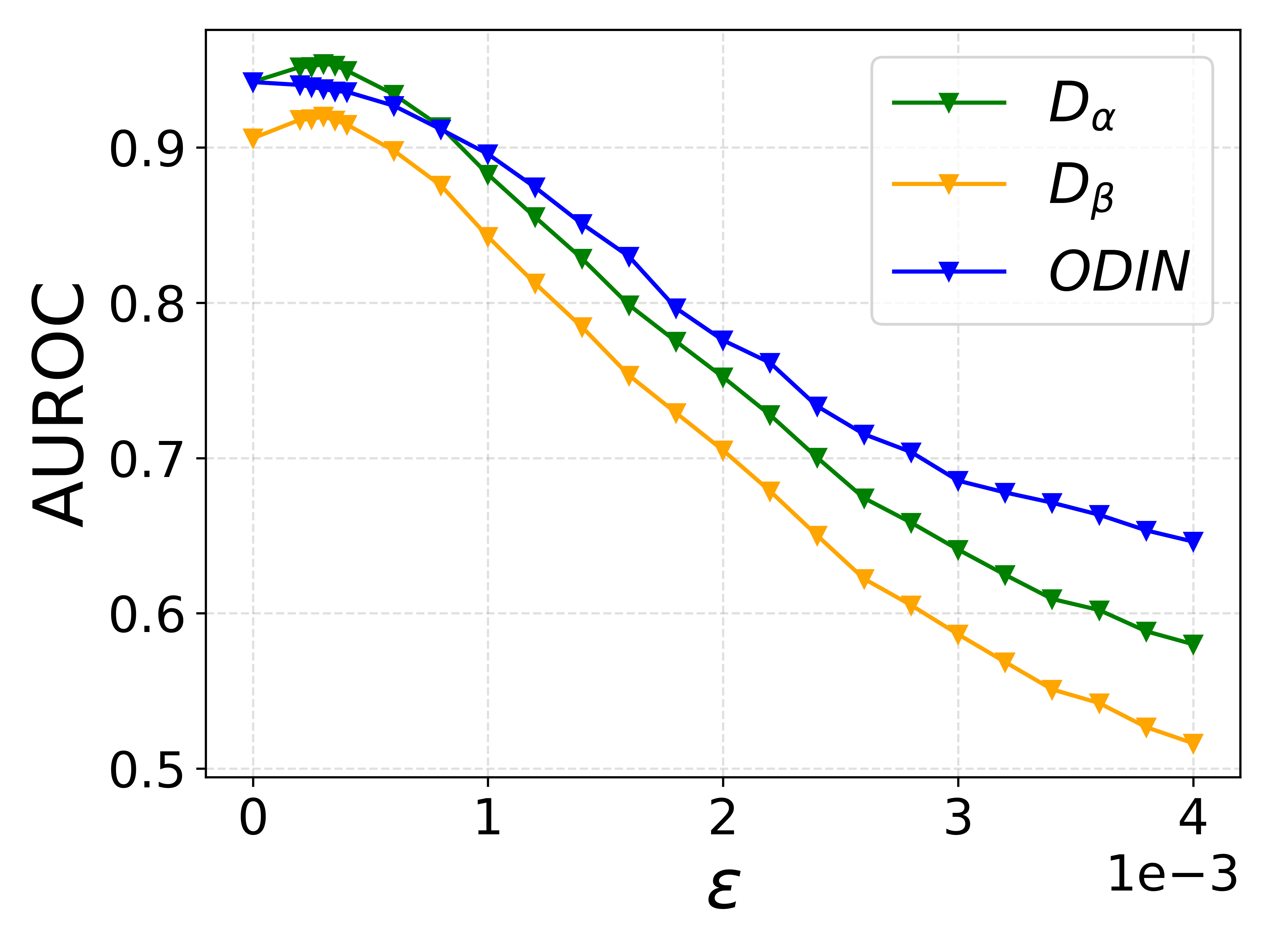}
	    \vspace{-1.5\baselineskip}
	    \caption{CIFAR10\\\centering{$T=1.3$}}
	    \label{fig:cifar10_T_1.3_all}
	\end{subfigure}
	\begin{subfigure}[b]{ 0.23\textwidth}
	    \centering
	    \includegraphics[width=\textwidth]{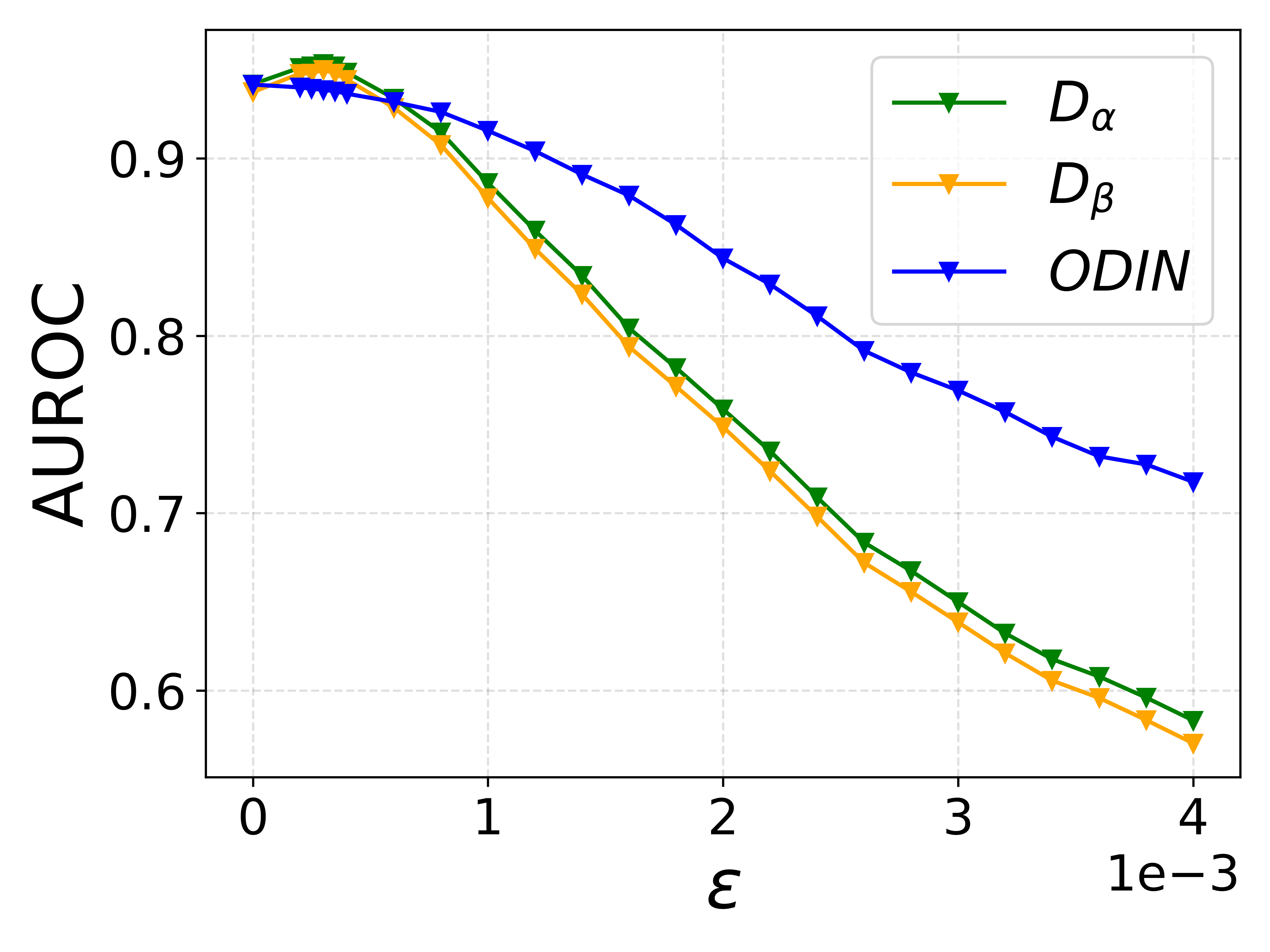}
	    \vspace{-1.5\baselineskip}
	    \caption{CIFAR10\\\centering{$T=1.5$}}
	    \label{fig:cifar10_T_1.5_all}
	\end{subfigure}
	\begin{subfigure}[b]{ 0.23\textwidth}
	    \centering
	    \includegraphics[width=\textwidth]{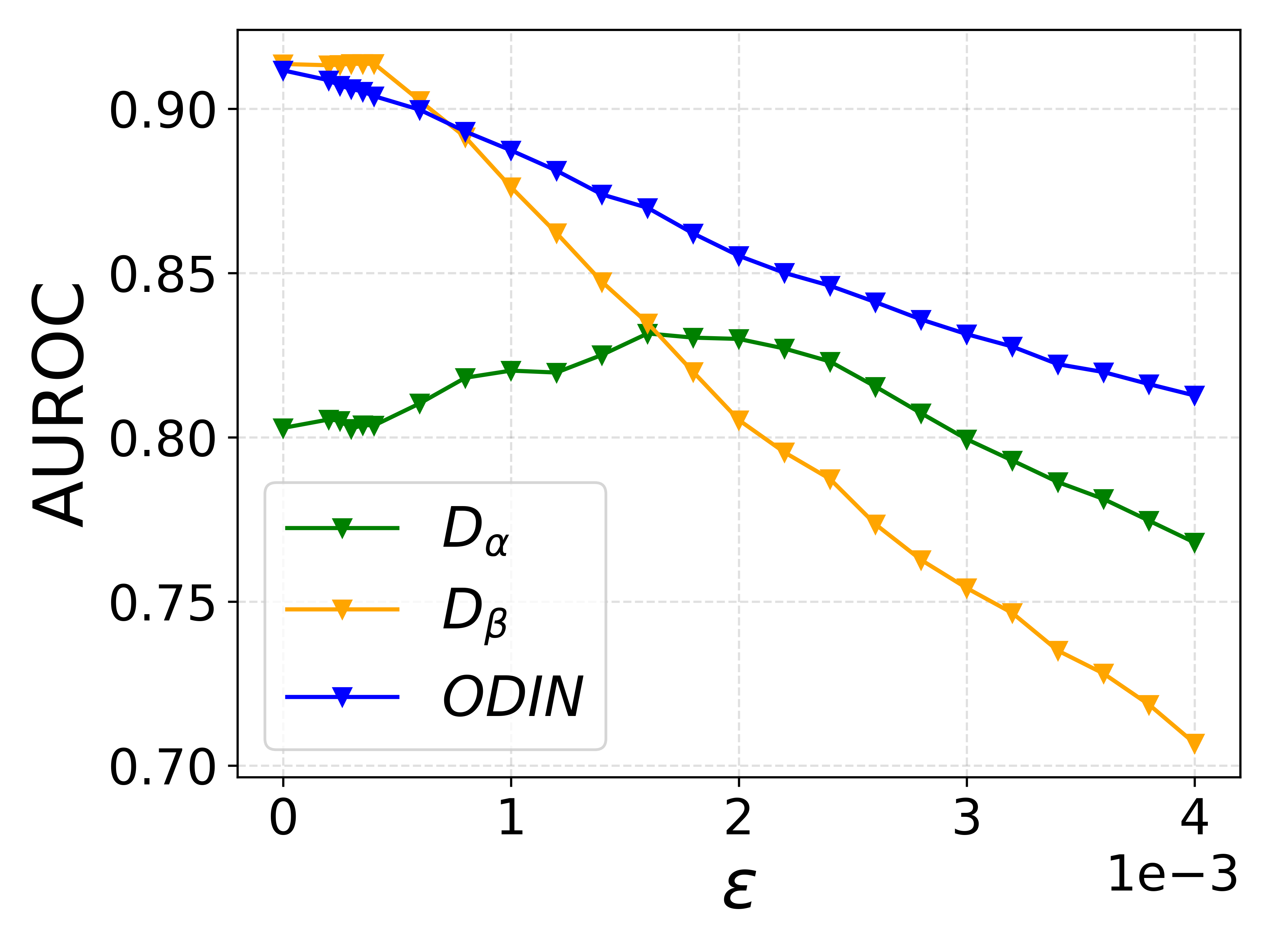}
	    \vspace{-1.5\baselineskip}
	    \caption{CIFAR10\\ \centering{$T=1000$}}
	    \label{fig:cifar10_T_1000_all}
	\end{subfigure}
	\begin{subfigure}[b]{  0.23\textwidth}
	    \centering
	    \includegraphics[width=\textwidth]{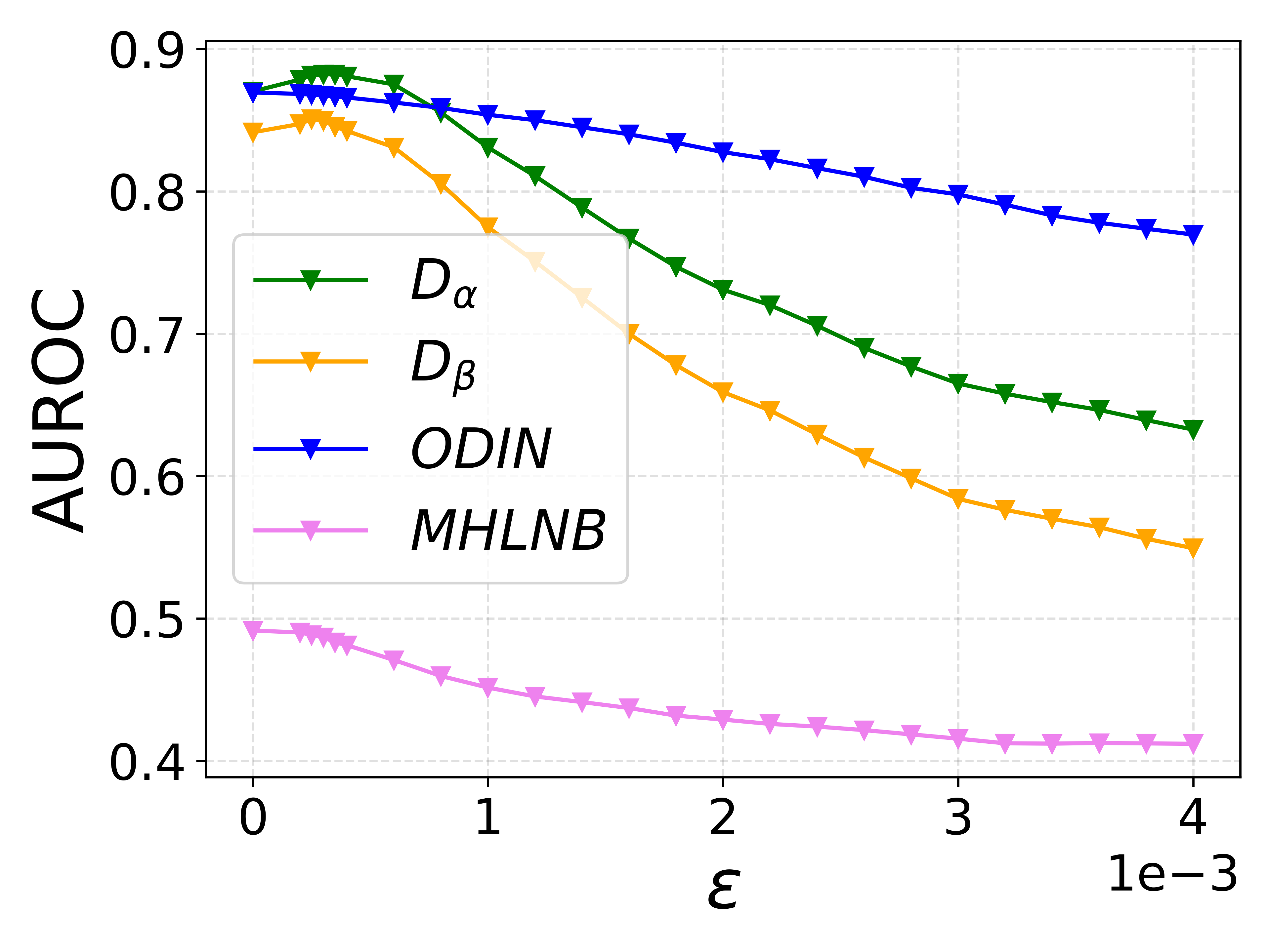}
	    \vspace{-1.5\baselineskip}
	    \caption{CIFAR100\\  \centering{$T=1$}}
	    \label{fig:cifar100_T_1_all}
	\end{subfigure}
	\begin{subfigure}[b]{  0.23\textwidth}
	    \centering
	    \includegraphics[width=\textwidth]{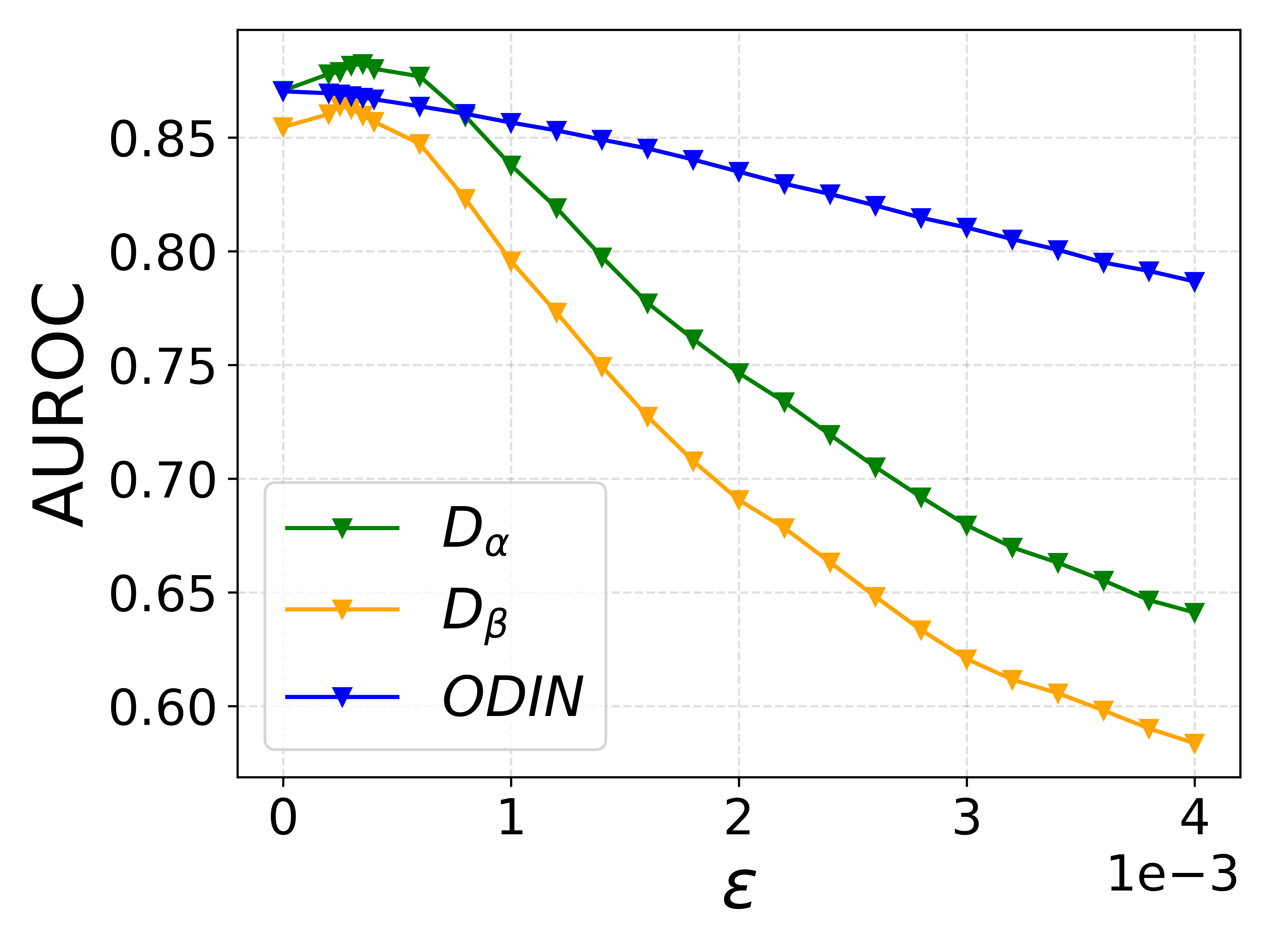}
	    \vspace{-1.5\baselineskip}
	    \caption{CIFAR100\\  \centering{$T=1.3$}}
	    \label{fig:cifar100_T_1.3_all}
	\end{subfigure}
	\begin{subfigure}[b]{  0.23\textwidth}
	    \centering
	    \includegraphics[width=\textwidth]{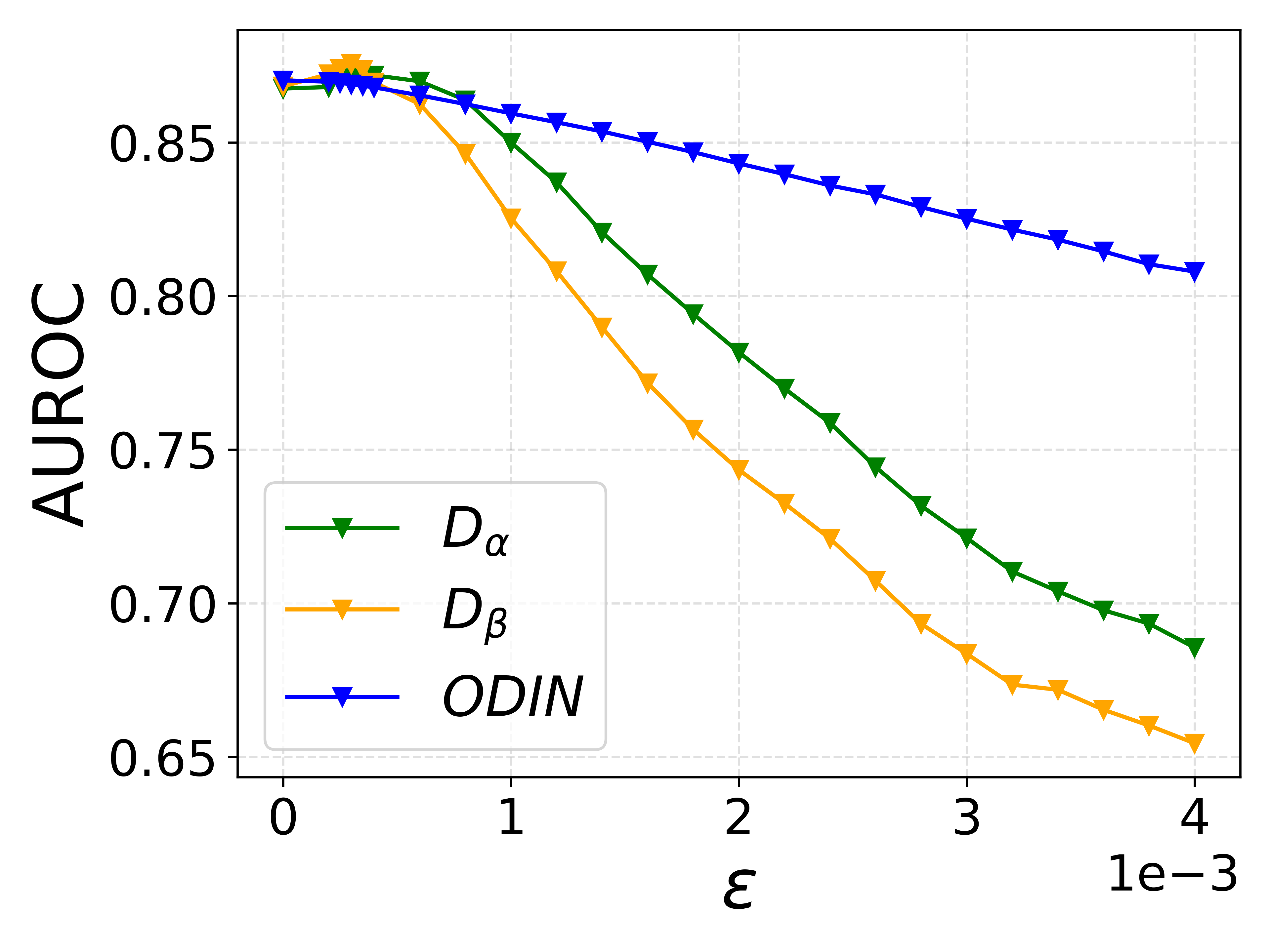}
	    \vspace{-1.5\baselineskip}
	    \caption{CIFAR100\\  \centering{$T=1.5$}}
	    \label{fig:cifar100_T_1.5_all}
	\end{subfigure}
	\begin{subfigure}[b]{  0.23\textwidth}
	    \centering
	    \includegraphics[width=\textwidth]{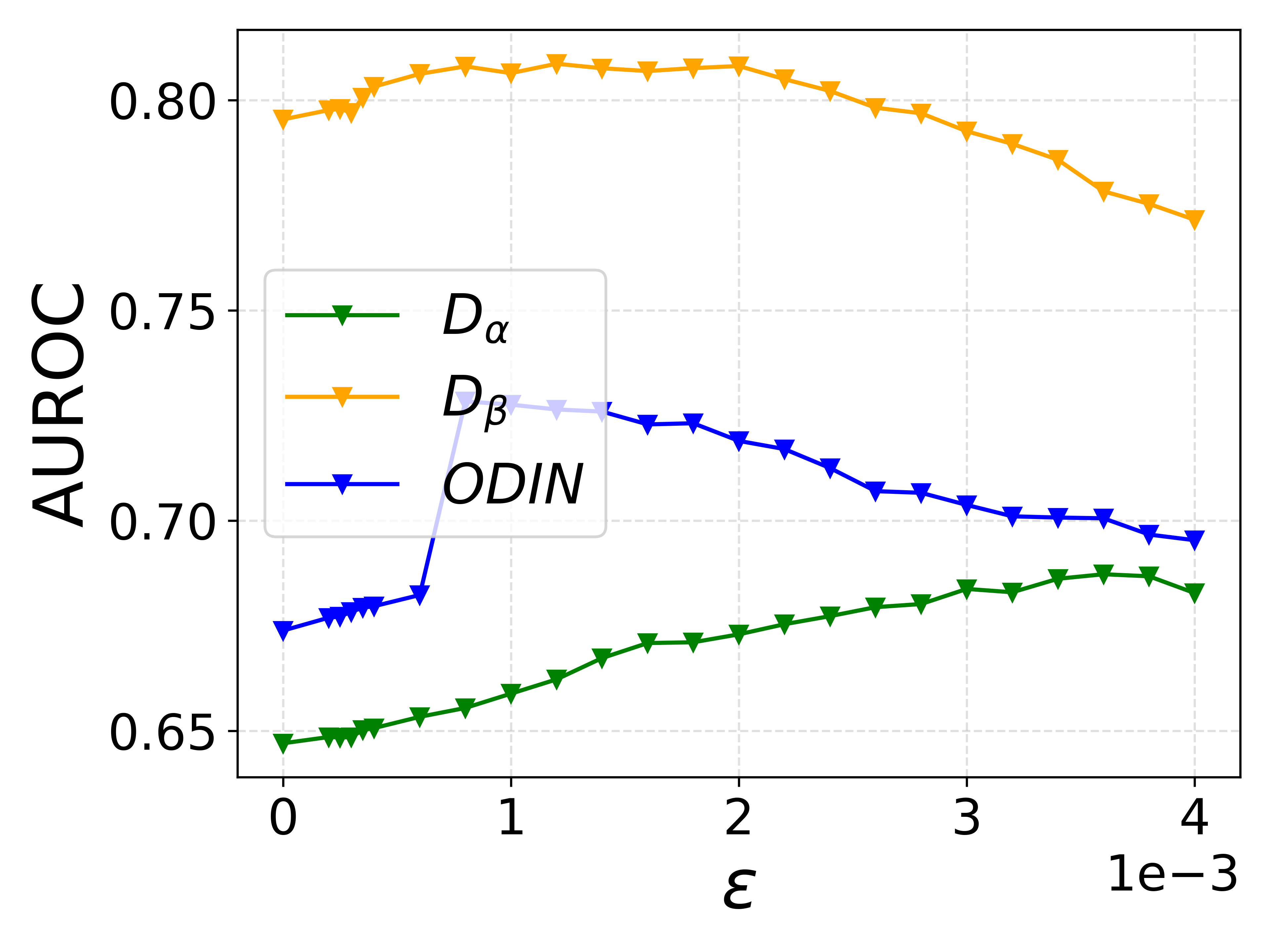}
	    \vspace{-1.5\baselineskip}
	    \caption{CIFAR100\\\centering{$T=1000$}}
	    \label{fig:cifar100_T_1000_all}
	\end{subfigure}
		\begin{subfigure}[b]{ 0.23\textwidth}
	    \centering
	    \includegraphics[width=\textwidth]{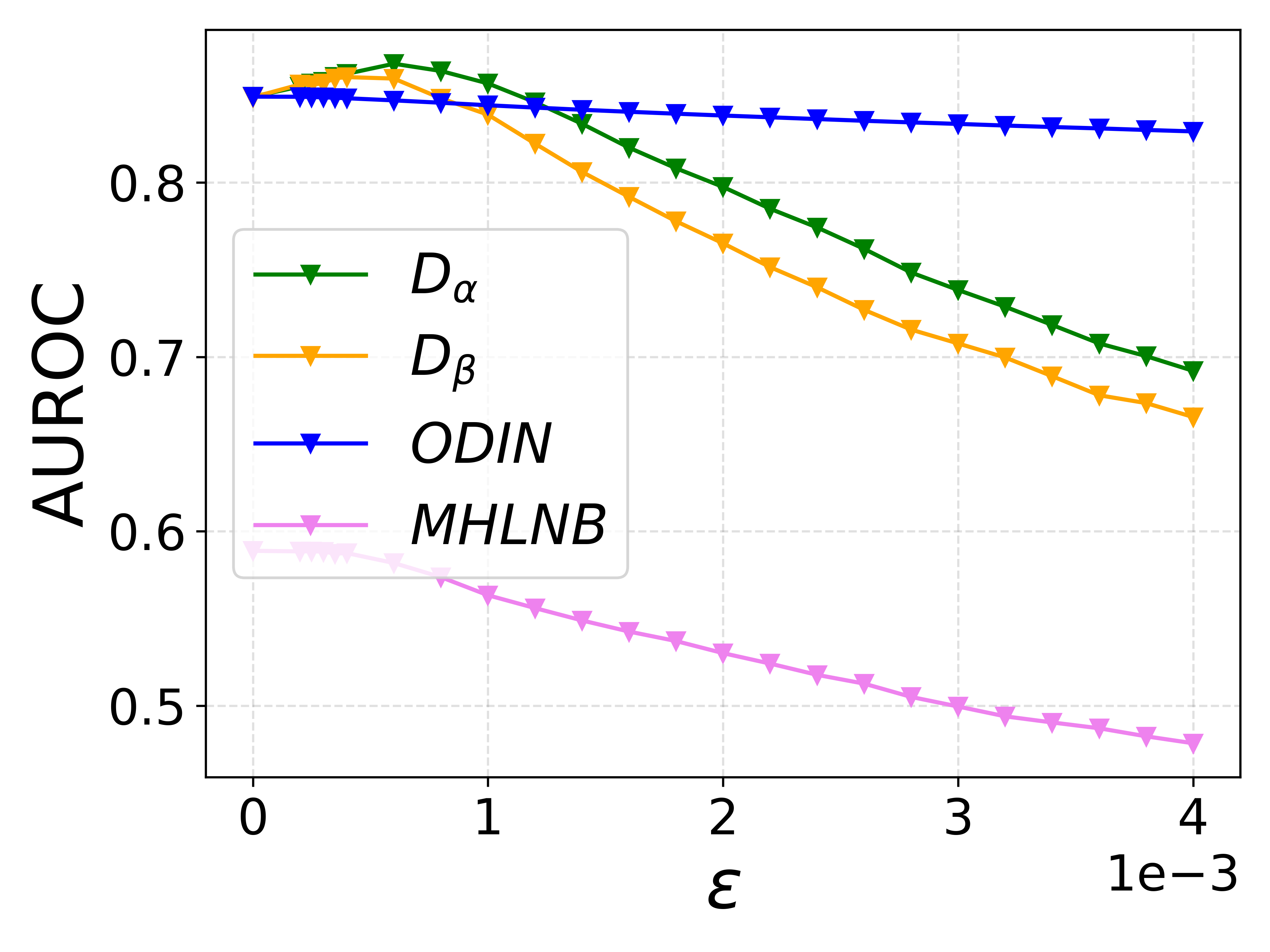}
	    \vspace{-1.5\baselineskip}
	    \caption{TinyImageNet\\\centering{$T=1$}}
	    \label{fig:tinyimagenet_T_1_all}
	\end{subfigure}
	\begin{subfigure}[b]{ 0.23\textwidth}
	    \centering
	    \includegraphics[width=\textwidth]{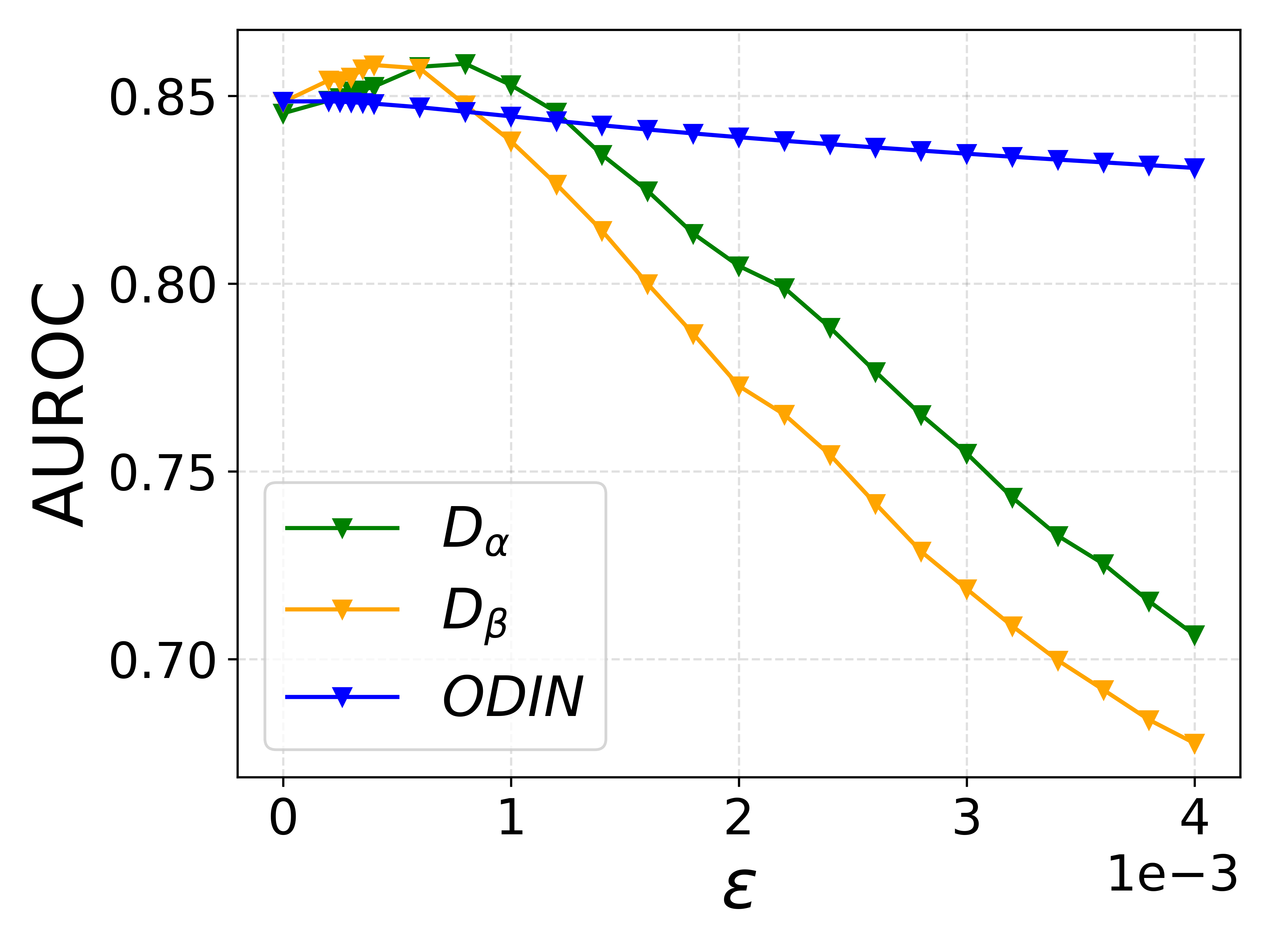}
	    \vspace{-1.5\baselineskip}
	    \caption{TinyImageNet\\\centering{$T=1.3$}}
	    \label{fig:tinyimagenet_T_1.3_all}
	\end{subfigure}
	\begin{subfigure}[b]{ 0.23\textwidth}
	    \centering
	    \includegraphics[width=\textwidth]{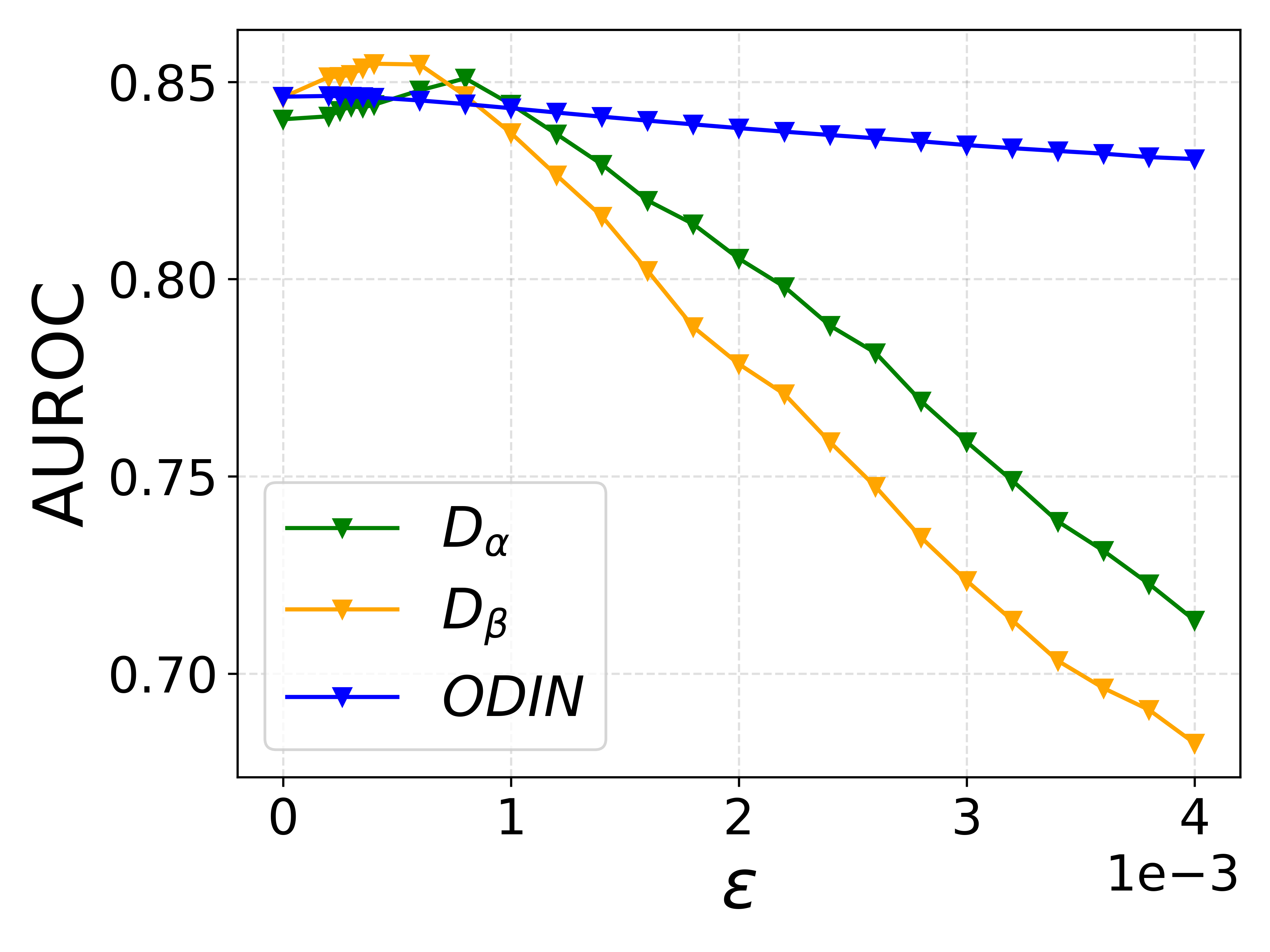}
	    \vspace{-1.5\baselineskip}
	    \caption{TinyImageNet\\\centering{$T=1.5$}}
	    \label{fig:tinyimagenet_T_1.5_all}
	\end{subfigure}
	\begin{subfigure}[b]{ 0.23\textwidth}
	    \centering
	    \includegraphics[width=\textwidth]{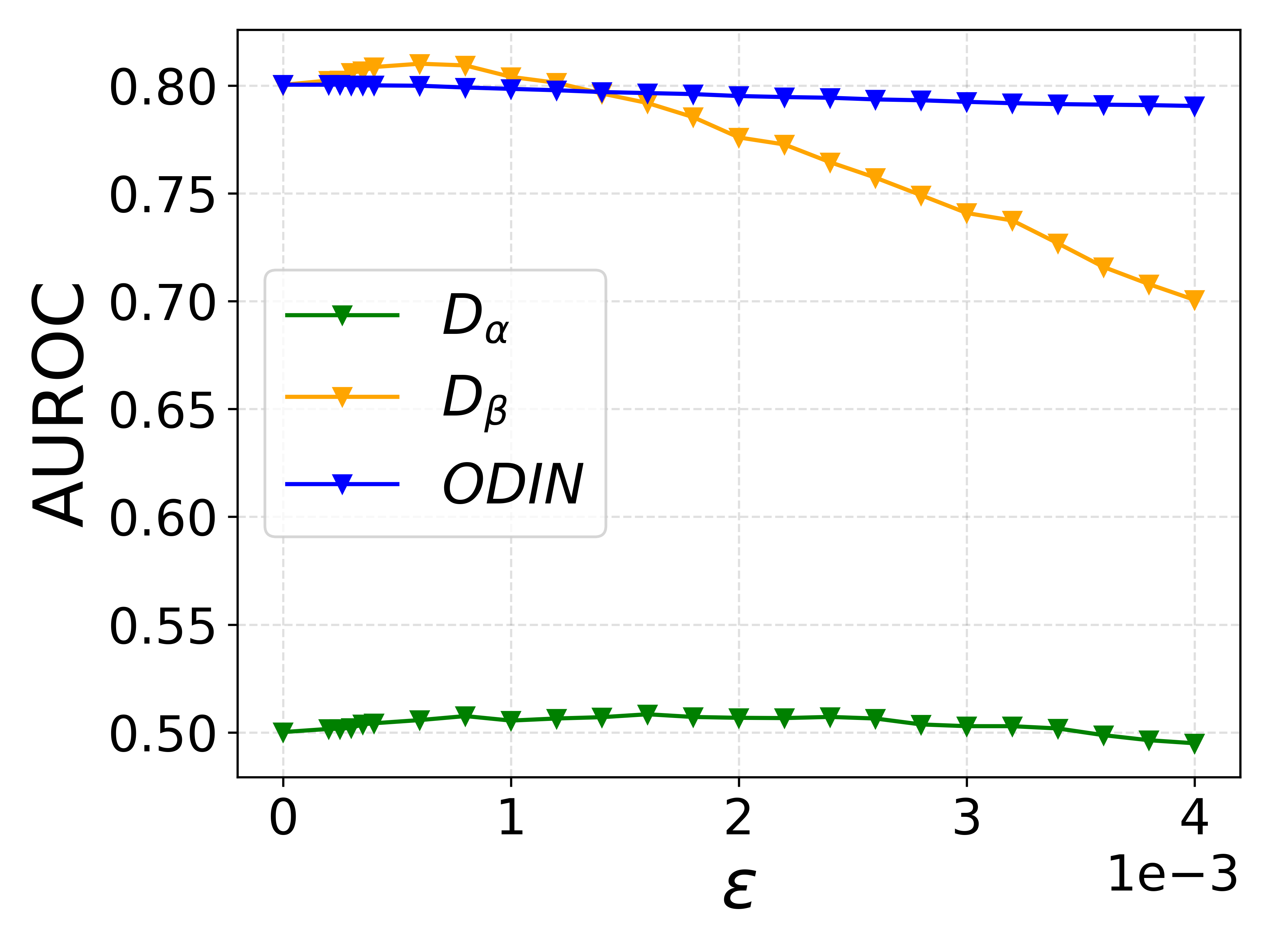}
	    \vspace{-1.5\baselineskip}
	    \caption{TinyImageNet\\\centering{$T=1000$}}
	    \label{fig:tinyimagenet_T_1000_all}
	\end{subfigure}
		\begin{subfigure}[b]{ 0.23\textwidth}
	    \centering
	    \includegraphics[width=\textwidth]{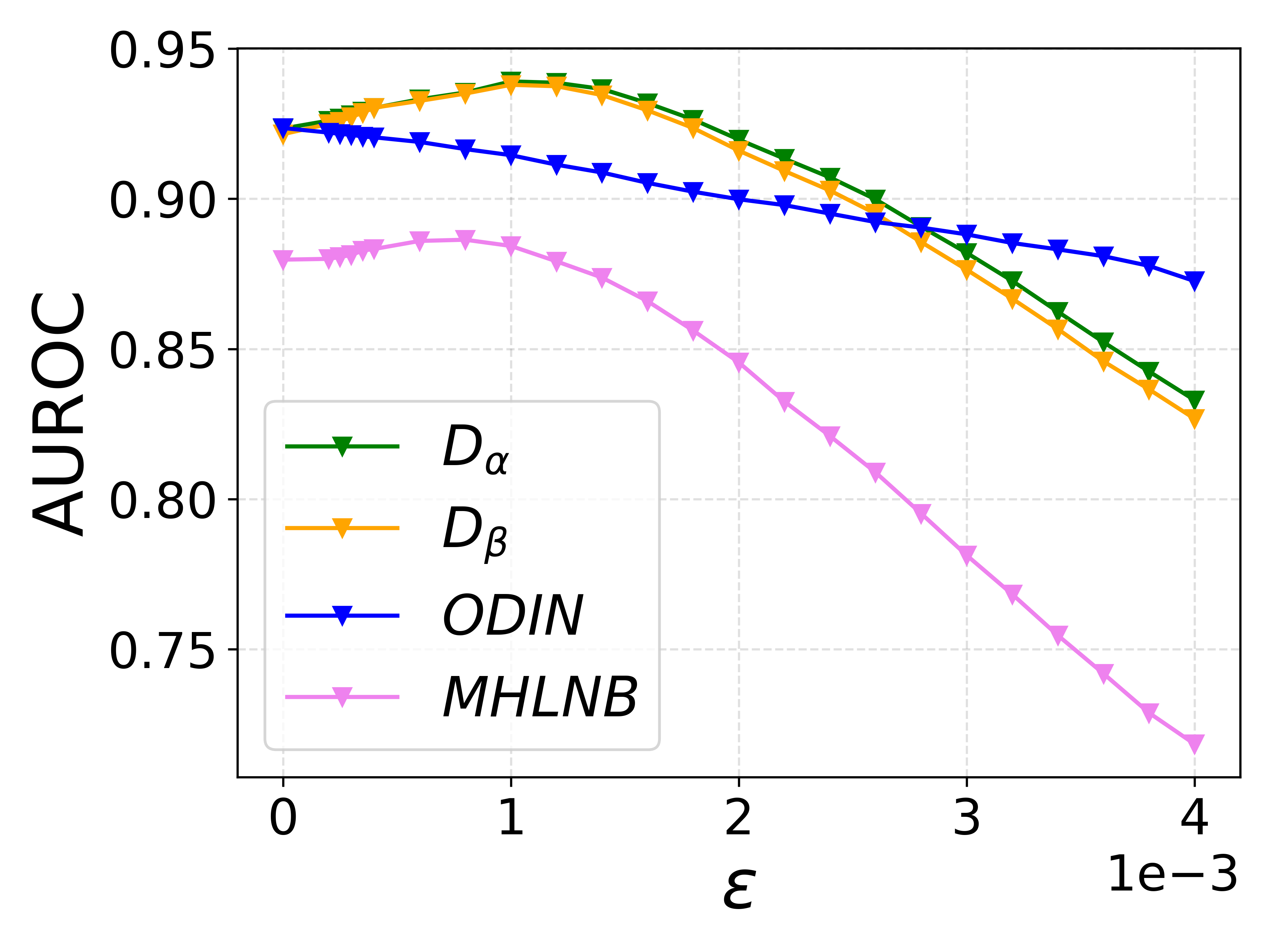}
	    \vspace{-1.5\baselineskip}
	    \caption{SVHN\\\centering{$T=1$}}
	    \label{fig:svhn_T_1_all}
	\end{subfigure}
	\begin{subfigure}[b]{ 0.23\textwidth}
	    \centering
	    \includegraphics[width=\textwidth]{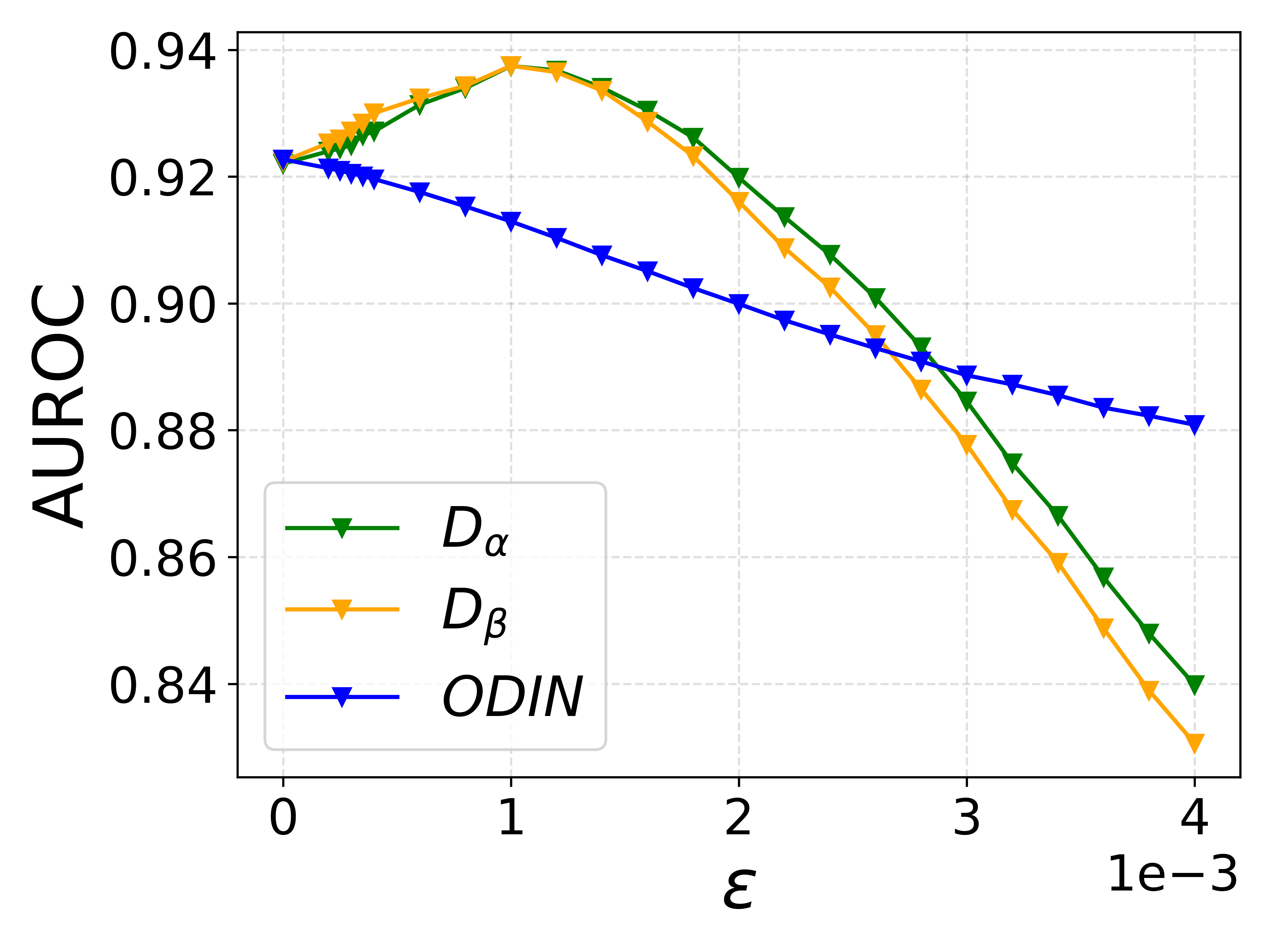}
	    \vspace{-1.5\baselineskip}
	    \caption{SVHN\\\centering{$T=1.3$}}
	    \label{fig:svhn_T_1.3_all}
	\end{subfigure}
	\begin{subfigure}[b]{ 0.23\textwidth}
	    \centering
	    \includegraphics[width=\textwidth]{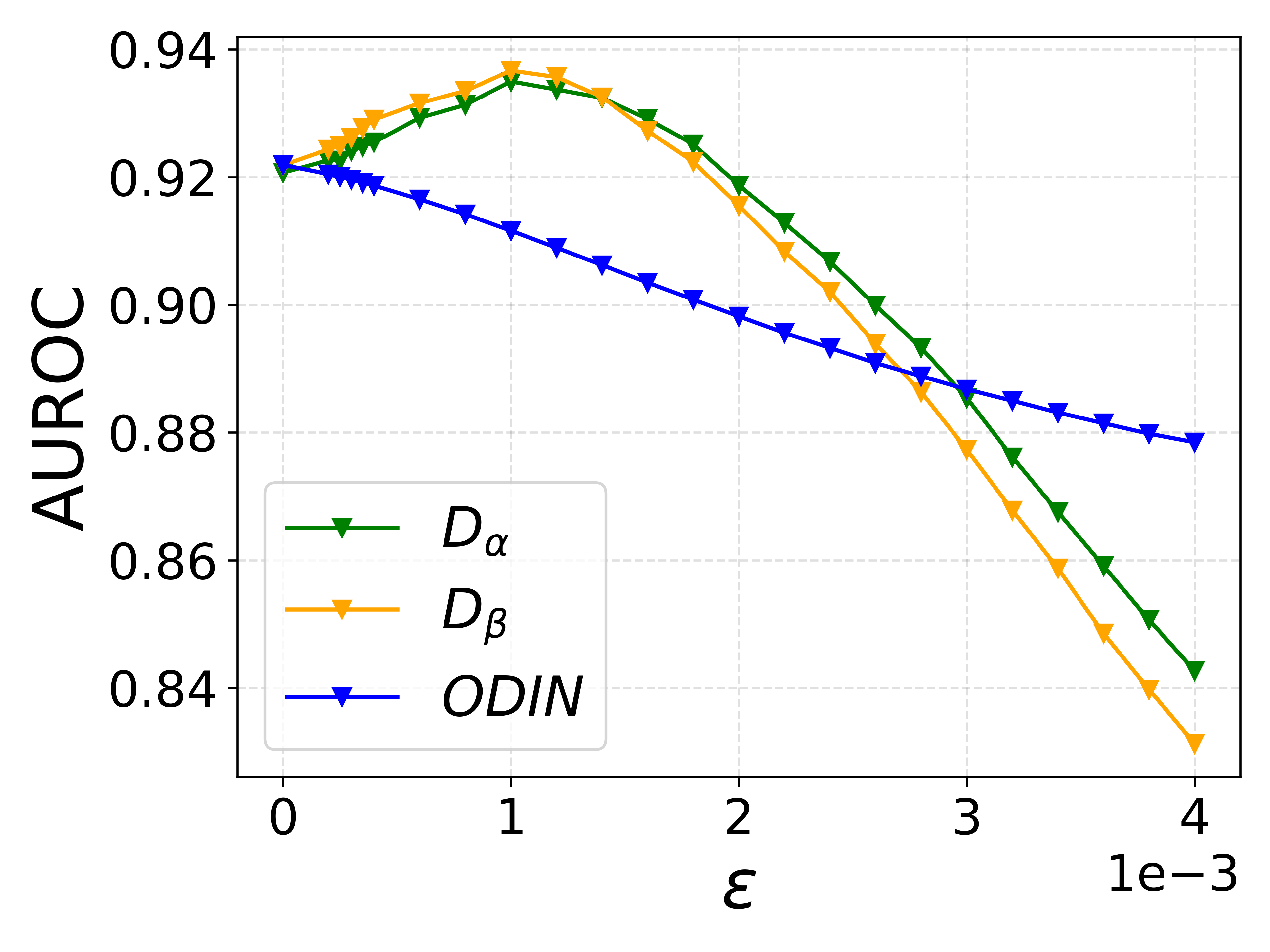}
	    \vspace{-1.5\baselineskip}
	    \caption{SVHN\\\centering{$T=1.5$}}
	    \label{fig:svhn_T_1.5_all}
	\end{subfigure}
	\begin{subfigure}[b]{ 0.23\textwidth}
	    \centering
	    \includegraphics[width=\textwidth]{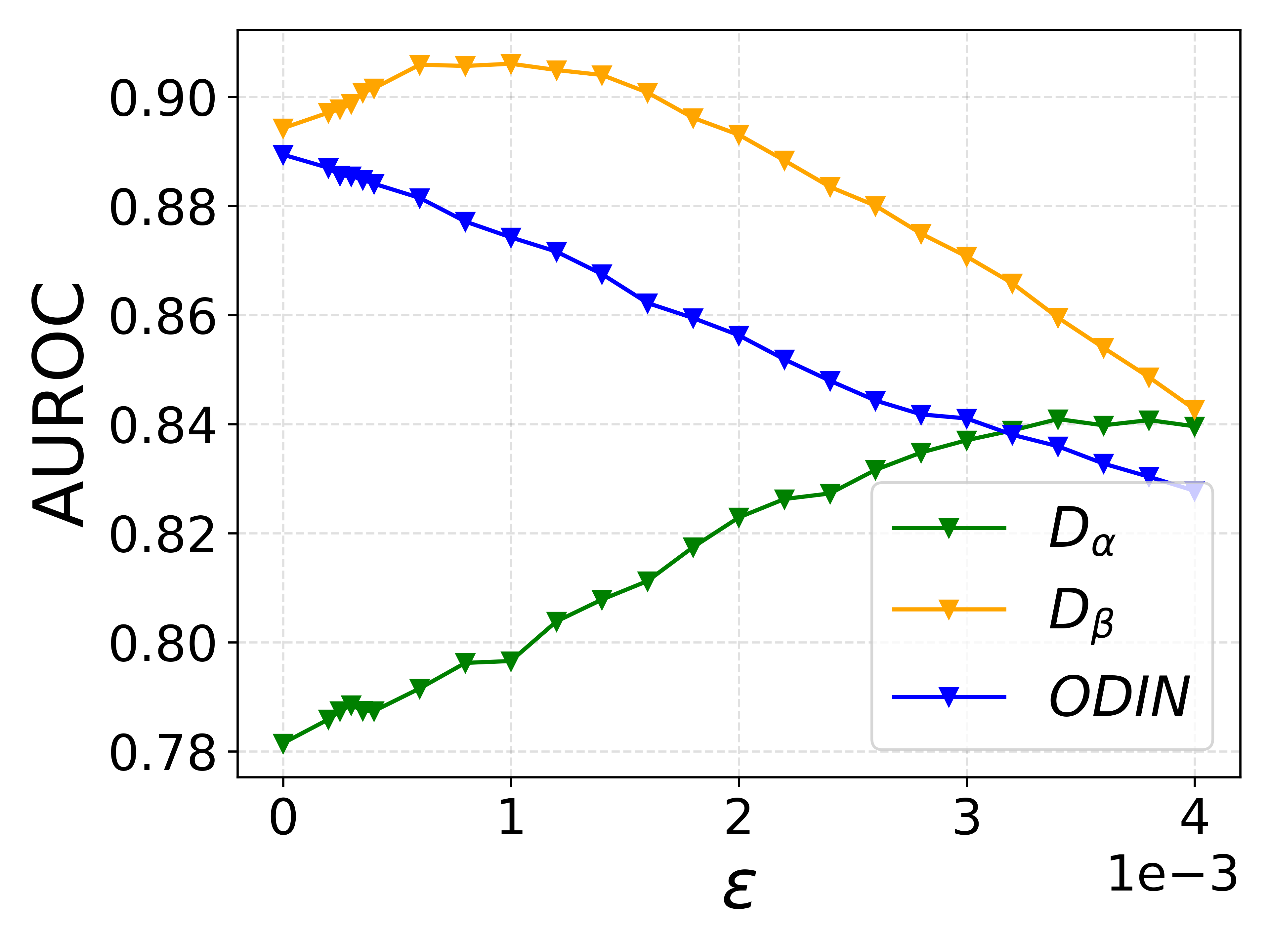}
	    \vspace{-1.5\baselineskip}
	    \caption{SVHN\\\centering{$T=1000$}}
	    \label{fig:svhn_T_1000_all}
	\end{subfigure}
	\label{fig:auc_cifar10_all}
\end{figure*}

We conclude by showing in~\cref{fig:auc_cifar10_all} the test results obtained by varying $T$ and $\epsilon$ in PBB for all the methods. We present $4$ groups of plots (one for each image dataset) and in each plot we pick $T$ from $\{1, 1.3, 1.5, 1000\}$ (the values selected for $D_\alpha$, $D_\beta$, ODIN and MHLNB~\cref{tab:best_aurocs}) and we let $\epsilon$ vary.

\subsubsection{Misclassification detection in presence of out-of-distribution samples}
\label{appendix:ood}
We include in~\cref{tab:ood} the results of all the simulations carried out for detecting misclassification detection in presence of out-of-distribution samples. The experimental setting is reported in~\cref{sec:experiments_setup}.
\begin{table}[!htb]
	\caption{In PBB we set $\epsilon_\alpha = 0.00035$ and $T_\alpha = 1$,
	$\epsilon_\beta = 0.00035$ and $T_\beta = 1.5$, 
	$\epsilon_\text{ODIN} = 0$ and $T_\text{ODIN} = 1.3$. By ODIN$_\text{ood}$, we mean ODIN with the parameter setting as in~\cite{LiangLS2018ICLR}. Since we proceed in a Monte Carlo fashion, the results are reported in terms of \textit{mean / standard deviation}.
	In TBB for by ODIN we report the results of SR, since both methods coincide when $T=1$ and $\epsilon=0$.
	}
	\begin{center}
		\begin{small}
		\resizebox{0.95\textwidth}{!}{
		\begin{sc}
			\begin{tabular}{c|c|c|c|c|c|c|c|c|c|c}
				\toprule
				\multirow{2}{5em}{\textbf{DATASET (In)}} & \multirow{2}{5em}{\textbf{DATASET (Out)}} & \multirow{2}{5em}{\textbf{Scenario}}& \multicolumn{4}{c}{\textbf{AUROC $\%$}} & \multicolumn{4}{|c}{\textbf{FRR $\%$ (95 $\%$ TRR)}}\Bstrut\\\cline{4-11}
								& & & $D_\alpha$ & $D_\beta$ & ODIN & ODIN$_{\text{ood}}$ & $D_\alpha$ & $D_\beta$ & ODIN & ODIN$_{\text{ood}}$\Tstrut\\
								\hline
								\midrule
								\multirow{5}{5em}{\parbox[t][][c]{3cm}{
 CIFAR10\\$\clubsuit$}}  
& \multirow{2}{5em}{iSUN} & PBB & \textbf{95.4} / 0.1  & 95.1 / 0.1  & 94.6 / 0.1  & 89.6 / 0  &  14 / 0.5 &  \textbf{13.5} / 0.4 &  17.2 / 0.3 &  38.9 / 0 \Bstrut\\\cline{3-11}
& & TBB & \textbf{94.6} / 0  & 69.3 / 0.1  & 94.5 / 0.1  & -   &  \textbf{17.7} / 0.1 &  \textbf{17.7} / 0.1 &  \textbf{17.7} / 0 &  -\Tstrut\Bstrut\\\cline{2-11}
& \multirow{2}{5em}{LSUN (crop)} & PBB & \textbf{95.5} / 0.1  & 95.1 / 0  & 94.7 / 0  & 92.6 / 0  &  13.1 / 0.5 &  \textbf{13} / 0.2 &  17.3 / 0 &  31.9 / 0.1\Tstrut\Bstrut\\\cline{3-11}
& & TBB & \textbf{94.4 }/ 0.1  & 69.2 / 0.1  & \textbf{94.4 }/ 0  & -  &  \textbf{17.6} / 0.2 &  \textbf{17.6} / 0.2 &  17.7 / 0.2 &  -\Tstrut\Bstrut\\\cline{2-11}
& \multirow{2}{5em}{LSUN (resize)} & PBB & \textbf{95.4} / 0.1  & 95.1 / 0  & 94.8 / 0  & 89.6 / 0  &  13.4 / 0.6 &  \textbf{13.2} / 0.3 &  17 / 0.3 &  38.9 / 0\Tstrut\Bstrut\\\cline{3-11}
& & TBB & \textbf{94.6} / 0.1  & 69.3 / 0.1  & 94.5 / 0.1  & -  &  \textbf{17.8} / 0.1 &  \textbf{17.8} / 0.1 &  \textbf{17.8} / 0.1 &  -\Tstrut\Bstrut\\\cline{2-11}
& \multirow{2}{5em}{Tiny (crop)} & PBB & \textbf{95.4} / 0  & 95.1 / 0.1  & 94.7 / 0  & 89.6 / 0  &  13.4 / 0.4 &  \textbf{13} / 0.2 &  17.2 / 0.3 &  38.9 / 0\Tstrut\Bstrut\\\cline{3-11}
& & TBB & \textbf{94.6} / 0  & 69.4 / 0.1  & \textbf{94.6} / 0  & - &  \textbf{17.8} / 0.1 &  \textbf{17.8} / 0.1 &  \textbf{17.8} / 0.1 &  -\Tstrut\Bstrut\\\cline{2-11}
& \multirow{2}{5em}{Tiny (res)} & PBB & \textbf{95.2} / 0.1  & 94.9 / 0  & 94.6 / 0.1  & 89.6 / 0  &  \textbf{14} / 0.4 &  \textbf{14} / 0.5 &  17.8 / 0.4 &  38.9 / 0\Tstrut\Bstrut\\\cline{3-11}
& & TBB & \textbf{94.4} / 0.1  & 69.2 / 0  & \textbf{94.4} / 0  & -  &  \textbf{17.8} / 0.1 &  \textbf{17.8} / 0.1 &  \textbf{17.8} / 0.1 &  -\Tstrut\\
\midrule
\multirow{5}{5em}{\parbox[t][][c]{3cm}{
 CIFAR100\\$\clubsuit$}}  
& \multirow{2}{5em}{iSUN} & PBB & \textbf{86.5} / 0.2  & 85.8 / 0  & 85.6 / 0.2  & 79 / 0.1  &  \textbf{45.3} / 1 &  46.1 / 0.5 &  46.8 / 1 &  65.9 / 0.4\Bstrut\\\cline{3-11}
& & TBB & \textbf{85.6} / 0.1  & 82.7 / 0.1  & 85.5 / 0.1  & -  &  46.9 / 0.4 &  \textbf{46.8} / 0.4 &  \textbf{46.8} / 0.4 &  -\Tstrut\Bstrut\\\cline{2-11}
& \multirow{2}{5em}{LSUN (crop)} & PBB & \textbf{89.1} / 0  & 88.5 / 0.1  & 88 / 0.1  & 80.6 / 0  &  \textbf{35.6} / 0.4 &  35.7 / 0.2 &  39.9 / 0.3 &  65.1 / 0\Tstrut\Bstrut\\\cline{3-11}
& & TBB & \textbf{87.9} / 0.1  & 84.9 / 0.1  & 87.7 / 0.1  & -  &  \textbf{39.8} / 0.6 &  \textbf{39.8} / 0.6 &  \textbf{39.8} / 0.6 &  -\Tstrut\Bstrut\\\cline{2-11}
& \multirow{2}{5em}{LSUN (resize)} & PBB & \textbf{86.8} / 0.1  & 86.2 / 0.1  & 86 / 0.1  & 79.1 / 0.1  &  \textbf{44.4} / 0.9 &  \textbf{44.4} / 0.6 &  45.3 / 0.3 &  65.4 / 0.3\Tstrut\Bstrut\\\cline{3-11}
& & TBB & \textbf{85.8} / 0.1  & 82.9 / 0.1  & 85.7 / 0.1  & - &  45.9 / 0.5 &  \textbf{45.8} / 0.5 &  \textbf{45.8} / 0.5 &  -\Tstrut\Bstrut\\\cline{2-11}
& \multirow{2}{5em}{Tiny (crop)} & PBB & \textbf{88.4} / 0.1  & 87.8 / 0.1  & 87.6 / 0.1  & 81.8 / 0.1  &  38.2 / 0.4 &  \textbf{37.8} / 0.9 &  40.6 / 0.5 &  63.4 / 0.1\Tstrut\Bstrut\\\cline{3-11}
& & TBB & \textbf{87.2} / 0.1  & 84.2 / 0.1  & 87 / 0.1  & -  &  \textbf{42} / 0.6 &  \textbf{42} / 0.6 &  \textbf{42} / 0.6 &  -\Tstrut\Bstrut\\\cline{2-11}
& \multirow{2}{5em}{Tiny (res)} & PBB & \textbf{86.8} / 0.1  & 86.3 / 0.1  & 85.9 / 0.1  & 79.2 / 0.1  &  44 / 0.1 &  \textbf{43.6} / 0.2 &  45.9 / 1.2 &  65.8 / 0.3\Tstrut\Bstrut\\\cline{3-11}
& & TBB & \textbf{85.9} / 0.2  & 83 / 0.2  & 85.8 / 0.2  & 85.8 / 0.2  &  \textbf{45.7} / 1.3 &  \textbf{45.7} / 1.3 &  \textbf{45.7} / 1.3 &  -\Tstrut\\
\midrule
				\multirow{5}{5em}{\parbox[t][][c]{3cm}{
 CIFAR10\\$\diamondsuit$}}  
& \multirow{2}{5em}{iSUN} & PBB  & \textbf{95.5} / 0.1  & 95.3 / 0.1  & 94.9 / 0.1 & 91.5 / 0 & 14.4 / 0.6 &  \textbf{13.4} / 0.2 &  16.8 / 0.5 & 34/ 0.1
 \Bstrut\\\cline{3-11}
& & TBB & \textbf{95 }/ 0  & 69.6 / 0  & 94.9 / 0.1  & - &  \textbf{16.4} / 0.2 &  \textbf{16.4} / 0.2 &  \textbf{16.4} / 0.2 & -\Tstrut\Bstrut\\\cline{2-11}
& \multirow{2}{5em}{LSUN (crop)} & PBB & \textbf{95.8} / 0.1  & 95.5 / 0.1  & 95 / 0.1 & 93.9 / 0.1 &  \textbf{12.4} / 0.2 &  12.6 / 0.1 &  16.1 / 0.4 & 24.8 / 0.1\Tstrut\Bstrut\\\cline{3-11}
& & TBB & \textbf{94.8} / 0.1  & 69.6 / 0.1  & \textbf{94.8} / 0.1  & - &  16.7 / 0.4 &  16.8 / 0.4 &  \textbf{16.6} / 0.4 & - \Tstrut\Bstrut\\\cline{2-11}
& \multirow{2}{5em}{LSUN (resize)} & PBB & \textbf{95.8} / 0  & 95.6 / 0  & 95.2 / 0  & 91.6 / 0&  \textbf{12.9} / 0.5 &  \textbf{12.9} / 0.3 &  15.8 / 0.2 & 33.9 / 0
\Tstrut\Bstrut\\\cline{3-11}
& & TBB & \textbf{95} / 0  & 69.7 / 0.1  & \textbf{95} / 0.1  & - &  \textbf{16.4} / 0.2 &  \textbf{16.4} / 0.3 &  \textbf{16.4} / 0.2 & - \Tstrut\Bstrut\\\cline{2-11}
& \multirow{2}{5em}{Tiny (crop)} & PBB & \textbf{95.8} / 0.1  & 95.5 / 0.1  & 95.2 / 0.1  & 91.5 / 0& \textbf{12.8} / 0.7 &  12.9 / 0.5 &  16 / 0 & 33.9 / 0\Tstrut\Bstrut\\\cline{3-11}
& & TBB & \textbf{95} / 0.2  & 69.8 / 0.1  & \textbf{95} / 0.1  & - &  \textbf{16.4} / 0.2 &  16.5 / 0.2 &  \textbf{16.4} / 0.2 & - \Tstrut\Bstrut\\\cline{2-11}
& \multirow{2}{5em}{Tiny (res)} & PBB & \textbf{95.4} / 0.1  & 95 / 0.1  & 94.8 / 0.1  & 91.4 / 0&  15 / 0.1 &  1\textbf{4.8} / 0.7 &  17 / 0.5 & 34.5 / 0.9
\Tstrut\Bstrut\\\cline{3-11}
& & TBB & \textbf{94.6} / 0.2  & 69.3 / 0.2  & \textbf{94.6} / 0.2  & - &  18.1 / 1 &  18.1 / 1.1 &  \textbf{18} / 1 & - \Tstrut\\
\midrule
\multirow{5}{5em}{\parbox[t][][c]{3cm}{
 CIFAR100\\$\diamondsuit$}}  
& \multirow{2}{5em}{iSUN} & PBB & \textbf{84.8} / 0.1  & 84.4 / 0.2  & 84.6 / 0.1  & 80.8 / 0.2 & 53.6 / 1 &  \textbf{51.2} / 0.2 &  51.3 / 0.1 & 63.5 / 0.3\Bstrut\\\cline{3-11}
& & TBB & \textbf{84.1} / 0.1  & 81.2 / 0.1  & 84 / 0.1 & - &  \textbf{52.5} / 0.5 &  \textbf{52.5} / 0.5 & \textbf{52.5} / 0.5 & -\Tstrut\Bstrut\\\cline{2-11}
& \multirow{2}{5em}{LSUN (crop)} & PBB & \textbf{89.9} / 0.1  & 89.6 / 0  & 89 / 0 &  84.1 / 0&  \textbf{35.2} / 0.7 &  35.4 / 0.2 &  39.3 / 0.1 & 62.2 / 0
\Tstrut\Bstrut\\\cline{3-11}
& & TBB & \textbf{88.7} / 0.1  & 85.7 / 0  & 88.5 / 0.1  & - &  \textbf{38.8} / 0.5 &  \textbf{38.8} / 0.5 &  \textbf{38.8} / 0.4 & - 
\Tstrut\Bstrut\\\cline{2-11}
& \multirow{2}{5em}{LSUN (resize)} & PBB & \textbf{85.3} / 0.3  & 85.1 / 0.2  & 84.9 / 0.1 & 81.1 / 0&  51.6 / 0.9 &  \textbf{48.8} / 1 &  49.2 / 0.7 & 63.3 / 0.1\Tstrut\Bstrut\\\cline{3-11}
& & TBB & \textbf{84.6} / 0.2  & 81.8 / 0.2  & \textbf{84.6} / 0.1  & - &  \textbf{50.6} / 0.8 &  50.7 / 0.8 &  \textbf{50.6} / 0.8 & - \Tstrut\Bstrut\\\cline{2-11}
& \multirow{2}{5em}{Tiny (crop)} & PBB & \textbf{88.2} / 0  & 88.1 / 0.2  & 87.7 / 0.1  & 84.8 / 0.1 & 41.2 / 0.3 &  \textbf{40.2} / 0.6 &  42.3 / 0.4 & 59/ 0.2\Tstrut\Bstrut\\\cline{3-11}
& & TBB & \textbf{87.7} / 0.1  & 84.7 / 0.1  & 87.5 / 0.1  & - &  \textbf{41.8} / 0.5 &  \textbf{41.8} / 0.5 &  \textbf{41.8} / 0.5 & - 
 \Tstrut\Bstrut\\\cline{2-11}
& \multirow{2}{5em}{Tiny (res)} & PBB & \textbf{85.4} / 0.2  & 84.8 / 0.2  & 85.1 / 0.3 & 81.2 / 0.1 &  51.8 / 1.6 &  52 / 0.8 &  \textbf{50.4} / 0.9 & 63.3 / 0.2\Tstrut\Bstrut\\\cline{3-11}
& & TBB & \textbf{84.8} / 0.1  & 81.9 / 0.1  & 84.7 / 0.1  & - &  \textbf{51.4} / 0.5 &  \textbf{51.4} / 0.5 &  \textbf{51.4} / 0.5 & - 
 \Tstrut\\
\midrule
				\multirow{5}{5em}{\parbox[t][][c]{3cm}{
 CIFAR10\\$\spadesuit$}}  
& \multirow{2}{5em}{iSUN} & PBB & \textbf{95.6} / 0.1  & \textbf{95.6} / 0  & 95.4 / 0 & 93.5 / 0 &  15.1 / 0.1 &  \textbf{13.6} / 0.5 &  16.1 / 0.2 & 30.6 / 0.4\Bstrut\\\cline{3-11}
& & TBB & \textbf{95.4} / 0.1  & 70 / 0.1  & 95.2 / 0.1  & - &  16.1 / 0.4 &  \textbf{16} / 0.5 &  \textbf{16} / 0.4 & - \Tstrut\Bstrut\\\cline{2-11}
& \multirow{2}{5em}{LSUN (crop)} & PBB & \textbf{96.1} / 0.1  & 95.9 / 0.1  & 95.5 / 0.2  & 95.2 / 0.1 & 12.6 / 0.5 &  \textbf{12.4} / 0.3 &  15.3 / 0.7 & 20.8 / 0.4 \Tstrut\Bstrut\\\cline{3-11}
& & TBB & \textbf{95.2} / 0.1  & 70 / 0.1  & \textbf{95.2} / 0.1  & - &  15.8 / 0.7 &  15.8 / 0.7 &  \textbf{15.7} / 0.7 & - 
 \Tstrut\Bstrut\\\cline{2-11}
& \multirow{2}{5em}{LSUN (resize)} & PBB & \textbf{96} / 0  & 95.8 / 0  & 95.7 / 0 & 93.6 / 0&  13.2 / 0.5 &  \textbf{13} / 0.2 &  15.2 / 0.4 & 30.3 / 0.4
\Tstrut\Bstrut\\\cline{3-11}
& & TBB & \textbf{95.5} / 0.1  & 70.2 / 0.1  & \textbf{95.5}/ 0.1  & - &  15.2 / 0.5 &  15.2 / 0.5 &  \textbf{15.1} / 0.5 & -  \Tstrut\Bstrut\\\cline{2-11}
& \multirow{2}{5em}{Tiny (crop)} & PBB & \textbf{96} / 0.1  & 95.9 / 0.1  & 95.7 / 0  & 93.6 / 0& 13.5 / 0.9 &  \textbf{12.7} / 0.4 &  15.2 / 0.4 & 30.3 / 0.4
\Tstrut\Bstrut\\\cline{3-11}
& & TBB & 95.5 / 0.1  & 70.3 / 0  & \textbf{95.6} / 0  & - &  15.1 / 0.2 &  \textbf{15} / 0.3 &  \textbf{15} / 0.2 & - \Tstrut\Bstrut\\\cline{2-11}
& \multirow{2}{5em}{Tiny (res)} & PBB & \textbf{95.5} / 0.1  & 95.2 / 0.1  & 95.1 / 0.1  & 93.2 & \textbf{14.7} / 0.3 &  14.8 / 0.5 &  17.1 / 0.4 & 31/ 0\Tstrut\Bstrut\\\cline{3-11}
& & TBB & \textbf{94.9} / 0.1  & 69.7 / 0.1  & \textbf{94.9} / 0.1  & - &  16.8 / 0.3 &  16.9 / 0.2 &  \textbf{16.7} / 0.2 & - \Tstrut\\
\midrule
\multirow{5}{5em}{\parbox[t][][c]{3cm}{
 CIFAR100\\$\spadesuit$}}  
& \multirow{2}{5em}{iSUN} & PBB & \textbf{83.3} / 0.1  & 83.1 / 0.1  & 83 / 0.2 & 82.6 / 0.2  &  57.8 / 0.3 &  57.1 / 1 &  \textbf{56.8} / 0.8 & 60/ 0.4\Bstrut\\\cline{3-11}
& & TBB & \textbf{82.6} / 0.2  & 79.7 / 0.2  & 82.5 / 0.2  & - &  \textbf{58.3} / 1 &  58.4 / 1.1 &  58.4 / 1 & -  \Tstrut\Bstrut\\\cline{2-11}
& \multirow{2}{5em}{LSUN (crop)} & PBB & 90.6 / 0  & \textbf{90.7} / 0  & 89.9 / 0.1 & 87.5 / 0&  35.9 / 0.2 &  \textbf{34.6} / 0.2 &  38.5 / 0.4 & 56.1 / 0.2
 \Tstrut\Bstrut\\\cline{3-11}
& & TBB & \textbf{89.4} / 0.1  & 86.2 / 0  & 89 / 0  & - &  \textbf{39.4} / 0.1 &  \textbf{39.4} / 0.1 &  \textbf{39.4} / 0.1 & - \Tstrut\Bstrut\\\cline{2-11}
& \multirow{2}{5em}{LSUN (resize)} & PBB & 83.6 / 0.2  & \textbf{83.8} / 0.1  & 83.6 / 0.2 & 83.2 / 0.1 &  55.8 / 0.4 &  54.2 / 0.7 &  \textbf{54.1} / 0.6 & 59.6 / 0.8\Tstrut\Bstrut\\\cline{3-11}
& & TBB & \textbf{83.2} / 0.1  & 80.4 / 0.1  & \textbf{83.2} / 0.1  & - &  \textbf{55} / 0.6 &  \textbf{55} / 0.7 &  \textbf{55} / 0.6 & - \Tstrut\Bstrut\\\cline{2-11}
& \multirow{2}{5em}{Tiny (crop)} & PBB & 88.3 / 0.1  & \textbf{88.5} / 0.1  & 88.1 / 0.1  & 87.7 / 0.1 & 43.2 / 0.5 &  \textbf{41.5} / 0.7 &  42.9 / 0.4 & 54.3 / 0.1\Tstrut\Bstrut\\\cline{3-11}
& & TBB & \textbf{87.8} / 0  & 84.7 / 0.1  & 87.5 / 0.1  & - &  \textbf{43.7} / 0.2 &  \textbf{43.7} / 0.2 &  \textbf{43.7} / 0.2 & - \Tstrut\Bstrut\\\cline{2-11}
& \multirow{2}{5em}{Tiny (res)} & PBB & 83.8 / 0.1  & 83.8 / 0.1  & \textbf{83.9} / 0.2  & 83/ 0.2 & 57.9 / 0.5 &  56.6 / 0.9 &  \textbf{55.6} / 1 & 61/ 0.6\Tstrut\Bstrut\\\cline{3-11}
& & TBB & \textbf{83.6} / 0.1  & 80.7 / 0.1  & 83.5 / 0.1  & - &  \textbf{55.5} / 0.8 &  \textbf{55.5} / 0.8 &  \textbf{55.5} / 0.8 & - \Tstrut\\
\midrule
\end{tabular}
		\end{sc}
		}
	\end{small}
\end{center}
\label{tab:ood}
\end{table}

{\subsection{\textsc{Doctor} for pure OOD detection}
It is worth emphasizing that DOCTOR is not targeting OOD detection, which is a rather different problem from the one investigated in this paper. So we did not optimize an ad-hoc input perturbation for DOCTOR within the OOD detection setup, i.e. we kept the same input perturbation proposed for the misclassification detection task. The baseline results reported in~\cref{tab:opt_ood} show that DOCTOR is competitive for OOD detection as well since it can reach similar scores or even outperform the baseline (e.g., the simulations with LSUN (CROP) show an improvement of the results of $3.3 \%$ in terms of FRR $\%$). We indicate the methods together with their parameter setting. ODIN$_{\textrm{OOD}}$ denotes the same parameter setting as in~\cite{LiangLS2018ICLR}.

\begin{table}[!htb]
\centering
	\caption{\textsc{Doctor} for pure OOD detection. We set : $\epsilon_\alpha = 0$ and $T_\alpha = 15$, $\epsilon_\beta = 0$ and $T_\beta = 1000$, as in~\cite{LiangLS2018ICLR} for  ODIN$_\text{OOD}$. The baseline results reported below show that DOCTOR is competitive for OOD detection as well since it can reach similar scores or even outperform the baseline.}
	\begin{center}
		\begin{small}
		\resizebox{0.9\textwidth}{!}{
		\begin{sc}
			\begin{tabular}{c|c|c|c|c|c|c|c}
				\toprule
				\multirow{2}{5em}{\centering\textbf{DATASET-In}} & \multirow{2}{5em}{\centering\textbf{DATASET-Out}} & \multicolumn{3}{c}{\textbf{AUROC $\%$}} & \multicolumn{3}{|c}{\textbf{FRR $\%$ (95 $\%$ TRR)}}\Bstrut\\\cline{3-8}
								& & $D_\alpha$ & $D_\beta$  & ODIN$_{\text{ood}}$ & $D_\alpha$ & $D_\beta$ & ODIN$_{\text{ood}}$\Tstrut\\
								\hline
								\midrule
								\multirow{4}{5em}{\parbox[t][][t]{1cm}{\centering
 CIFAR10}}  
& iSUN & 98.1  & 97.9  & \textbf{98.8}  &  8 &  9.1 &  \textbf{6.3} \Bstrut\\\cline{2-8}
& Tiny (res)  & 97.6 & 97.3	& \textbf{98.5} & 9.9 & 11.2 & \textbf{7.2} \Tstrut\Bstrut\\\cline{2-8}
& LSUN (crop) & \textbf{98.6} &	98.2 &	98.2 &	\textbf{5.4} &	6.9 & 8.7\Tstrut\Bstrut\\\cline{2-8}
& Tiny (crop) & 98.9 &	98.5 &	\textbf{99.1} &	4.6 &	6.4 &	\textbf{4.3} \Tstrut\Bstrut\\
\bottomrule
\end{tabular}
		\end{sc}
		}
	\end{small}
\end{center}
\label{tab:opt_ood}
\end{table}

{
\subsubsection{\textsc{Doctor} in presence of OOD samples that are similar to in-distribution ones}
We tested \textsc{Doctor} in pure OOD setting, considering CIFAR100 as in-distribution and CIFAR10 as out-distribution. The results below show that \textsc{Doctor} optimized as in the following paper outperforms ODIN (optimized as described in~\cite{LiangLS2018ICLR}) and ENERGY. This is particularly promising as it shows that \textsc{Doctor}, without performing any training and without been particularly optimized for OOD detection, can perform well on a wider variety of problems. 
\begin{table}[!htb]
\centering
	\caption{Comparison of $D_\alpha$ with ENERGY and ODIN (parameter setting as in~\cite{LiangLS2018ICLR}) when OOD samples are similar to in-distribution samples.}
	\begin{center}
		\begin{small}
		\resizebox{0.9\textwidth}{!}{
		\begin{sc}
			\begin{tabular}{c|c|c|c|c}
				\toprule
				\textbf{DATASET-In} & \textbf{DATASET-Out}& \textbf{METHODS} & \textbf{AUROC $\%$} & \textbf{FRR $\%$ (95 $\%$ TRR)}\\
								\hline
								\midrule
								\multirow{3}{5em}{\parbox[t][][t]{1cm}{\centering
 CIFAR100}} & \multirow{3}{5em}{\parbox[t][][t]{1cm}{\centering
 CIFAR10}} 
& $D_\alpha$ (PBB) & 76.8 & 64.2\Bstrut\\\cline{3-5}
& & ENERGY & 73.3 & 76.4\Tstrut\Bstrut\\\cline{3-5}
&  & ODIN (OOD) & 70.5 & 79.5\Tstrut\Bstrut\\
\bottomrule
\end{tabular}
		\end{sc}
		}
	\end{small}
\end{center}
\label{tab:similar}
\end{table}
}
}
{\subsection{Some observations on the white-box scenario (WB)}
\begin{table}[!htb]
\centering
	\caption{Comparison of MHLNB (WB) and $D_\alpha$ (PBB).}
	\begin{center}
		\begin{small}
		\resizebox{0.9\textwidth}{!}{
		\begin{sc}
			\begin{tabular}{c|c|c|c}
				\toprule
				\textbf{DATASET-In} & \textbf{METHODS} & \textbf{AUROC $\%$} & \textbf{FRR $\%$ (95 $\%$ TRR)}\\
								\hline
								\midrule
								\multirow{3}{5em}{\parbox[t][][t]{1cm}{\centering
 CIFAR10}}  
&  $D_\alpha$ (PBB) & 95.2 & 13.9\Bstrut\\\cline{2-4}
&  MHLNB (WB) & 49.5 & 97.3\Tstrut\Bstrut\\
\midrule
\multirow{3}{5em}{\parbox[t][][t]{1cm}{\centering
 CIFAR100}}
&  $D_\alpha$ (PBB) & 88.2 & 35.7\Bstrut\\\cline{2-4}
&  MHLNB (WB) & 51.6 & 94.9\Tstrut\Bstrut\\
\bottomrule
\end{tabular}
		\end{sc}
		}
	\end{small}
\end{center}
\label{tab:wb}
\end{table}
It is worth clarifying the results in~\Cref{tab:wb} to motivate the performance obtained using the Mahalanobis-based discriminator (MHLNB - WB) for the misclassification detection problem and the issues it raises. First of all, we emphasize that given a network and an input sample \textsc{Doctor} only needs to access the logits output of the network in order to perform the detection. On the contrary, the detector based on Mahalanobis distance consists of 3 steps:
\begin{itemize}
    \item Estimation of the class mean and covariance matrix;
    \item Features extraction according to the Mahalanobis score function;
    \item Aggregation of the scores obtained layer by layer in order to obtain a decision a rule for the discriminator.
\end{itemize}
Clearly, the Mahalanobis distance-based method requires additional samples compared to \textsc{Doctor}. Although estimating the mean and the covariance matrix is possible by exploiting samples from the benchmark tra.et (e.g. CIFAR10, CIFAR100, ...), this method still needs additional (different from training) samples for learning the linear regressor intended to distinguish between correctly (positive) and incorrectly (negative) classified samples. In order to generate the negative samples, we consider the use of adversarial examples generated through Projected Gradient Descent Attack (magnitude of the perturbation 0.0031), which does not assume any knowledge about the test set.}

\end{document}